\documentclass[10pt,twocolumn,letterpaper]{article}

\usepackage{wacv}
\usepackage{times}
\usepackage{epsfig}
\usepackage{graphicx}
\usepackage{amsmath}
\usepackage{amssymb}
\usepackage{booktabs}
\usepackage[accsupp]{axessibility}

\usepackage{mathtools}
\usepackage{xspace} 
\usepackage{booktabs}
\usepackage{multicol}
\usepackage{multirow}
\usepackage{makecell}
\usepackage{color}
\usepackage{colortbl}
\usepackage{placeins}
\usepackage{pifont}  %

\newcommand{\oneincircle}{\ding{172}\xspace}
\newcommand{\twoincircle}{\ding{173}\xspace}
\newcommand{\threeincircle}{\ding{174}\xspace}
\newcommand{\fourincircle}{\ding{175}\xspace}

\definecolor{ballblue}{rgb}{0.0, 0.53, 0.74}
\definecolor{red}{rgb}{1.0, 0., 0.0}
\definecolor{applegreen}{rgb}{0.55, 0.71, 0.0}

\newif\ifArxivVersion
\newcommand{\switchArxiv}[2]{\ifArxivVersion #1\xspace\else #2\xspace\fi}
\ArxivVersiontrue

\def\wacvPaperID{451} %

\wacvalgorithmstrack   %

\wacvfinalcopy %

\ifwacvfinal
\def\assignedStartPage{1} %
\fi

\ifwacvfinal
\usepackage[breaklinks=true,colorlinks,bookmarks=false,pagebackref]{hyperref}
\else
\usepackage[pagebackref=true,breaklinks=true,colorlinks,bookmarks=false]{hyperref}
\fi

\usepackage[capitalize]{cleveref}
\crefname{section}{Sec.}{Secs.}
\Crefname{section}{Section}{Sections}
\Crefname{table}{Table}{Tables}
\crefname{table}{Tab.}{Tabs.}

\newcommand{\topic}[1]{\vspace{1mm}\noindent\textbf{#1}}
\newcommand{\modelname}{SIMPLI\xspace}

\begin{document}

\title{Self-improving Multiplane-to-layer Images for Novel View Synthesis}

\author{
    Pavel Solovev$^{1*}$ \qquad
    Taras Khakhulin$^{1,2*}$ \qquad
    Denis Korzhenkov$^{3}$\thanks{All authors contributed equally.}
    \\[7pt] $^1$Samsung AI Center -- Moscow
    \\ $^2$Skolkovo Institute of Science and Technology
    \\ $^3$Yandex Research
    \\[7pt]  \url{https://samsunglabs.github.io/MLI/}
}

\maketitle
\newlength{\mrgone}
\newlength{\wid}

\begin{abstract}
We present a new method for lightweight novel-view synthesis that generalizes to an arbitrary forward-facing scene.
Recent approaches are computationally expensive, require per-scene optimization, or produce a memory-expensive representation.
We start by representing the scene with a set of fronto-parallel semitransparent planes and afterwards convert them to deformable layers in an end-to-end manner. Additionally, we employ a feed-forward refinement procedure that corrects  the estimated representation by aggregating information from input views.
Our method does not require any fine-tuning when a new scene is processed and can handle an arbitrary number of views without any restrictions.
Experimental results show that our approach surpasses recent models in terms of both common metrics and human evaluation, with the noticeable advantage in inference speed and compactness of the inferred layered geometry. 
\end{abstract}

\section{Introduction}
\label{sec:introduction}

A problem of novel view synthesis (NVS) consists in predicting the view $I_n$ of a scene from a novel camera viewpoint $\pi_n$, given a set of input views $\{I_i\}_{i=1}^V$ for that scene (also referred to as source views) and the corresponding camera poses $\{\pi_i\}_{i=1}^V$ and intrinsic parameters $\{K_i\}_{i=1}^V$~\cite{deep_blending}.
Two natural origins for such input views are handheld videos of static scenes~\cite{nerf,ibrnet,neural_rays} and shots from a multi-camera rig~\cite{deep_view,immersive_lf_video}. To this day, the best image quality is obtained by estimating the radiance field of the scene, deriving it from frames of the source video that are the closest to the novel camera pose~\cite{nerf,Xian2021SpacetimeNI,Barron2021MipNeRF3U,Neff2021DONeRFTR}. 
Nevertheless, such approaches require fine-tuning of the model on a new scene to achieve the best results. This limitation prevents them from being used in settings where fast rendering is needed.

On the other hand, a class of methods based on \emph{multiplane images} (MPI)~\cite{stereomag,singleview_mpi,pushing_bound} provides real-time rendering and good generalization, by representing the scene with a set of fronto-parallel planes given several input images~\cite{llff,deep_view}.
One of their restrictions is the relatively large number of semitransparent planes required to approximate the scene geometry. To cope with this, recent works~\cite{immersive_lf_video,multidepth_panorama} proposed to generate a dense set of planes (as many as 128) and merge them into a \emph{multilayer image} (MLI) with a non-learnable post-processing operation. This research direction was followed by the methods which estimate the MLI end-to-end with a neural network~\cite{worldsheet,stereo_layers}.

In this paper, we present a new method for photorealistic view synthesis, which estimates a multilayer geometry of the scene in the form of an MLI, given an arbitrary set of forward-facing views.
A network that predicts a proxy geometry is trained on a dataset of scenes, and after that, MLI is obtained in a feed-forward fashion for any new scene.
In contrast with prior solutions, our method is free of a predefined number of input views or any neural computation during rendering. Thus,  the output representation can be rendered in high resolution with standard graphics engines even on devices with computational restrictions.

\begin{figure*}[t]
   \centering 
   \includegraphics[width=\linewidth]{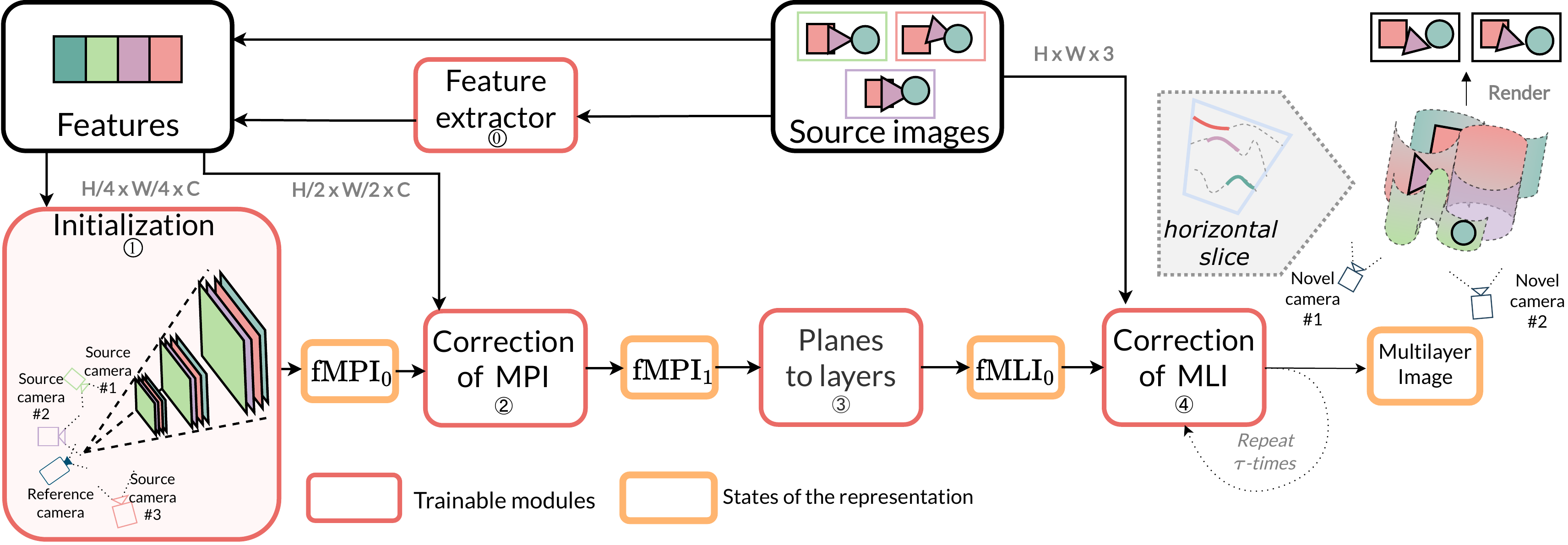}
   \caption{
    Scheme of the proposed \modelname model.
    Our multistage system consists of 4 steps: 
    \oneincircle a preliminary multiplane representation (MPI) is initialized and \twoincircle refined with an error correction step. 
    Afterwards, \threeincircle MPI is converted to the multilayer geometry (MLI), which is again passed through an error correction procedure \fourincircle.
    The whole pipeline is implemented in the coarse-to-fine manner.
    }
   \label{fig:method_scheme}
\vspace{-10pt}
\end{figure*}

We refer to the proposed model as \textbf{S}elf-\textbf{i}mproving \textbf{M}ulti\textbf{p}lane-to-\textbf{l}ayer \textbf{I}mage, or just \modelname.
The basic steps of our approach are outlined in the name: First, we estimate the geometry of a scene in the form of the multiplane image, which is converted into the multilayer image in an \emph{end-to-end} manner, see~\cref{fig:method_scheme}.
Additionally, we refine the representation with a feed-forward \emph{error correction} procedure, inspired by the DeepView paper~\cite{deep_view}, based on input views (hence \emph{self-improving}). 
Our evaluation shows that for test scenes not seen during training, we can synthesize images on par with or superior to state-of-the-art methods designed to generalize to new scenes. Moreover, our method provides faster inference speed.

\begin{figure*}[ht!]
    \setlength{\mrgone}{-0.5cm}
    \setlength{\wid}{0.34\columnwidth}

    \centering
        \begin{tabular}{cccccc }
           \hspace{\mrgone} \includegraphics[width=\wid]{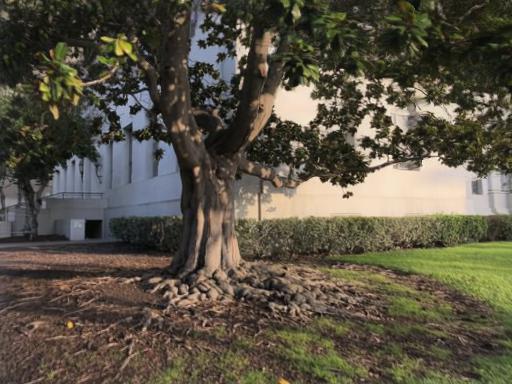}
            &  \hspace{\mrgone} \includegraphics[width=\wid]{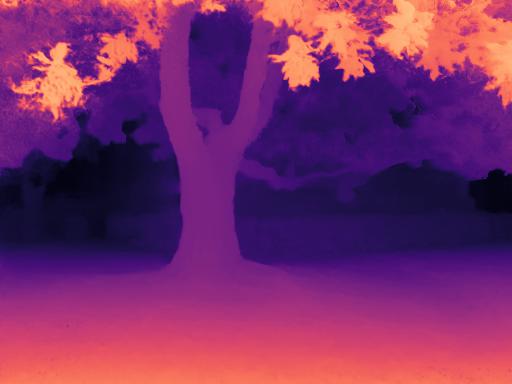}
             & \hspace{\mrgone} \includegraphics[width=\wid]{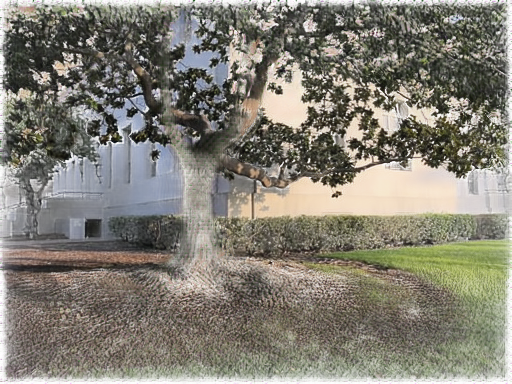}
             & \hspace{\mrgone} \includegraphics[width=\wid]{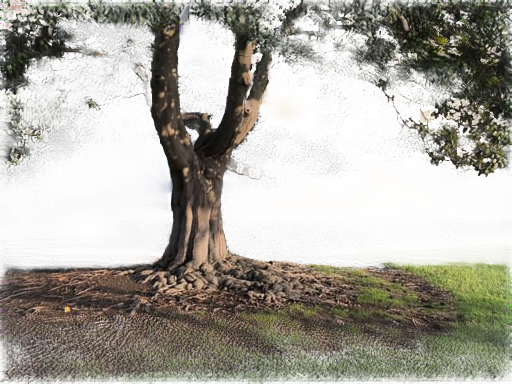}
             & \hspace{\mrgone} \includegraphics[width=\wid]{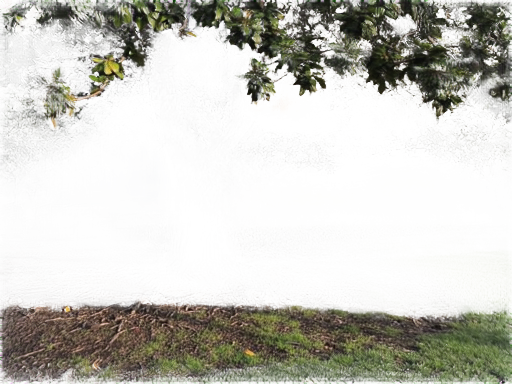}            
             & \hspace{\mrgone} \includegraphics[width
             =\wid]{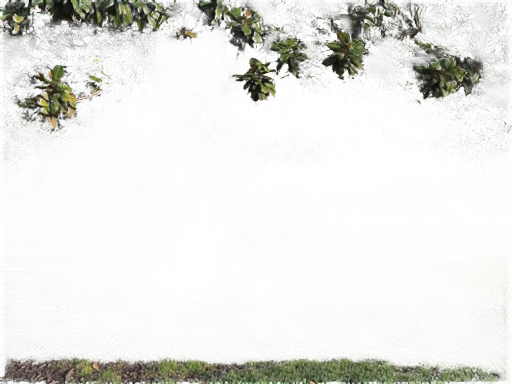}  \\
            
              \hspace{\mrgone} \includegraphics[width=\wid]{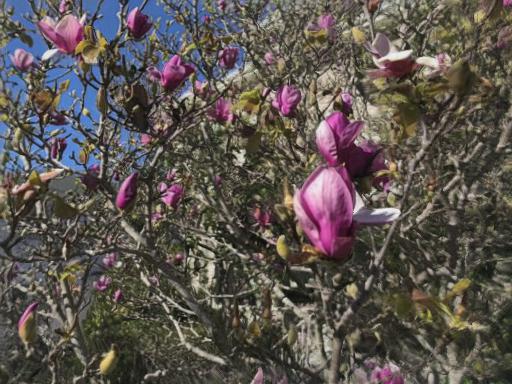}  & \hspace{\mrgone}
            \includegraphics[width=\wid]{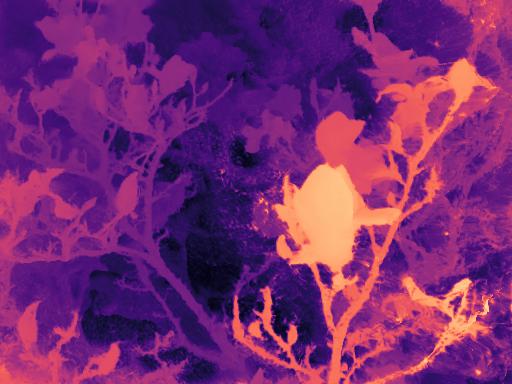} & \hspace{\mrgone}
              \includegraphics[width=\wid]{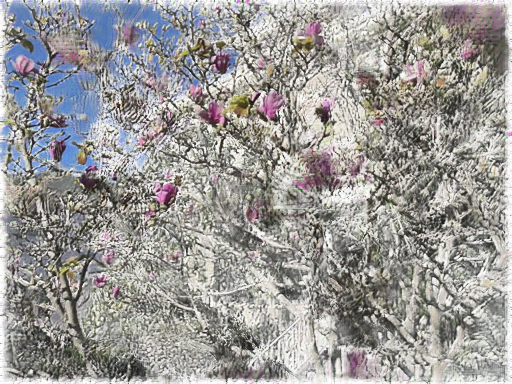} & \hspace{\mrgone}
             \includegraphics[width=\wid]{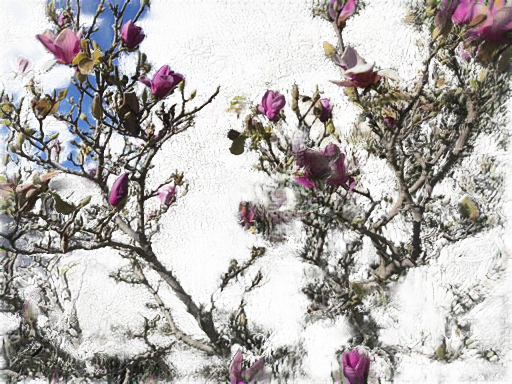} & \hspace{\mrgone}
              \includegraphics[width=\wid]{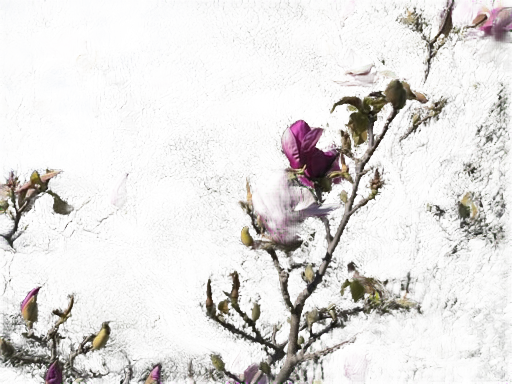} &    \hspace{\mrgone}         
             \includegraphics[width=\wid]{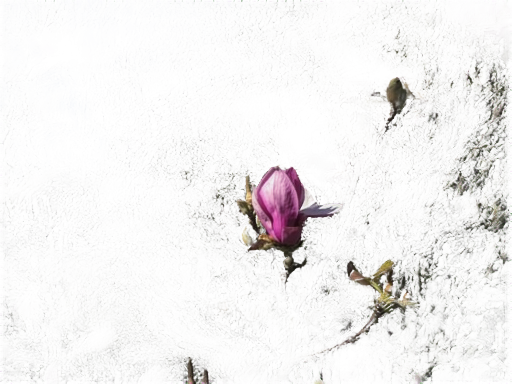} 
              \\
             \hspace{\mrgone} \includegraphics[width=\wid]{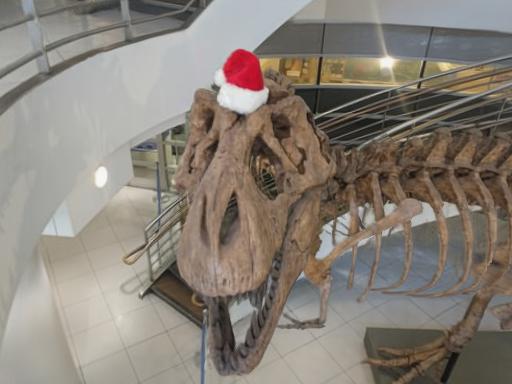} &  \hspace{\mrgone} \includegraphics[width=\wid]{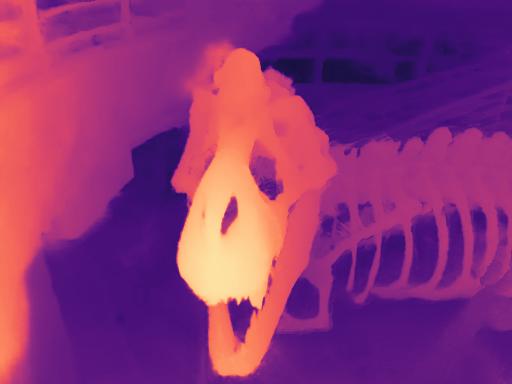} &  \hspace{\mrgone}
             \includegraphics[width=\wid]{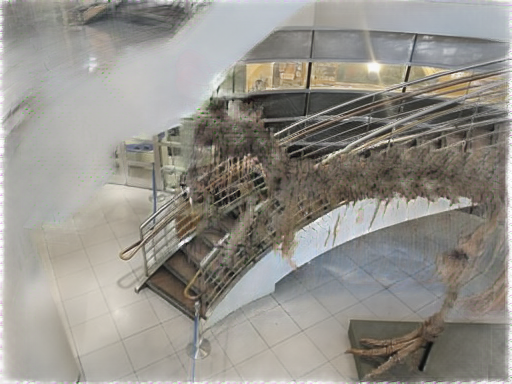} & \hspace{\mrgone}
              \includegraphics[width=\wid]{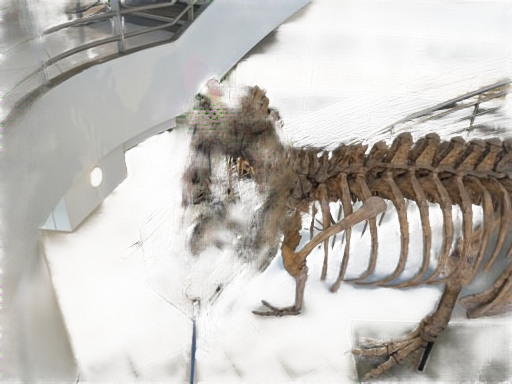} & \hspace{\mrgone}
              \includegraphics[width=\wid]{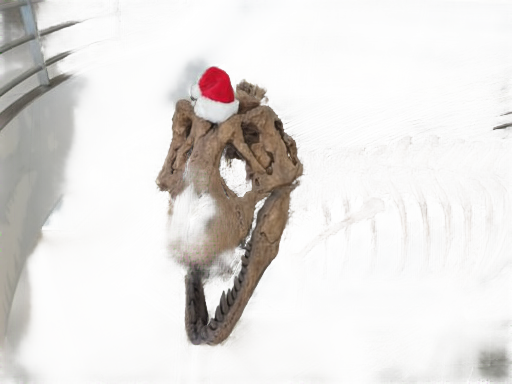}  &  \hspace{\mrgone}         
              \includegraphics[width=\wid]{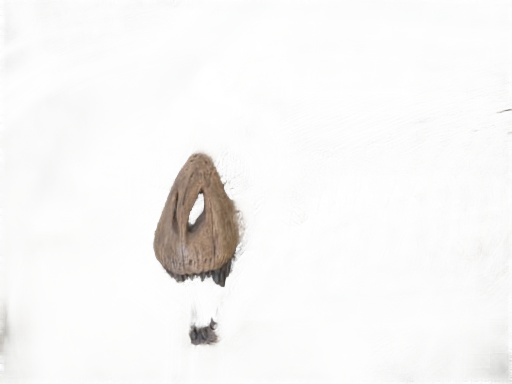} 
              
              \\ 
\hspace{\mrgone} \includegraphics[width=\wid]{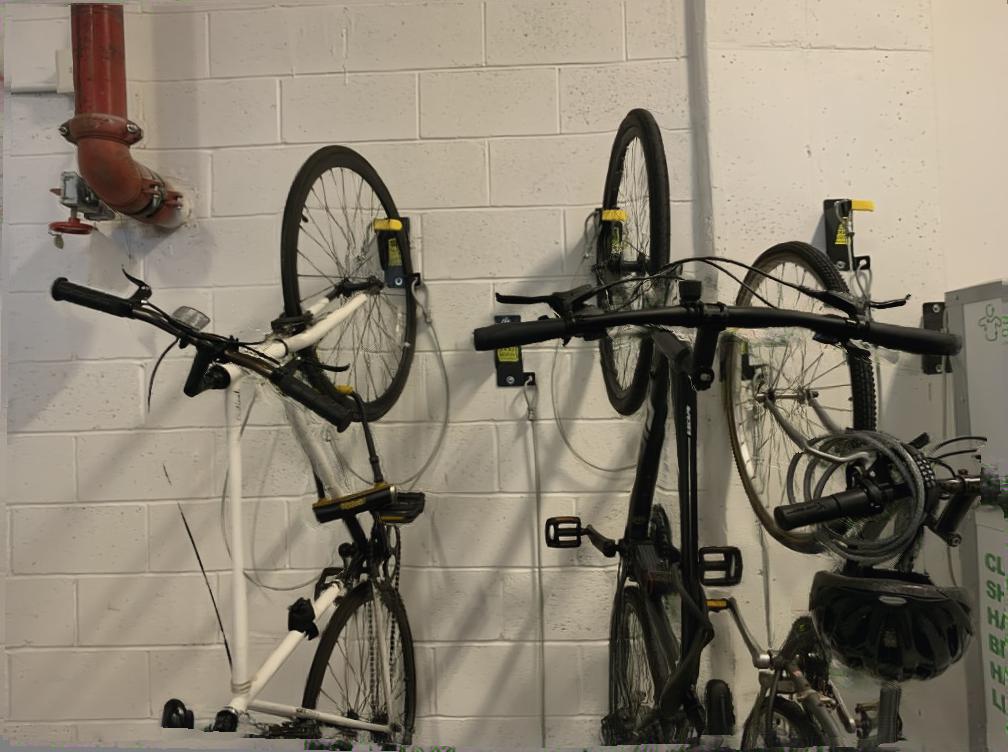} &  \hspace{\mrgone}
 \includegraphics[width=\wid]{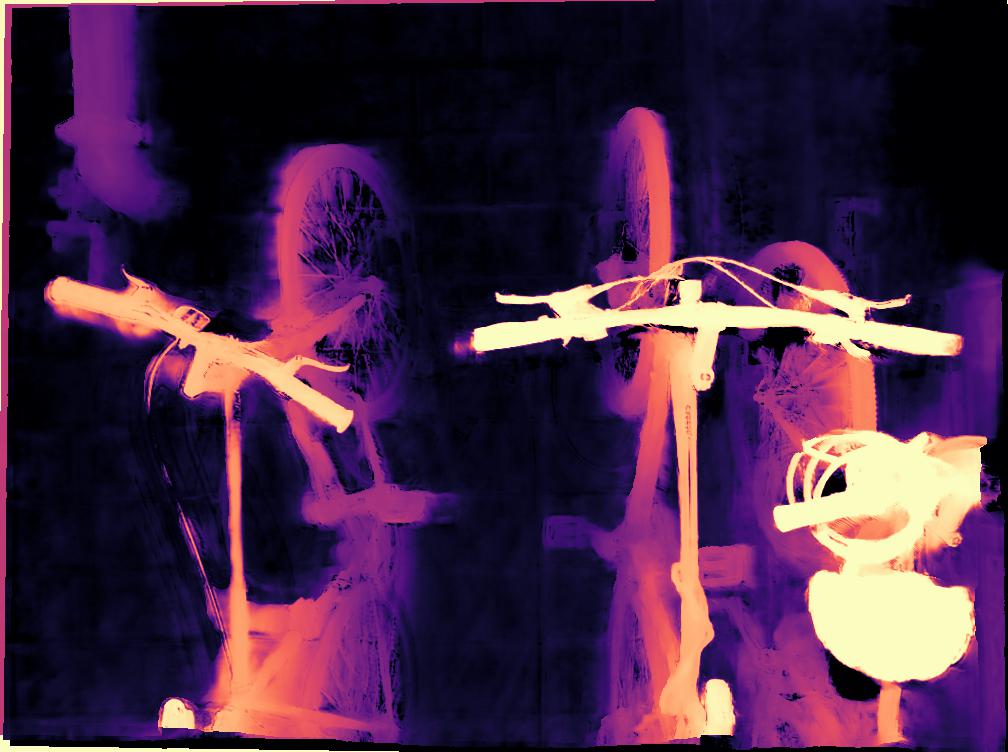} &  \hspace{\mrgone}
\includegraphics[width=\wid]{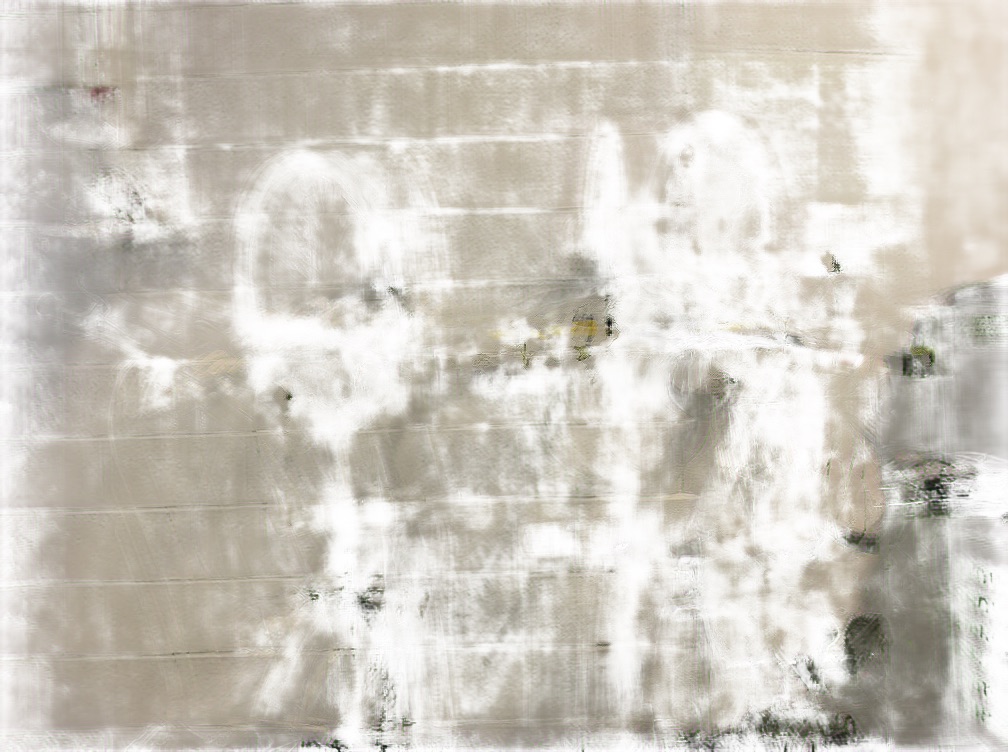} & \hspace{\mrgone}
\includegraphics[width=\wid]{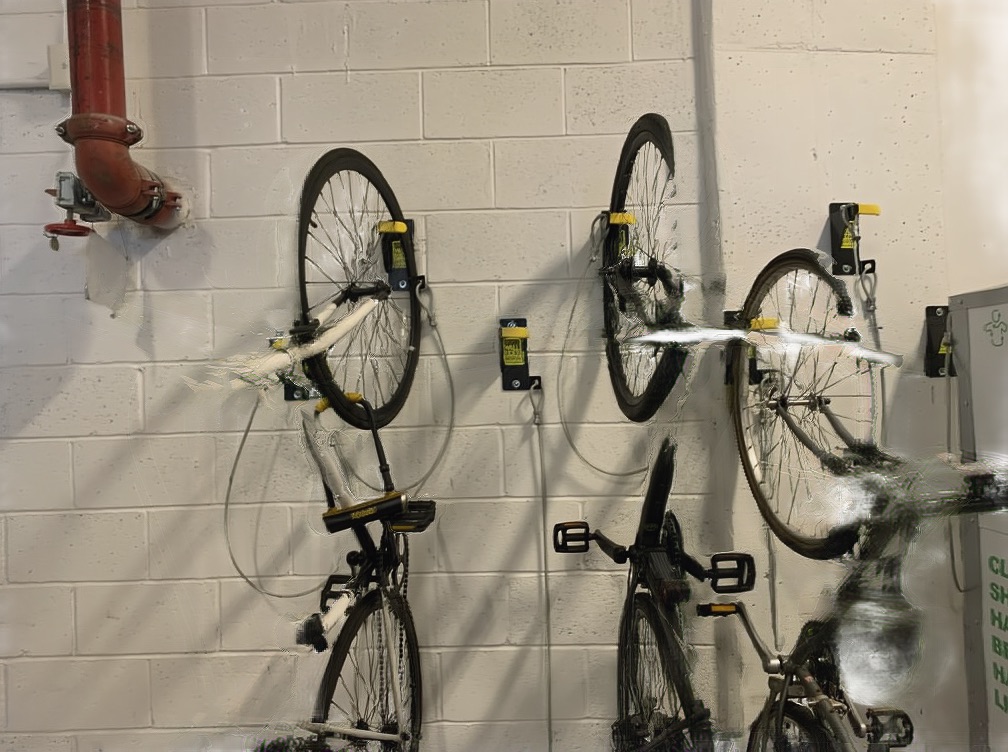} & \hspace{\mrgone}
\includegraphics[width=\wid]{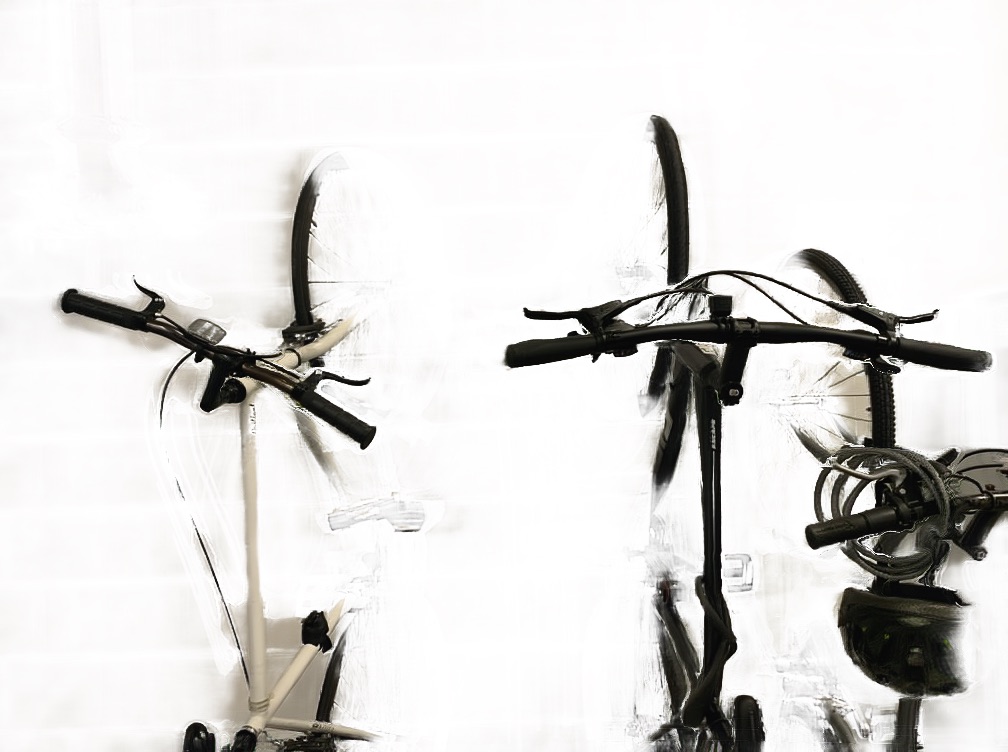} & \hspace{\mrgone}
\includegraphics[width=\wid]{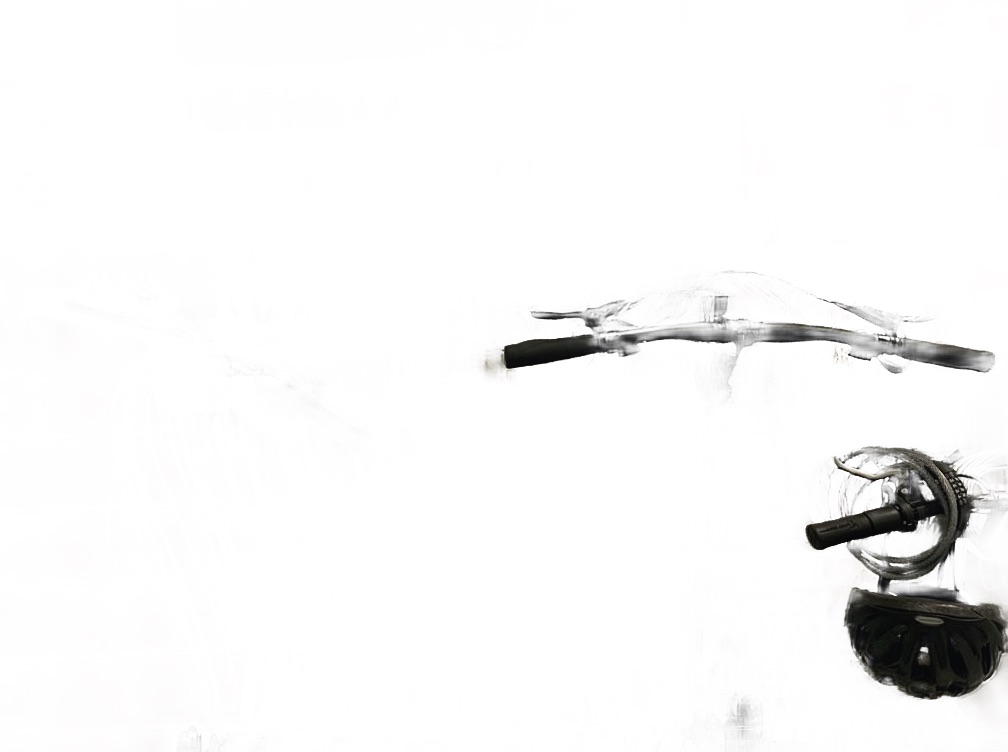} \\
\hspace{\mrgone} Rendered view &  \hspace{\mrgone}
Rendered depth &  \hspace{\mrgone}
Layer 1 (farmost) & \hspace{\mrgone}
Layer 2 & \hspace{\mrgone}
Layer 3 & \hspace{\mrgone}
Layer 4 (nearest) 
\end{tabular}
\vspace{0.5ex}
\caption{
    The MLI representation estimated by \modelname with 4 deformable layers.
    Semitransparent layers are enumerated in the back-to-front order.
    The inferred depth map is computed by overcomposing the per-layer depth maps \wrt the opacity extracted from the corresponding RGBA textures.
}
\label{fig:mli_representation}
\vspace{-1ex}
\end{figure*}

\section{Related works}
\label{sec:related_works}

Novel view synthesis for a static scene is a well-known problem of computer vision. The main task in this field is to produce plausible images corresponding to the camera's motion through the scene, based on the input information.
Early methods interpolated directly between the pixels of the input images with restrictions on the positions of the source cameras~\cite{lumigraph,learning_lf,hedman2016scalable}, used the proxy geometry of the scene~\cite{debevec96} or optimization accompanied with heuristics~\cite{soft3D} to render the scene from a new viewpoint.

The approach called \emph{Stereo magnification} used MPI geometry with semitransparent planes placed in the frustum of one of the source cameras~\cite{stereomag}. 
The \emph{DeepView} method refined the textures of planes step-by-step with a procedure similar to the learned gradient descent~\cite{deep_view,immersive_lf_video}. Although it was reportedly able to handle any number of input images, this was not demonstrated.
The authors of the \emph{LLFF} method~\cite{llff} built a separate MPI in the frustum of each source camera and used a heuristic to blend multiple preliminary novel views obtained using those MPIs to get the resulting image. In contrast, we aggregate information from an arbitrary number of input views and construct a single representation for a scene.

Several works~\cite{3d_layered_inpainting,Dhamo2019ldi} found the usage of layered depth image (LDI)~\cite{ldi} efficient for single-image NVS, due to its sparsity and simplicity of rendering. At the same time, estimating this representation in a differentiable way without relaxation is still challenging~\cite{lsiTulsiani18}. 
\emph{StereoLayers}~\cite{stereo_layers} and \emph{Worldsheet}~\cite{worldsheet} methods represented a scene with a number of semitransparent deformable layers, a structure also called the ``multilayer image'', which is less compact than LDI (see the discussion on terminology in \switchArxiv{\cref{sec:architecture_details}}{Supplementary}).
Nevertheless, due to its scene-adaptive nature, this proxy geometry is still much more lightweight than MPI while preserving the ability to render in real time with modern graphic engines.
Some prior works proposed to convert MPI to MLI with a postprocessing procedure, based on a heuristic~\cite{immersive_lf_video,multidepth_panorama}.
Unlike them, we perform this conversion end-to-end, using the techniques applied in the StereoLayeres~\cite{stereo_layers} paper to the case of only two input frames.

Recently, several methods~\cite{pixelnerf,mvsnerf,GRF,ibrnet,srfnet,tancik2021learned} have attempted to estimate the implicit 3D representation from input images in the form of a neural radiance field~\cite{nerf}. However, massive queries to the neural network at inference time result in slow rendering speed. Moreover, \emph{IBRNet}~\cite{ibrnet} additionally applied the self-attention operation along the ray to estimate the volumetric density of the points, further slowing the speed. 
Another group of methods~\cite{nex,plenoxel,directvox,instantngp} improved the speed of querying and training convergence for a single scene, but these approaches have not been generalized to new scenes.
In contrast, our system allows rendering novel views at higher speed while generalizing across various scenes.

\section{Preliminaries}
\label{sec:preliminaries}

\topic{MPI representation.}
MPI represents a scene with a sequence of $P$ planes, placed in the frustum of a pinhole camera, referred to as a \emph{reference} camera.
We construct a virtual reference camera by averaging the poses of the source cameras~\cite{markley_averaging_2007}. 
These $P$ planes are placed with uniform disparity in the predefined depth range, which depends on the preprocessing of the dataset.
In our experiments $P = 40$.

Each plane has a  semitransparent RGBA texture of resolution $h \times w$, disregarding its depth.
To render the final image using this representation, the planes' textures are ``projected'' to the novel camera pose using homography warping and composed-over \wrt their opacities~\cite{compose_over}.
Therefore, rendering an arbitrary number of new images is fast and does not involve any neural networks.
The critical question here is how to estimate the textures of the planes.
Typically, this is done by processing a \emph{plane-sweep volume} (PSV).

\topic{Building the PSV.}
Here we define the procedure of building a plane-sweep volume $\mathit{PSV} = \texttt{unproj}\left(\{F_i\}_{i=1}^V, h, w\right),$ given a set of $V$ source feature tensors $\{F_i\}_{i=1}^V$ consisting of $C$ channels and the resolution of MPI planes $h \times w$.
The features $\{F_i\}$ may coincide with the source images $\{I_i\}$ or be the outputs of some encoding network.
To construct the volume, each of the feature tensors is ``unprojected'' onto the planes with homography warping.
The result of the procedure $\texttt{unproj}$ is a tensor of shape $V \times P \times C \times h  \times w$.
Informally speaking, to obtain MPI from the built PSV, we need to ``reduce'' the $V$ axis, \ie to aggregate all source features, and subsequently convert them to RGBA domain~\cite{stereomag,deep_view}.

\topic{Attention pooling}. Attention pooling~\cite{lee_set_2019} is a modification of a standard QKV-attention module~\cite{attentionisallyouneed}, where queries are represented with trainable vectors called \emph{anchors}, independent of input, while keys and values are equal to input vectors.
This operation allows to compress an arbitrary number of inputs to the predefined number of outputs.

\begin{table*}[t]
\centering
\resizebox{\textwidth}{!}{
\begin{tabular}{llccccccccccc} %
    \toprule
    \multirowcell{2}{ \vspace{-4pt} \# \\ [-1pt] source \\ [-1pt] views} & \multirowcell{2}[0pt][l]{Model}
     & \multicolumn{3}{c}{\textbf{SWORD} }
     && \multicolumn{3}{c}{\textbf{Real Forward-Facing} }
     && \multicolumn{3}{c}{\textbf{Shiny} } \\
    \cmidrule{3-5}  \cmidrule{7-9}  \cmidrule{11-13} 
    && PSNR $\uparrow$ & SSIM $\uparrow$ & LPIPS $\downarrow$  &&  
    PSNR $\uparrow$ & SSIM $\uparrow$ & LPIPS $\downarrow$ && PSNR $\uparrow$ & SSIM $\uparrow$ & LPIPS $\downarrow$ \\
    \midrule
    \multirowcell{4}{2}
    & IBRNet    & 19.02$_{4.59}$ & 0.54$_{0.19}$ & 0.35$_{0.19}$   && 19.13$_{3.07}$ & 0.57$_{0.14}$ & 0.32$_{0.13}$   && 21.89$_{3.76}$ & 0.67$_{0.15}$ & 0.24$_{0.14}$  \\
    & StereoMag                 & 18.71$_{3.95}$ & 0.53$_{0.19}$ & 0.29$_{0.18}$   &&  17.22$_{2.85}$ & 0.47$_{0.15}$ & 0.31$_{0.12}$             && 20.11$_{3.63}$ & 0.58$_{0.16}$ & 0.19$_{0.11}$  \\
    & DeepView$^{\dagger}$ & 20.41$_{4.04}$ & 0.64$_{0.17}$ & 0.22$_{0.14}$ && 20.46$_{3.00}$ & 0.65$_{0.13}$ & 0.20$_{0.08}$    && 22.96$_{3.73}$ & 0.72$_{0.14}$ & 0.12$_{0.07}$     \\

    & \modelname-4L                     & 20.78$_{3.83}$ & 0.64$_{0.16}$ & 0.23$_{0.15}$  && 20.46$_{3.00}$ & 0.65$_{0.13}$ & 0.20$_{0.08}$    && 22.96$_{3.73}$ & 0.72$_{0.14}$ & 0.12$_{0.07}$       \\
    & \modelname-8L                     & \bf{20.84}$_{3.76}$ & \bf{0.64}$_{0.16}$ & \bf{0.22}$_{0.14}$ &&  \bf{21.17}$_{3.09}$ & \bf{0.69}$_{0.12}$ & \bf{0.16}$_{0.06}$  && \bf{23.59}$_{3.27}$ & \bf{0.76}$_{0.12}$ & \bf{0.10}$_{0.05}$  \\
    \midrule
    \multirowcell{4}{5}
    & IBRNet                            & 22.79$_{3.92}$ & 0.71$_{0.15}$ & 0.22$_{0.12}$  && 22.69$_{3.35}$ & 0.73$_{0.10}$ & 0.19$_{0.08}$   && 25.29$_{3.30}$ & 0.80$_{0.09}$ & 0.13$_{0.07}$  \\
    & LLFF                              & 19.56$_{3.19}$ & 0.52$_{0.17}$ & 0.33$_{0.11}$ && 21.76$_{3.02}$ & 0.72$_{0.10}$ & 0.20$_{0.07}$    && 23.31$_{2.74}$ & 0.75$_{0.11}$ & 0.16$_{0.05}$  \\
    & DeepView$^{\dagger}$ & 21.99$_{3.85}$ & 0.73$_{0.14}$ & 0.18$_{0.12}$  && 23.11$_{2.86}$ & 0.76$_{0.10}$ & 0.13$_{0.05}$   && 24.99$_{3.24}$ & 0.81$_{0.09}$ & 0.09$_{0.04}$     \\
    & \modelname-4L   & 22.95$_{3.01}$ & 0.74$_{0.12}$ & \bf{0.17}$_{0.13}$   && 23.37$_{3.12}$ & 0.78$_{0.09}$ & 0.12$_{0.05}$     && \bf{25.47}$_{2.73}$ & 0.83$_{0.07}$ & 0.08$_{0.03}$   \\
    & \modelname-8L                     & \bf{23.10}$_{3.09}$ & \bf{0.75}$_{0.11}$ & \bf{0.17}$_{0.12}$ &&  \bf{23.58}$_{3.06}$ & \bf{0.79}$_{0.09}$ & \bf{0.11}$_{0.04}$ && \bf{25.47}$_{2.64}$ & \bf{0.83}$_{0.07}$ & \bf{0.07}$_{0.03}$    \\
    \midrule
    \multirowcell{4}{8}
    & IBRNet                            & \bf{24.51}$_{3.16}$ & 0.77$_{0.10}$ & 0.18$_{0.07}$ && \bf{23.98}$_{3.31}$ & 0.78$_{0.09}$ & 0.16$_{0.06}$    && \bf{26.27}$_{2.80}$ & 0.83$_{0.07}$ & 0.11$_{0.05}$   \\
    & LLFF                              & 21.22$_{3.15}$ & 0.59$_{0.17}$ & 0.28$_{0.10}$  && 22.91$_{3.08}$ & 0.77$_{0.08}$ & 0.17$_{0.05}$   && 24.29$_{2.83}$ & 0.78$_{0.10}$ & 0.14$_{0.05}$  \\
    & DeepView$^{\dagger}$  & 22.71$_{3.60}$ & 0.77$_{0.11}$ & 0.16$_{0.09}$  && 23.79$_{2.49}$ & 0.80$_{0.07}$ & 0.11$_{0.03}$   && 25.71$_{2.75}$ & 0.84$_{0.07}$ &0.07$_{0.03}$     \\

    & \modelname-4L                & 24.02$_{3.37}$ & 0.78$_{0.11}$ & 0.16$_{0.10}$   && 23.86$_{2.87}$ & 0.80$_{0.08}$ & \bf{0.10}$_{0.03}$   && 26.03$_{2.36}$ & \bf{0.85}$_{0.05}$ & 0.07$_{0.02}$  \\
    & \modelname-8L                     & 24.32$_{3.37}$ & \bf{0.79}$_{0.11}$ & \bf{0.15}$_{0.10}$ && 23.81$_{2.66}$ & \bf{0.81}$_{0.07}$ & \bf{0.10}$_{0.03}$   && 26.05$_{2.29}$ & \bf{0.85}$_{0.05}$ &  \bf{0.06}$_{0.02}$     \\
    \bottomrule
    \end{tabular}
}
\vspace{0.7ex}
\caption{%
    Results of evaluation on the hold-out test part of SWORD (30 scenes), RFF dataset (8 scenes)~\cite{llff}, and Shiny dataset (8 scenes)~\cite{nex}.
    $V$ denotes the number of source views.
    MPI representation for both LLFF and DeepView$^{\dagger}$ consists of 40 planes. The dagger ${\dagger}$ indicates our re-implementation of the model.
    Note that the difference in metrics between our model and IBRNet is most often not significant due to the large std (indicated by subscripts), while \modelname produces a more compact representation, suitable for real-time rendering.
}
\label{tab:main_scores}
\end{table*}

\begin{table}[t]
\centering
\begin{minipage}{.9\columnwidth}\resizebox{\textwidth}{!}{
    \begin{tabular}{lcc ccc} %
    \toprule
    Model
    & & PSNR $\uparrow$ & SSIM $\uparrow$ & LPIPS $\downarrow$ \\
    \midrule
    IBRNet                    &&  27.02$_{2.09}$ & 0.84$_{0.05}$ & 0.11$_{0.03}$      \\
   DeepView                                &&  \textbf{29.52}$_{2.92}$ &   \textbf{0.90}$_{0.06}$ &   \textbf{0.05}$_{0.01}$   \\
     DeepView$^\dagger$   && 28.54$_{2.35}$ & 0.89$_{0.04}$ & \bf{0.05}$_{0.02}$  \\
    \modelname-4L                 && 27.73$_{1.86}$ & 0.87$_{0.04}$ & 0.07$_{0.02}$      \\
     \modelname-8L                 && 28.01$_{1.88}$ & 0.88$_{0.04}$ & 0.06$_{0.01}$         \\
    \bottomrule
    \end{tabular}
}\end{minipage}
\vspace{0.7ex}
\caption{Results of evaluation on the test part of Spaces dataset (10 scenes, large baseline)~\cite{deep_view}. 
    The number of source views equals 4.
    DeepView (original) and DeepView$^\dagger$ (our modification) both employ 40 planes while \modelname produces only 4-8 layers.
    For this dataset one needs to trade off quality for representation compactness.
}
\label{tab:spaces_scores}
\end{table}

\section{Method}
\label{sec:method}
\subsection{Overview}
\label{subsec:overview}
\topic{\oneincircle Initialization.}
First, we process each of the $V$ source views with a feature extractor based on the feature pyramid architecture $E_\theta: I_i \mapsto F'_i$ yielding $V$ tensors of the same resolution $H \times W$ as original images.
Each tensor $F'_i$ is concatenated with $I_i$ channel-wise, providing a feature tensor $F_i$.
The features $\{F_i\}_i$ are used to build the plane-sweep volume $\mathit{PSV}_0$ of resolution $\frac{H}{4} \times \frac{W}{4}$ with $P$ planes. The volume serves as input for the aggregation module $T_0$ (described below).
The output of this module is an initial version of MPI in the feature domain, denoted as $\mathit{fMPI}_0$.

\topic{\twoincircle Correction of MPI.} Then we conduct an error correction step. We project $\mathit{fMPI}_0$ on the source cameras at resolution of $\frac{H}{2} \times \frac{W}{2}$ and compute the difference with feature tensors $\{F_i\}$, downsampled to the same size.
Using the ``unprojected'' tensor of discrepancies of shape  \resizebox{0.35\hsize}{!}{$V \times P \times C \times \frac{H}{2} \times \frac{W}{2}$}, we update the representation to the state $\mathit{fMPI}_1$.

\topic{\threeincircle Planes-to-layer conversion.}
At this step, we  merge the $P$ rigid planes of $\mathit{fMPI}_1$ into $L$ deformable layers (see the detailed diagram in \switchArxiv{\cref{fig:method_diagrams}}{Supplementary}).
The procedure is performed in two steps:
First, attention pooling (see~\cref{sec:preliminaries}) with $L$ anchors is applied along the $P$ axis to aggregate planes into $L$ groups.
The resulting tensor contains textures in the feature domain for the layers.
Second, the depth maps for the deformable layers are predicted: a self-attention module is applied along the $P$ axis of $\mathit{fMPI}_1$ to predict the pixel-wise opacity for each plane. Then we divide the planes into $\frac{P}{L}$ consecutive groups of equal size, and the depths of the planes are overcomposed within each group with predicted opacities. This produces $L$ depth maps which do not intersected each other by design.
Third, to obtain a mesh from each depth map, we treat each pixel of the map as a vertex and connect it to the six nearby vertices (top, right, bottom-right, bottom, left, top-left) with edges. 
Therefore, each quad ($2 \times 2$ block of pixels) is converted into two triangle faces. 
The obtained multilayer representation is denoted as $\mathit{fMLI_0}$.

\topic{\fourincircle Correction of MLI.}
After the multilayer geometry $\mathit{fMLI_0}$ is obtained, we perform the correction step similar to \twoincircle, increasing the resolution of our representation to $H \times W$. %
Since the multilayer representation is relatively lightweight, it is possible to perform multiple error correction steps at this stage before obtaining the final representation. The number of such steps is denoted as $\tau$.
The influence of these additional steps is discussed in the ablation study below.
The updated state $\mathit{fMLI}_\tau$ is converted from the feature domain to RGBA, and we refer to this final state as $\mathit{MLI}$.
Examples of the resulting representations are presented in \cref{fig:mli_representation}.

\subsection{Error correction}
\label{subsec:method_details}
Here we provide more details on the error correction procedure.
It consists of computing the discrepancy with the input views and updating the representation based on this information.
This procedure is performed similarly for MPI and MLI, therefore we assume the case of MPI for simplicity.
Also we slightly abuse the notation in comparison with the previous subsection, as the described procedure does not depend on the exact step of our pipeline.
The detailed schemes of all the steps are provided in \switchArxiv{\cref{fig:method_diagrams,fig:aggregating_modules}}{Supplementary}.

\topic{Discrepancy computation.}
Let $\mathit{fMPI}$ (MPI in the feature domain) of the shape $P \times C \times h \times w$ be the input of the procedure.
We use an \emph{RGBA decoder} to predict RGB color and opacity from features.
RGBA decoder is implemented with a fully connected network, applied to each of $P \times h \times w$ positions independently. 
The predicted color and opacity are concatenated with the original $\mathit{fMPI}$ along the channel axis before rendering on the source cameras. After rendering, we obtain tensors $\{\hat{F}_i\}_{i=1}^V$ of resolution $h' \times w'$, which is the target resolution for the error computation. 
Then we compute the pixel-wise difference between the original feature tensor and the rendered view.
These discrepancies are ``unprojected'' back onto the planes and serve as input to the aggregating block.
Weights of the aggregating blocks and RGBA decoders are not shared between different steps of the pipeline.

In case of MPI, rendering is done with homography warping, while MLI requires the usage of a differentiable renderer~\cite{nvdiffrast}.
Additionally, in case of MLI, the \texttt{unproj} operation ``unprojects'' the inputs on the predefined deformable layers instead of rigid planes.
As the correction of MLI is done in full resolution, for the sake of memory consumption, we do not concatenate features to the predicted color before rendering and compute an error in the RGB domain only.

\topic{``Unprojection'' on the deformable layers.}
In our pipeline the resolution of the depth maps and textures of layers are equal.
Therefore, there exists a one-to-one correspondence between the vertices of the mesh layers and the texels with integer coordinates.
To ``unproject'' a feature tensor, related to the certain source view, to the deformable layers, we project each vertex on that view using the pinhole camera model and take the feature value from the given tensor with bilinear grid sampling.
This operation does not require us to use any differentiable renderer.

\topic{Aggregating modules.}
The aggregating module $T_{\theta}$ receives the current state $\mathit{fMPI}$ of shape $P \times  C \times h' \times w'$  and the unprojected discrepancies $\mathit{PSV}$  of the shape $V \times P \times \left(3 + C\right) \times h' \times w'$ (3 extra channels of $\mathit{PSV}$ correspond to the output of the RGBA decoder).
The self-attention module is applied to $\mathit{PSV}$ along the $V$ axis, following by the attention pooling with a single anchor.

The output of the pooling is concatenated with $\mathit{fMPI}$ channel-wise, as well as mean and variance calculated along the same $V$ axis of $\mathit{PSV}$.
The resulting tensor is passed through convolutional residual blocks~\cite{resnet}, and each of the $P$ planes is processed independently.
The output is treated as a new state of $\mathit{fMPI}$.

At initialization \oneincircle, the aggregating module $T_0$ receives only the $\mathit{PSV}$, since there is no previous state of representation.
Therefore, ResNet blocks operate on the output of the pooling step only.
Also, as the resolution is low at this step, we employ 3D convolutional kernels, while at steps \twoincircle and \fourincircle 2D convolutions are used.

\subsection{Training.}
The proposed system is trained on the dataset of scenes, unlike some prior approaches~\cite{nerf,free_synthesis,deferred_rendering,nex} that require dedicated training per scene.
The training process of the presented model is typical for NVS pipelines.
First, we sample $V$ source views and use them to construct an MLI representation, as explained above.
Then we render MLI on holdout cameras sampled from the same scene and compare the generated images $\{I_j^n\}_{j=1}^N$ with the ground truth images $\{I_j^{gt}\}_{j=1}^N$ using the pixelwise $\ell_1$ loss \scalebox{0.85}{$L_1 = \frac{1}{3NHW} \sum_{j=1}^N \|I^n_j - I^{gt}_j\|_1$} as well as perceptual loss~\cite{johnson2016perceptual} \scalebox{0.85}{$
    L_{\text{perc}} = \frac{1}{N} \sum_{j=1}^N \sum_t w_t \|VGG_t \left( I^n_j \right) - VGG_t \left( I^{gt}_j \right)\|_1, 
$} where the weights $w_t$ correspond to the different layers of the VGG network~\cite{vgg}.
To smooth the geometry of the deformable layers, the total variation loss $L_{\text{tv}}$ is imposed on the depth of each layer (the loss is computed for each of the $L$ maps independently).
Overall, our loss function is equal to   
$
    \lambda_1 L_1 
    + \lambda_{\text{perc}} L_{\text{perc}} 
    + \lambda_{\text{tv}}L_{\text{tv}},
$
and by default $\lambda_1 = 1$,  $\lambda_{\text{perc}} = 2$, $\lambda_{\text{tv}} = 0.1$.
We use Adam optimizer~\cite{adam} with an initial learning rate equal to $10^{-4}$ and cosine annealing schedule~\cite{cosine_schedule}.
More details can be found in our released code.

\begin{figure*}[t]
    \centering
        \includegraphics[width=\linewidth]{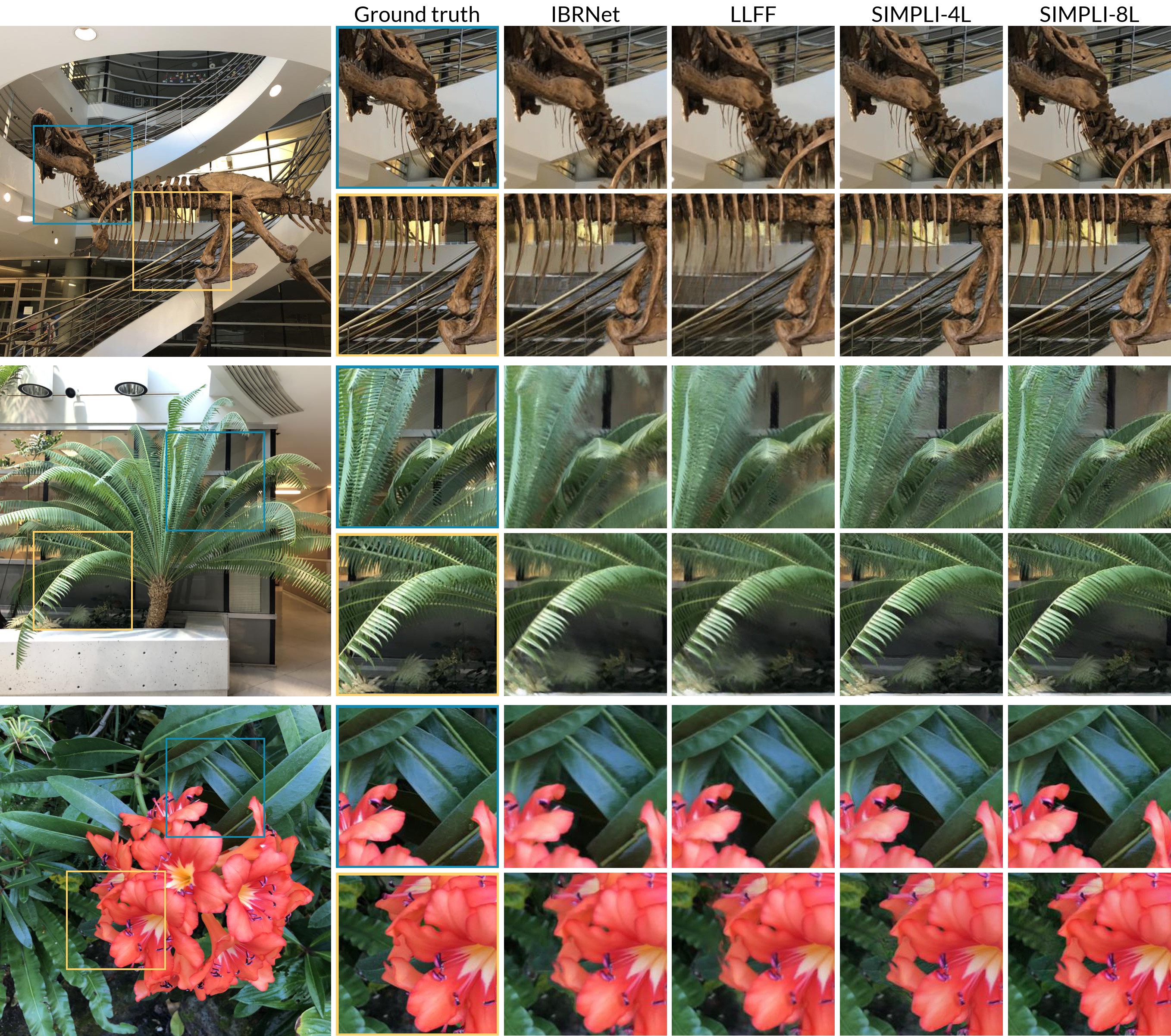}
    \caption{Comparison on a scene from the hold-out RFF dataset. In this experiment 8 source views were provided to all the models.
    Outputs of IBRNet and LLFF models are more blurry than results produced by \modelname (ours).
    As expected, using more layers in the MLI representation leads to better performance: note the bones of a T-Rex. 
    }
\label{fig:crop_ours}
\end{figure*}

\section{Experiments}
\label{sec:experiments}

\subsection{Baselines, datasets and metrics}
\topic{Baselines.}
To measure the performance of our system, we compare it with four baseline models: Stereo magnification (StereoMag)~\cite{stereomag}, LLFF~\cite{llff},  DeepView~\cite{deep_view} and the more recent IBRNet~\cite{ibrnet}.
StereoMag was designed for the  extreme case of only two input views, so we trained this system on our data in this setting. 
The authors of LLFF have not open-sourced the training code for their network, therefore for the evaluation we use the inference code and checkpoint of this model provided by its authors.

The source code and checkpoints for the DeepView approach, which is most close in spirit to ours, were not released, while the authors admit the model is hard to implement~\cite{deep_view}.
However, they have shared the results of their model trained on the Spaces dataset~\cite{deep_view}.
Therefore, we compare our approach with theirs on this data.
To make the comparison more fair, we train a special modification of our \modelname model called DeepView$^\dagger$ (marked with dagger), which does not convert planes to layers and instead performs several error correction steps for the initial 40 planes.

Although the recently presented NeX model~\cite{nex} outperforms LLFF and DeepView by a significant margin, we do not consider this approach as our baseline since it requires the full training procedure for each new scene which is far from our setting.

The authors of IBRNet evaluated its quality with and without fine-tuning for new scenes and concluded that their model shows the best results after additional per-scene optimization. However, for a fair comparison among methods, we do not fine-tune the IBRNet network and, therefore, measure the ability to generalize for all considered systems.
We have tried to fit IBRNet on our training data, but obtained significantly worse results than the released pre-trained model: PSNR on our holdout dataset equals 17.61 vs 22.69 for the released checkpoint.
We suggest that this discrepancy is mostly due to the different sampling strategy of training camera poses, as well as more complex structure of our training dataset.
Therefore, for the evaluation purposes, we stick to the pre-trained weights provided by the authors of IBRNet.

\topic{Training set.}
For training, we used 1,800 scenes from the training part of SWORD dataset~\cite{stereo_layers}.
The pipeline of data preparation was similar to that described in~\cite{stereomag}.
The source and novel poses are randomly sampled from the continuous range of frames within a scene, and the number of source images $V$ varied from 2 to 8.
To compare with DeepView, we also trained our models on the Spaces set (91 scenes).

\topic{Validation set.}
To validate various configurations of our model (see \cref{subsec:ablation_study} for the list), we measured their quality on Real Forward-Facing data, referred to as RFF (40 scenes)~\cite{llff} and a subsample (18 scenes) of data released with IBRNet paper, as well as Shiny dataset (8 scenes)~\cite{nex}.

\topic{Test set.}
Since IBRNet was trained on a mixture of datasets, including part of the RFF data and RealEstate10K~\cite{stereomag}, for the test set we selected 8 RFF scenes that IBRNet did not see and 8 scenes from the Shiny dataset~\cite{nex}. We repeat the evaluation 20 times to compute the standard deviation.
The resolution of test images equals $384 \times 512$. 
Additionally, we report the results of different methods on 38 scenes from the test part of SWORD.%

\topic{Sampling test views.}
We follow the evaluation protocol of DeepView, but extend it from photos taken from a camera rig to frames sampled from a short monocular video.
Initially, a range of subsequent frames is chosen from the scene video, and afterward, source and novel cameras are sampled from this range without replacement. 

In contrast, the authors of IBRNet selected the source camera poses for each novel pose as the nearest ones from the whole video to estimate the radiance field.
Therefore, the scenario of DeepView looks more realistic for real-life applications, although this setup was not investigated for other baseline methods we have chosen.
In addition, this setting also evaluates the robustness of different methods since it involves both the interpolation and extrapolation regimes.

\topic{Metrics.}
For evaluation, we employ several common metrics: perceptual similarity (LPIPS)~\cite{lpips}, peak signal-to-noise ratio (PSNR), and structural similarity (SSIM). 
Before computing the metrics, we take central crops from the predicted and ground-truth images, preserving 90\% of the image area, since the methods, based on MPI and MLI geometry, cannot fill the area outside the reference camera frustum~\cite{synsin}.
To measure human preference, we showed pairs of videos to users of a crowdsourcing platform.  Within each pair, a virtual camera was following the same predefined trajectory, and the assessors were to answer the question of which of the candidates looked more realistic (method known as 2AFC). 
For user study,
entire set of RFF data (40 scenes), 18 scenes from the data released in IBRNet paper~\cite{ibrnet},
and the Shiny dataset (8 scenes) were used. 
For each scene we generated two trajectories: rotation around the scene center (as in the IBRNet demo) and a spiral-like path with camera moving forward and backward.
800 different workers participated in the study, each pair was assessed by 40 of them.

\begin{table}[t]
\resizebox{\columnwidth}{!}{
\begin{tabular}{lccccccc}
\toprule
\multirowcell{2}[0pt][l]{Configuration} & \multicolumn{3}{c}{5 source views} && \multicolumn{3}{c}{8 source views} \\%
\cmidrule{2-4}\cmidrule{6-8}
& PSNR $\uparrow$ & SSIM$\uparrow$ &  LPIPS$\downarrow$ && PSNR $\uparrow$& SSIM$\uparrow$ &  LPIPS$\downarrow$ \\%
\midrule
w/o perceptual loss   & \bf{24.16}$_{3.06}$ & \bf{0.79}$_{0.09}$ & 0.13$_{0.06}$    && \bf{24.58}$_{2.98}$ & \bf{0.80}$_{0.08}$ & 0.13$_{0.05}$    \\%
w/o attention pooling  & 23.67$_{3.11}$ & 0.77$_{0.10}$ & 0.12$_{0.05}$      && 24.09$_{3.04}$ & 0.79$_{0.09}$ & 0.11$_{0.05}$     \\%
w/o error correction  & 21.32$_{4.15}$ & 0.69$_{0.12}$ & 0.16$_{0.05}$      && 22.13$_{3.12}$ & 0.70$_{0.11}$ & 0.14$_{0.05}$     \\%
$\tau = 0$ (\fourincircle omitted) & 23.15$_{4.01}$ & 0.76$_{0.11}$ & 0.12$_{0.05}$     && 23.92$_{3.27}$ & 0.78$_{0.11}$ & 0.11$_{0.05}$     \\%
$\tau = 1$ (less steps) & 23.73$_{3.16}$ & 0.78$_{0.10}$ & \bf{0.11}$_{0.05}$     && 24.16$_{3.08}$ & 0.79$_{0.09}$ & 0.11$_{0.04}$     \\%
$\tau = 2$ (less steps) & 23.90$_{3.13}$ & 0.79$_{0.09}$ & \bf{0.11}$_{0.05}$     && 24.30$_{3.04}$ & \bf{0.80}$_{0.08}$ & 0.11$_{0.04}$     \\%
$L=2$  (fewer layers) & 23.41$_{3.20}$ & 0.77$_{0.10}$ & 0.12$_{0.06}$    && 23.76$_{3.13}$ & 0.78$_{0.09}$ & 0.12$_{0.05}$    \\%
$L=8$  (more layers)  & 23.96$_{3.10}$ & \bf{0.79}$_{0.09}$ & \bf{0.11}$_{0.05}$    && 24.39$_{3.00}$ & \bf{0.80}$_{0.08}$ & \bf{0.10}$_{0.04}$     \\%
\midrule
default model & 23.79$_{3.14}$ & 0.78$_{0.10}$ & \bf{0.11}$_{0.05}$     && 24.21$_{3.05}$ & 0.79$_{0.09}$ & 0.11$_{0.05}$     \\%
\bottomrule
\end{tabular}
}
\centering
\vspace{0.7ex}
\caption{Ablation study.
    The default model contains $P=40$ planes at intermediate steps and $L=4$ layers in the final MLI representation.
    $\tau$ is the number of correction steps at stage \fourincircle, by default $\tau = 3$. 
    Note that a model without correction steps both at stages \twoincircle and \fourincircle demonstrates the worst performance.
    Despite removal of perceptual loss improves PSNR and SSIM, the results get blurry which leads to worse LPIPS value.
    Increasing the number of layers predictably raises the quality of results.
    Subscripts indicate the standard deviation.
}
\label{tab:ablation}
\vspace{10pt}
\centering
\resizebox{0.8\linewidth}{!}{\begin{tabular}{lccl}
\toprule
\multirowcell{2}[0pt][l]{Model} & \multirowcell{2}{\# params,\\ mln} &  \multirowcell{2}{Building\\ representation, sec} &  \multirowcell{2}{Rendering\\  speed, fps} \\%
&&& \\%
\midrule
IBRNet          & 9.0  &  --  &  $\sim 0.4$ \\
LLFF        & 0.7 &  31.3  &  $\sim 60$  \\
DeepView$^\dagger$ & 1.7 &  65.4  &  $\sim 80$ \\
\modelname-4L    & 1.9 &  9.6 &  $\sim 200$ \\
\modelname-8L & 2.0 &  10.6  &  $\sim 120$ \\
\bottomrule
\end{tabular}
}
\vspace{0.7ex}
\caption{Rendering speed for 8 source views at resolution of $768 \times 1024$.
For all methods except IBRNet, we measure the time of representation building and rendering separately, as IBRNet requires forward pass of a neural network for each new frame, while other models may be used with graphics engines. DeepView$^\dagger$ was implemented by us.
Measurements are provided for NVIDIA P40 GPU.
}
\label{tab:speed}
\end{table}

\begin{figure}[t]
\includegraphics[width=\columnwidth]{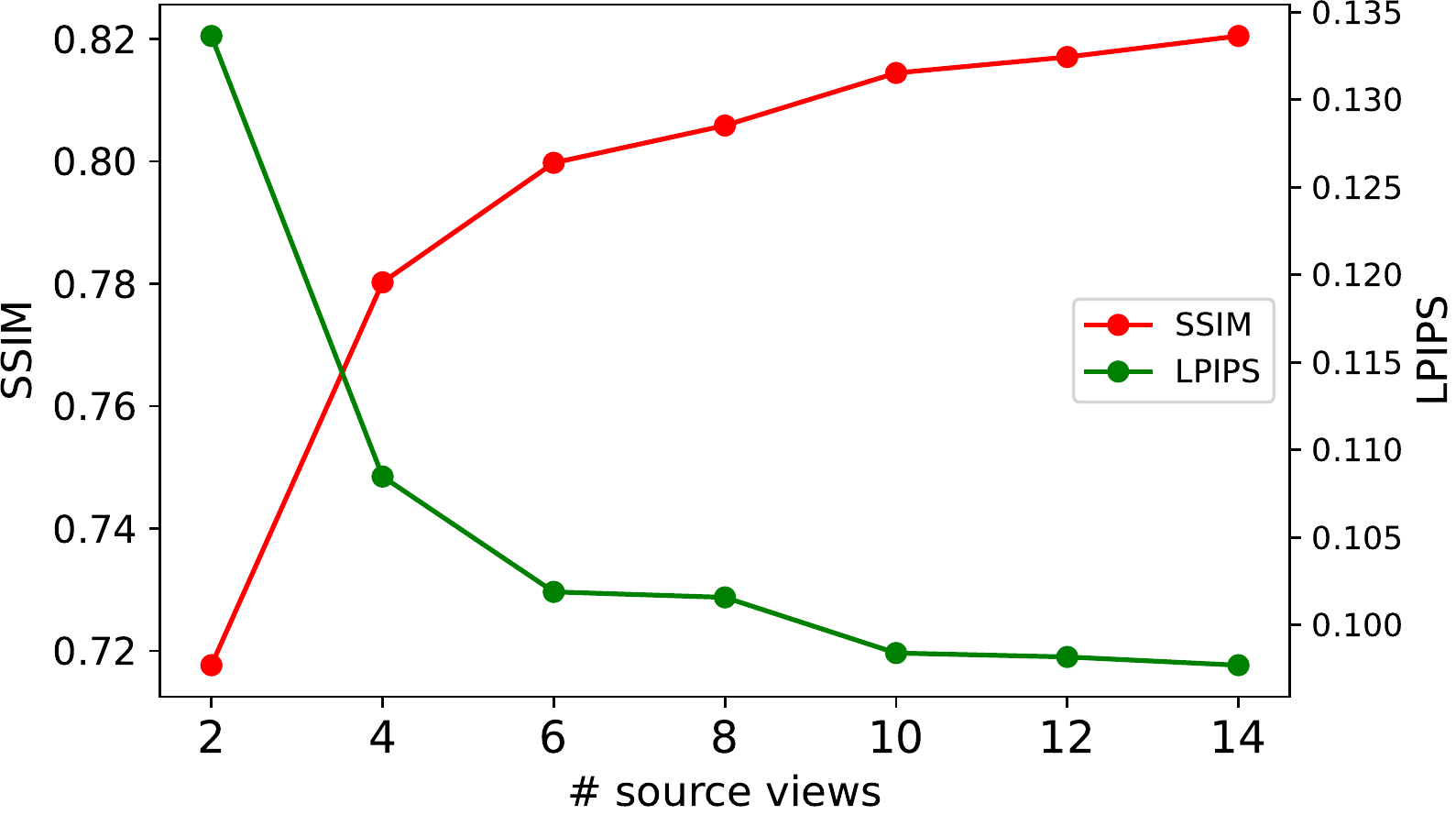}
\caption{Increasing the number of source views results in better performance of the proposed \modelname model.}
\label{fig:nviews_vs_quality}
\end{figure}

\subsection{Model configurations}
\label{subsec:ablation_study}
The number of planes $P$ in the preliminary MPI representation equals 40, and the default number of layers is $L=4$, unless other values are explicitly specified,
\ie \modelname-8L denotes the architecture with $L=8$ deformable layers.
The modification called DeepView$^\dagger$ does not convert planes to layers (\ie step \threeincircle is omitted), and step \fourincircle is performed for MPI as well as \twoincircle.

We conducted an ablation study to evaluate the significance of different parts of our \modelname system and choose the best configuration.
For this purpose, we trained versions of our model on two NVIDIA P40 GPUs for 90,000 iterations.
\cref{tab:ablation} shows that an increase in the number of correcting steps can slightly improve quality.
As expected, perceptual loss is important for the quality of high-frequency details measured by LPIPS.
Using less layers worsens the model performance as well, while adding more layers or correction steps boosts the quality.
This may be caused by the employed algorithm of converting planes to layers. We are going to explore alternative schemes to improve the performance of the \modelname model in future.
Additionally, \cref{fig:nviews_vs_quality} demonstrates that a greater number of source views fed to the model results in better quality.

\subsection{Main results}
To compare our approach with baselines, we train \modelname on 8 P40 GPUs for 500,000 iterations (approximately 5 days) with an effective batch size of 8. 
The main results are reported in \cref{tab:main_scores} and \cref{tab:spaces_scores}.
\modelname outperforms LLFF and typically is better than IBRNet, although in some cases obtaining slightly worse PSNR.
This result may come from the fact that IBRNet was trained with $\ell_2$ loss only, corresponding to PSNR (\cf \cref{tab:ablation}), while our model used perceptual loss.
This is a clear advantage of our approach that it allows the training on patches large enough to use VGG-based losses, while IBRNet is trained on sparsely sampled pixels due to higher memory consumption. 
As reported in \cref{tab:speed},  \modelname allows for orders of magnitude faster rendering than IBRNet.
While rendering in real time is also possible for LLFF, the  quality it obtains is poor according to \cref{tab:main_scores}.

We observe that due to the large standard deviations, none of the models except DeepView outperforms \modelname by a statistically significant margin.
Comparison of \modelname with DeepView$^\dagger$ and original DeepView shows that there is a clear trade-off between quality and compactness on Spaces data, while on other datasets \modelname is consistently better.
At the same time, \modelname provides a much more compact representation than DeepView: 4 layers against 40 planes.

\cref{fig:crop_ours} provides a qualitative comparison of different approaches, demonstrating that the proposed model has fewer artifacts in most cases. 
Although the metrics do not show a significant difference between the models considered, the study of human preference demonstrates the decisive advantage of \modelname over recently proposed baselines: our model achieves \emph{81\%} compared to LLFF and \emph{79\%} against IBRNet.
See \switchArxiv{\cref{sec:additional_results}}{the supplementary text} and the accompanying video for more qualitative results.

\subsection{Limitations}
\label{sec:limitations}

\begin{figure}[t]
\includegraphics[width=\columnwidth]{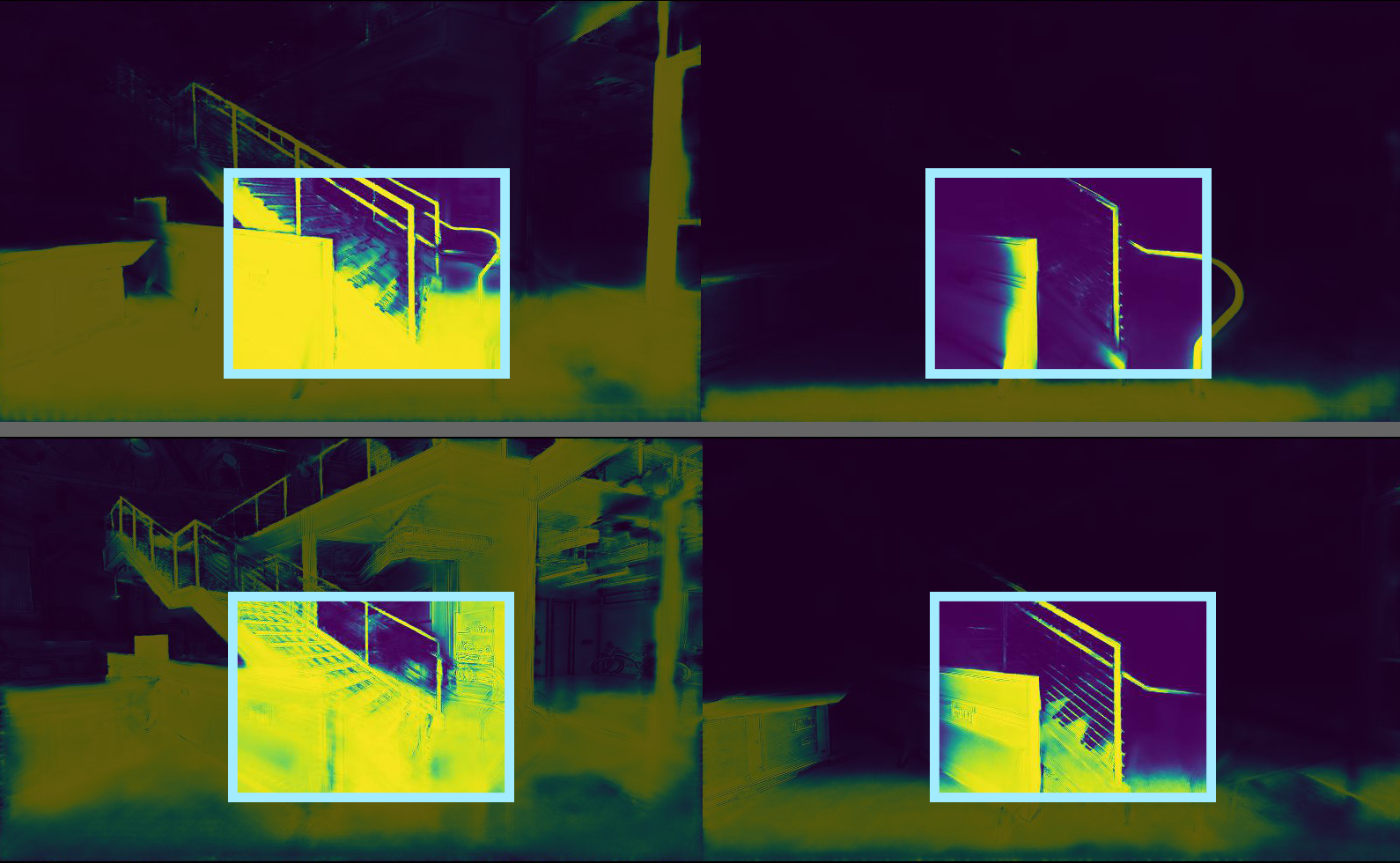}
\caption{
    Example of the limitations. 
    The opacity channel of predicted textures for the layers in the depth range corresponding to the ground truth position of the stair railing. 
    Left: model with 4 layers, right: model with 8 layers.
    Obviously, when using too few layers, the proposed \modelname is not capable of faithfully representing thin objects.
}
\label{fig:limitation_deepview}
\vspace{-10pt}
\end{figure}

Although our model provides good quality of results in most cases, it still struggles from some limitations.
First, similar to StereoMag, LLFF, StereoLayers and other methods which produce only the RGBA textures for the employed scene representation, \modelname cannot plausibly reconstruct view-dependent effects, \eg specular reflections.
One possible way to address this drawback is to predict the spherical harmonic coefficients instead of RGB color (\cf NeX~\cite{nex}), and we leave this for future work.
Second, as shown in \cref{fig:limitation_deepview} for complicated scenes with fine-grained details, using as few as four layers is not enough.
It is not immediately clear whether this effect should only be attributed to the small internal capacity of the representation with a few layers.
Instead, the suboptimal procedure for converting planes to layers (step \threeincircle) may be to blame.
Therefore, searching for a more appropriate merging algorithm, as well as extending the error correction step to the depth maps, are other ways to improve our model.

\section{Conclusion}
\label{sec:conclusion}
We presented a method of novel view synthesis called \modelname.
It employs a mesh-based scene representation that consists of a set of non-intersecting semitransparent  layers. 
RGBA textures for the mesh layers are inferred with a multistage neural pipeline that refines the representation.
Our network is trained on a dataset of real-life scenes and generalizes well to the unseen data without fine-tuning, allowing for on-device usage.

Our approach extends existing multiview methods in three major ways: (i) we propose to use an adaptive and more compact representation of the scene (4 or 8 layers instead of 40 planes), (ii) we use input frames from the natural handheld video instead of the calibrated rig, and (iii) the presented system does not rely on the predefined number of input views, instead being able to work with an arbitrary number of source images.

Evaluation shows that \modelname outperforms recent state-of-the-art methods or produces results of similar quality, while excelling over them in the speed of rendering or the size of the obtained scene representation.

\section*{Acknowledgements}
\label{sec:acknowledgments}
The authors thank 
    V. Aliev, 
    A.-T. Ardelean, 
    A. Ashukha, 
    R. Fabbricatore,
    A. Kharlamov, 
    V. Lempitsky, 
    and R. Suvorov 
for their comments, which greatly improved the manuscript.

\FloatBarrier
\ifnum\value{page}>8 \errmessage{Number of pages exceeded!!!!}\fi
\clearpage

\renewcommand{\thetable}{S\arabic{table}}
\renewcommand{\thefigure}{S\arabic{figure}}
\setcounter{figure}{0}
\setcounter{table}{0}
\setcounter{equation}{0}

\appendix

\section{Additional details}
\label{sec:architecture_details}
\begin{figure*}[t]
\centering
\includegraphics[width=1.05\linewidth]{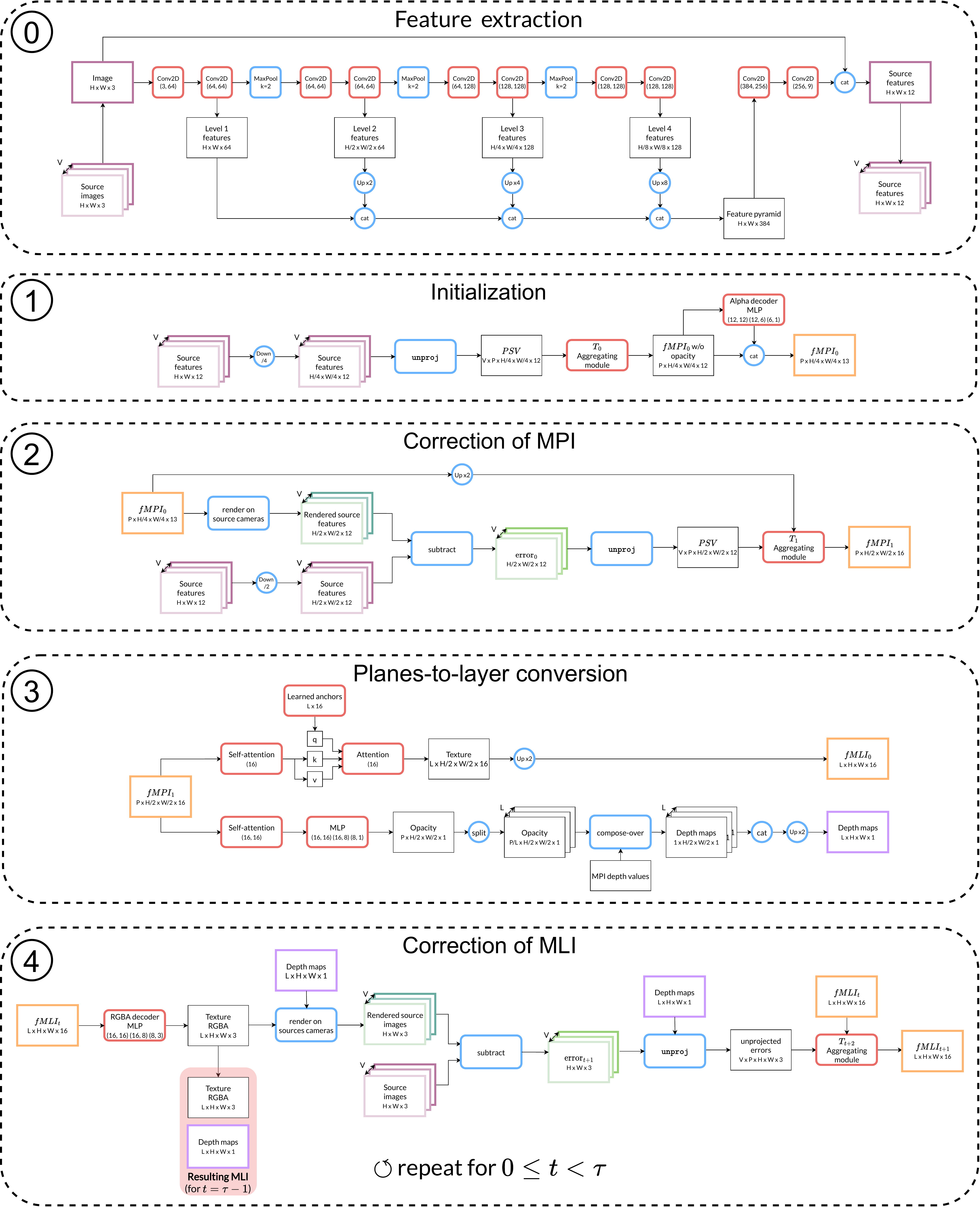}
\caption{%
     The detailed diagram of our pipeline.
     Please zoom in for details.
}
\label{fig:method_diagrams}
\end{figure*}

\begin{figure*}[t]
\centering
\includegraphics[width=\linewidth]{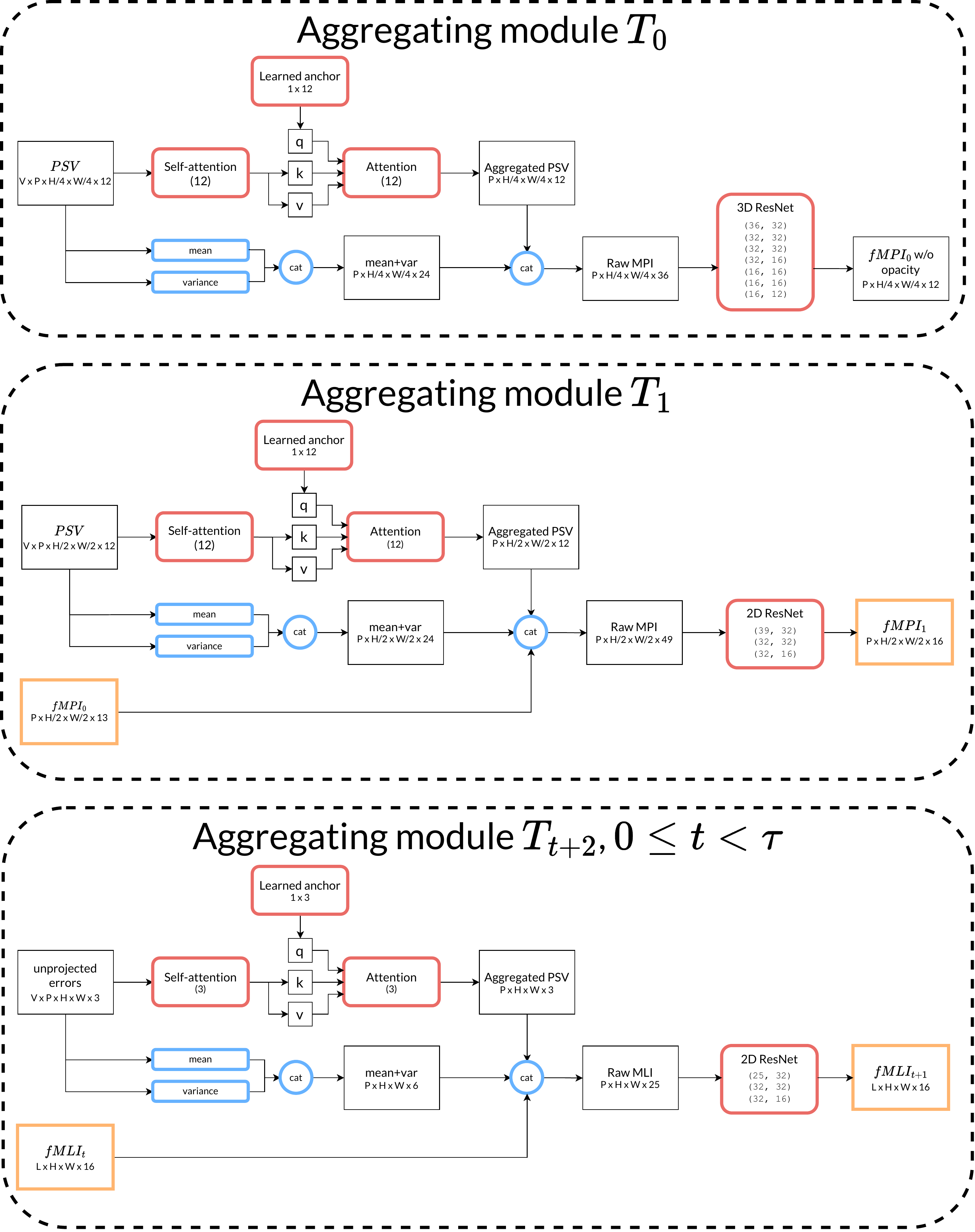}
\caption{%
     Architecture of aggregating modules.
     Please zoom in for details.
}
\label{fig:aggregating_modules}
\end{figure*}

\topic{Architectures.}
Detailed diagrams of the steps of our pipeline are shown in \cref{fig:method_diagrams}.
The architectures of the aggregating modules are described in~\cref{fig:aggregating_modules}.
\switchArxiv{}{The implementation of our system is released on the website  \url{https://samsunglabs.github.io/MLI}.}

\topic{MLI vs LDI.}
Since the seminal work on layered depth images (LDI)~\cite{ldi}, multiple papers considered equipping each pixel of the reference image with a stack of depth values.
However, the original definition of this representation~\cite{ldi} assumed that the size of this stack may differ between pixels.
In addition, any connections between pixels were not imposed.
Although later manuscripts introduced explicit local connectivity of neighboring pixels~\cite{3d_layered_inpainting}, we stick to another terminology and refer to our representation as a layered mesh~\cite{immersive_lf_video} or a multilayer image (MLI)~\cite{stereo_layers}.
The latter name is preferred, as it reveals the relation to multiplane images~\cite{stereomag}: just like them, our proxy geometry contains semitransparent RGBA textures.
On the contrary, many methods that have been reported to employ LDI representation do not use the opacity channel~\cite{Dhamo2019ldi,lsi,3d_layered_inpainting,one_shot_3d}.
Besides, MLI contains a predefined number of layers; therefore, each pixel gets the same number of depth values.
While the layers are non-overlapping by design, a ray from a novel camera can intersect each layer in several points, justifying the usage of z-buffer during the rasterization step.
In contrast, the original LDI representation did not need the z-buffer, and McMillan's warp ordering algorithm was used instead.

\section{Additional results}
\label{sec:additional_results}

\begin{figure*}[ht]
\setlength{\mrgone}{-0.4cm}
\setlength{\wid}{0.32\columnwidth}
\centering
\begin{tabular}{ccccccc}
\hspace{\mrgone} \includegraphics[width=\wid]{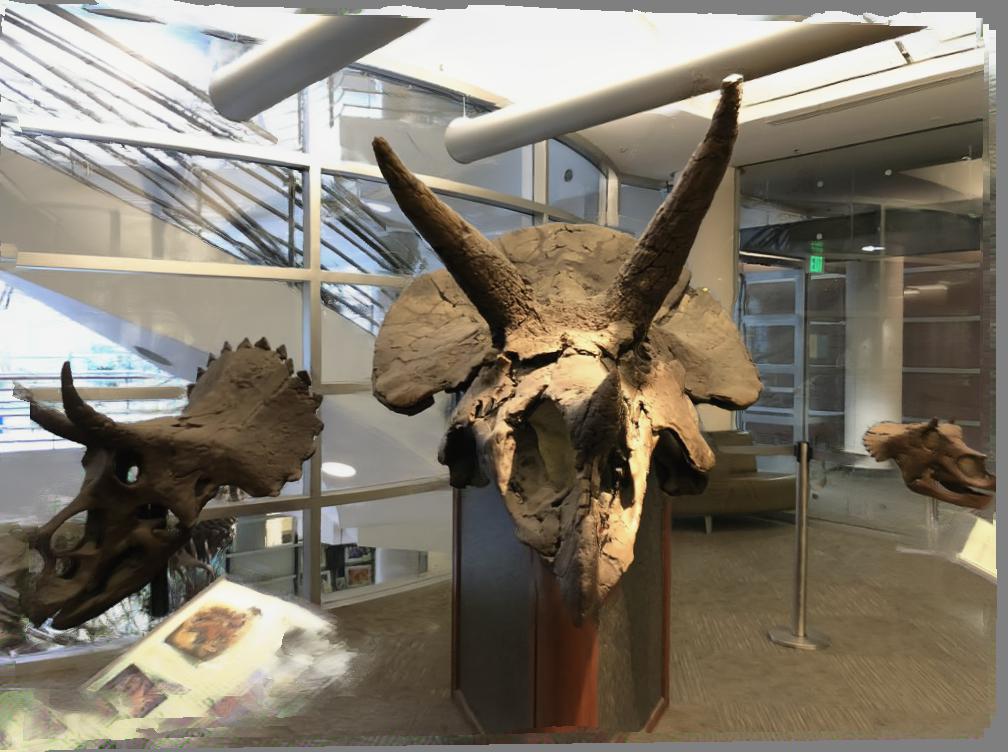} &  \hspace{\mrgone}
\hspace{\mrgone} \includegraphics[width=\wid]{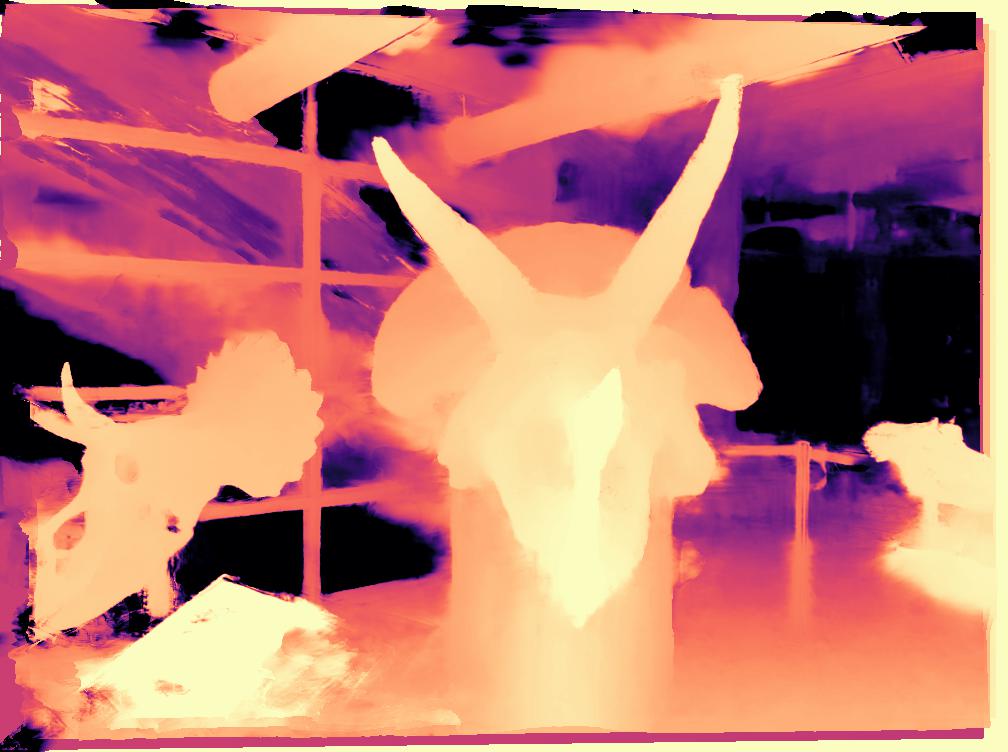} &  \hspace{\mrgone}
\includegraphics[width=\wid]{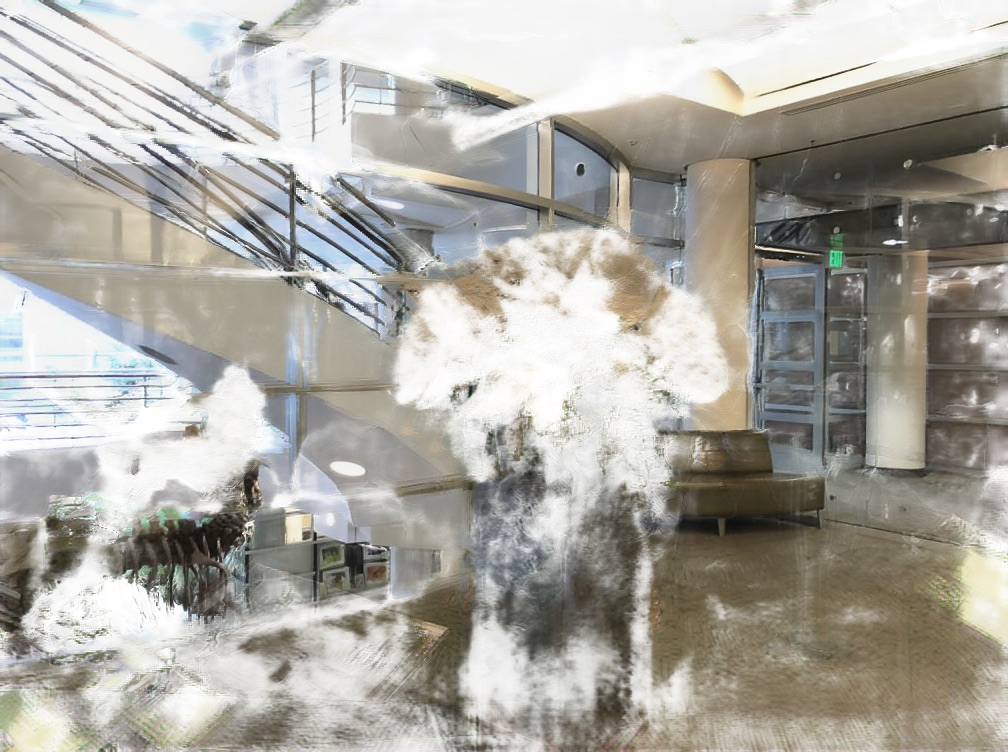} & \hspace{\mrgone}
\includegraphics[width=\wid]{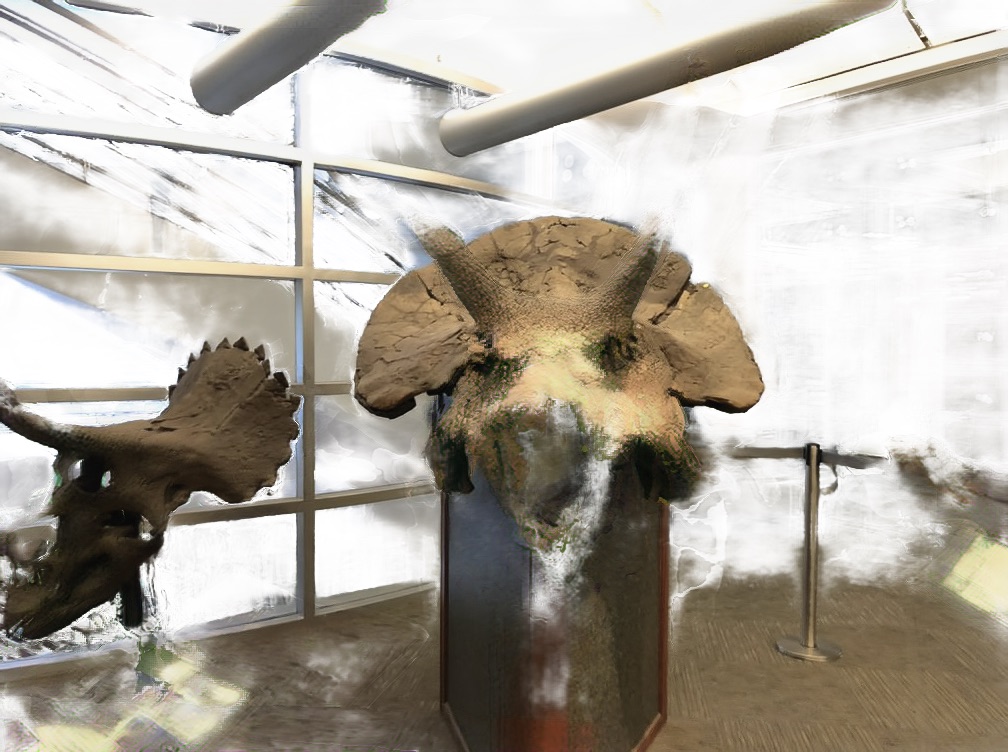} & \hspace{\mrgone}
\includegraphics[width=\wid]{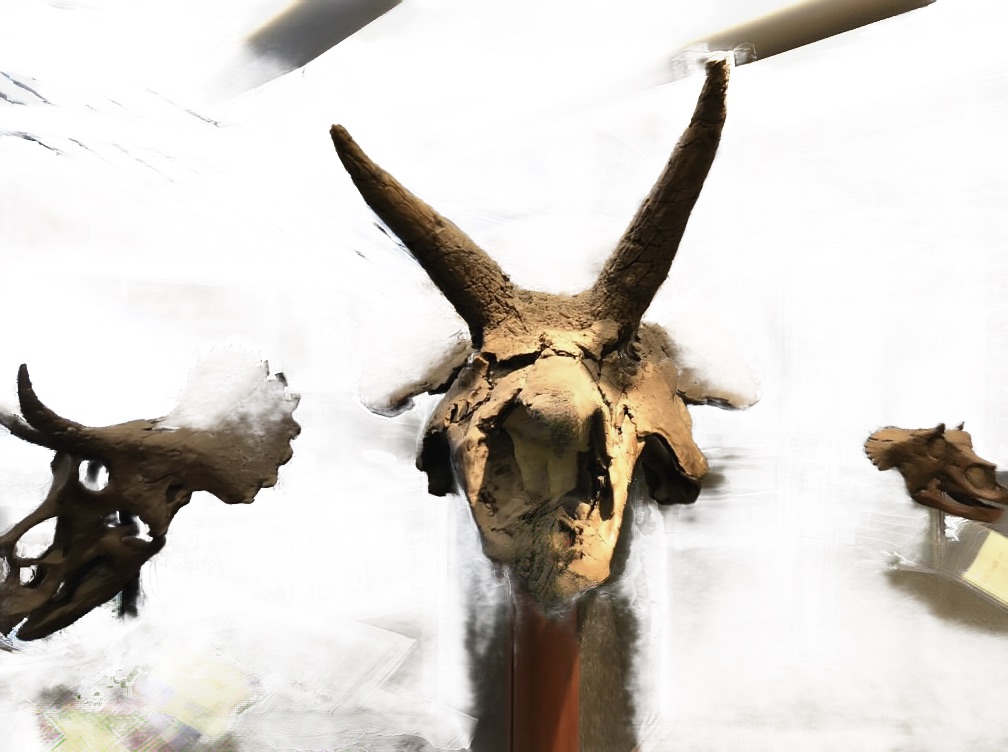} & \hspace{\mrgone}
\includegraphics[width=\wid]{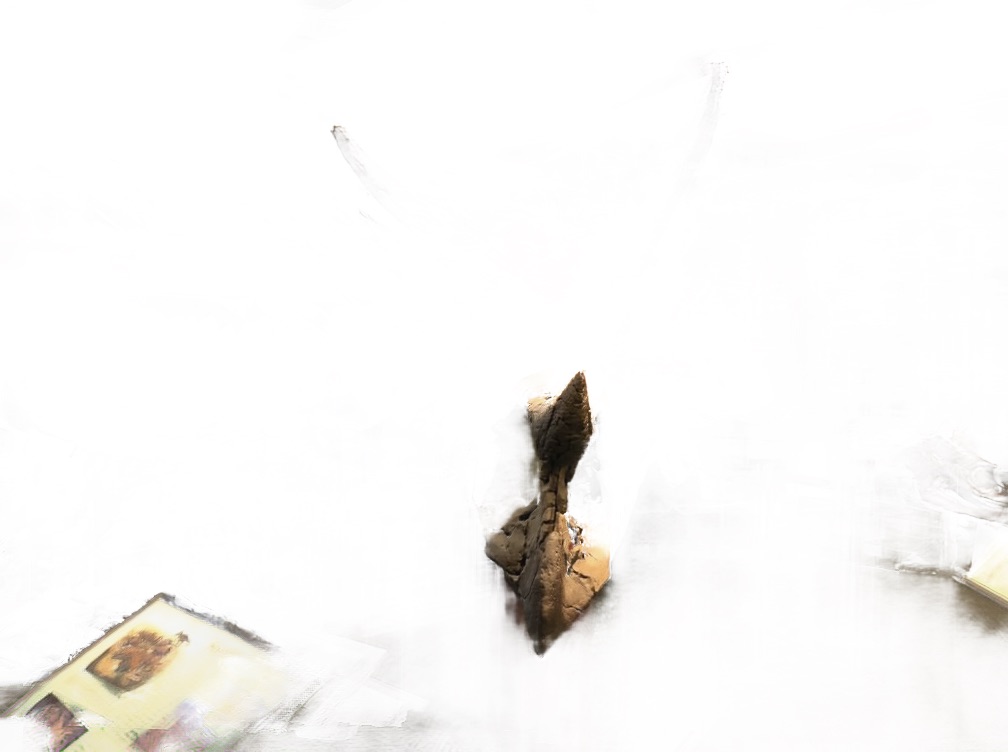} 
\\
\hspace{\mrgone} \includegraphics[width=\wid]{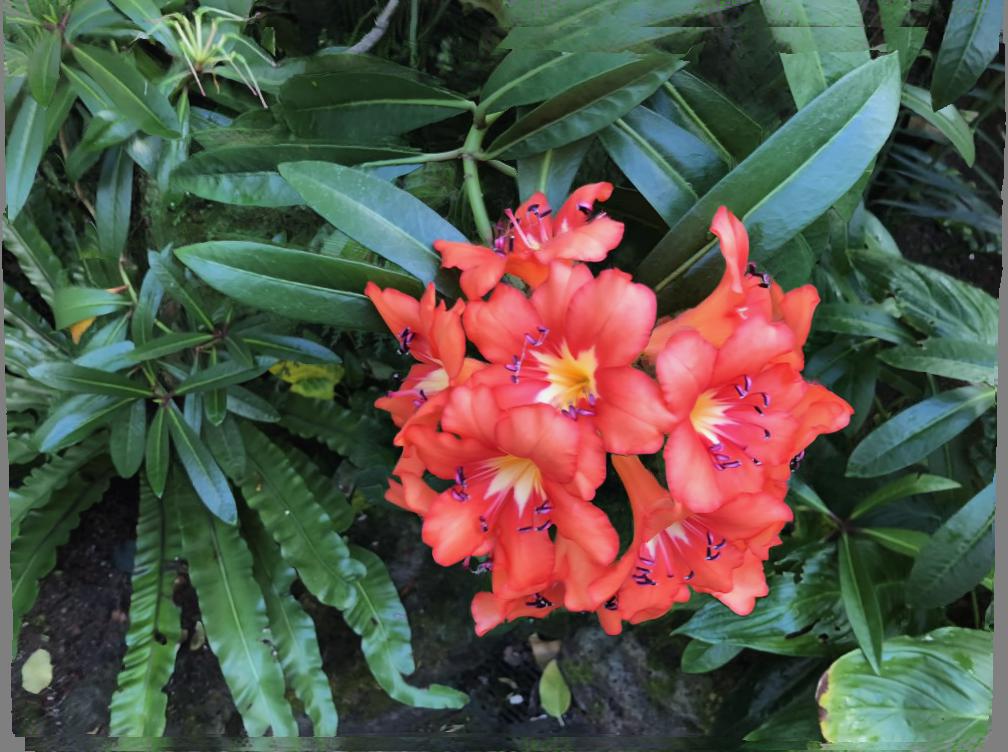} &  \hspace{\mrgone}
\hspace{\mrgone} \includegraphics[width=\wid]{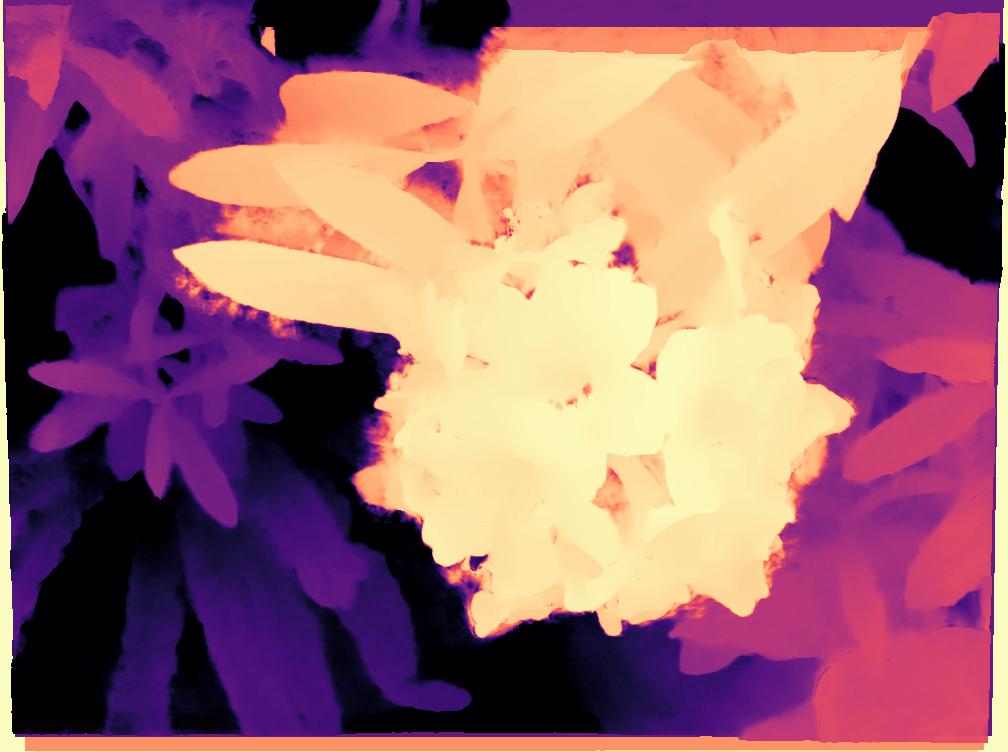} &  \hspace{\mrgone}
\includegraphics[width=\wid]{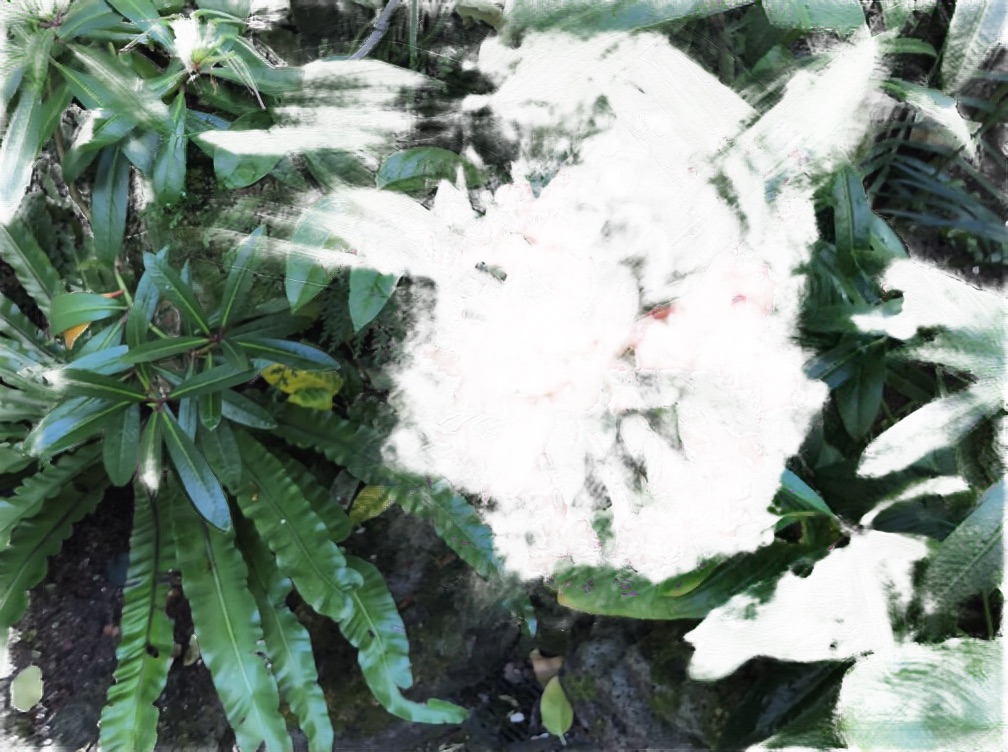} & \hspace{\mrgone}
\includegraphics[width=\wid]{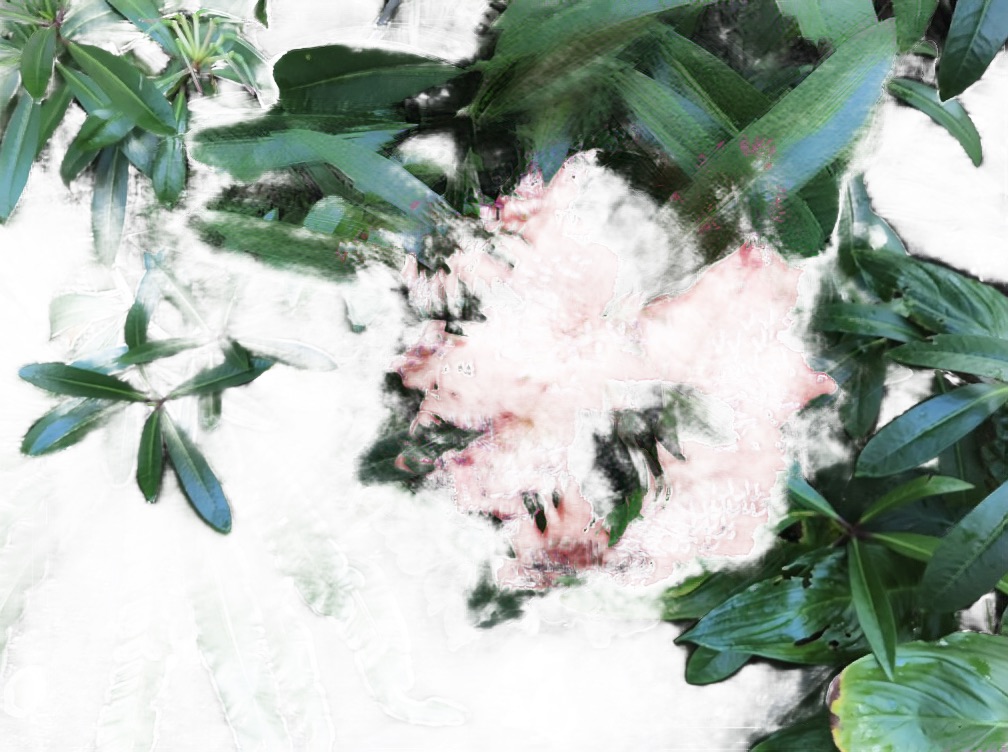} & \hspace{\mrgone}
\includegraphics[width=\wid]{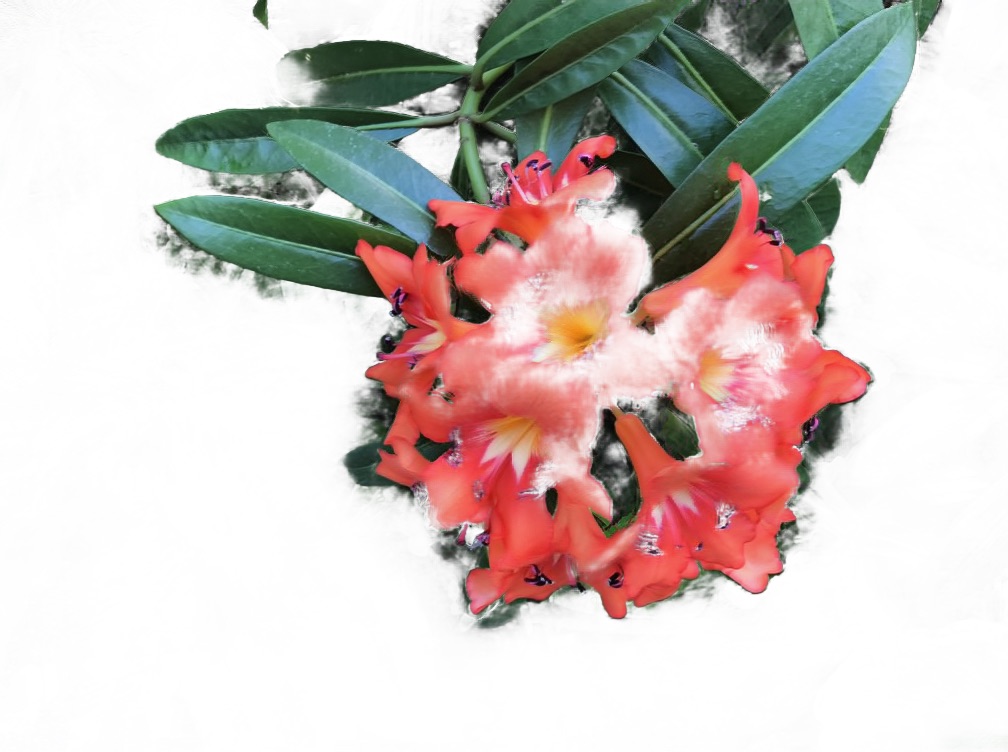} & \hspace{\mrgone}
\includegraphics[width=\wid]{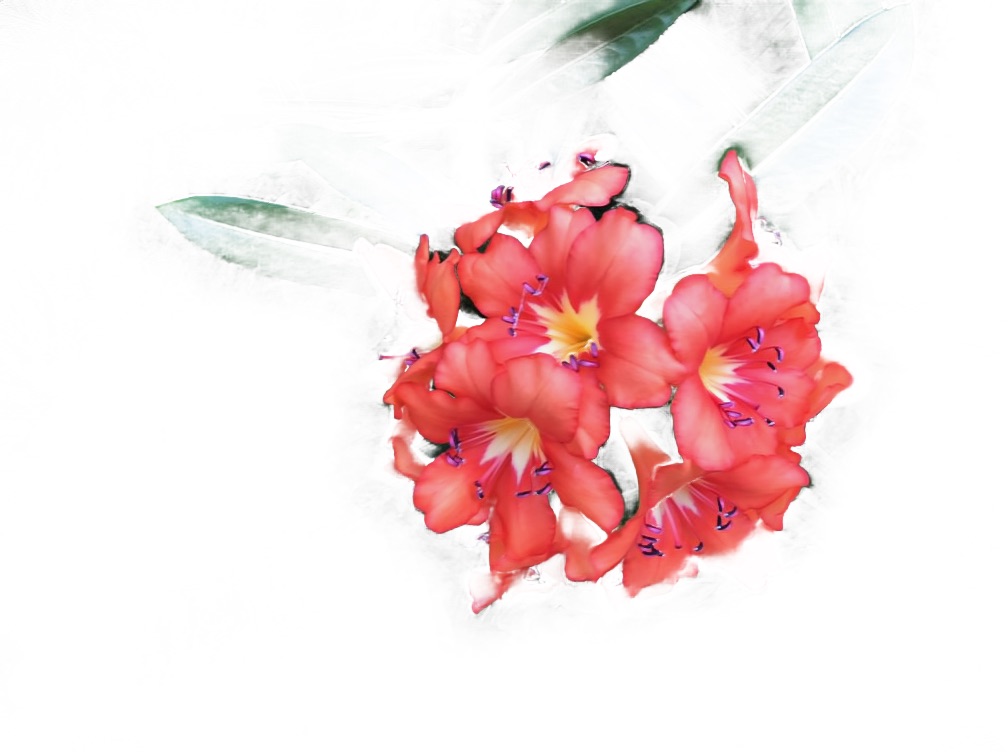} 
\\

\hspace{\mrgone} \includegraphics[width=\wid]{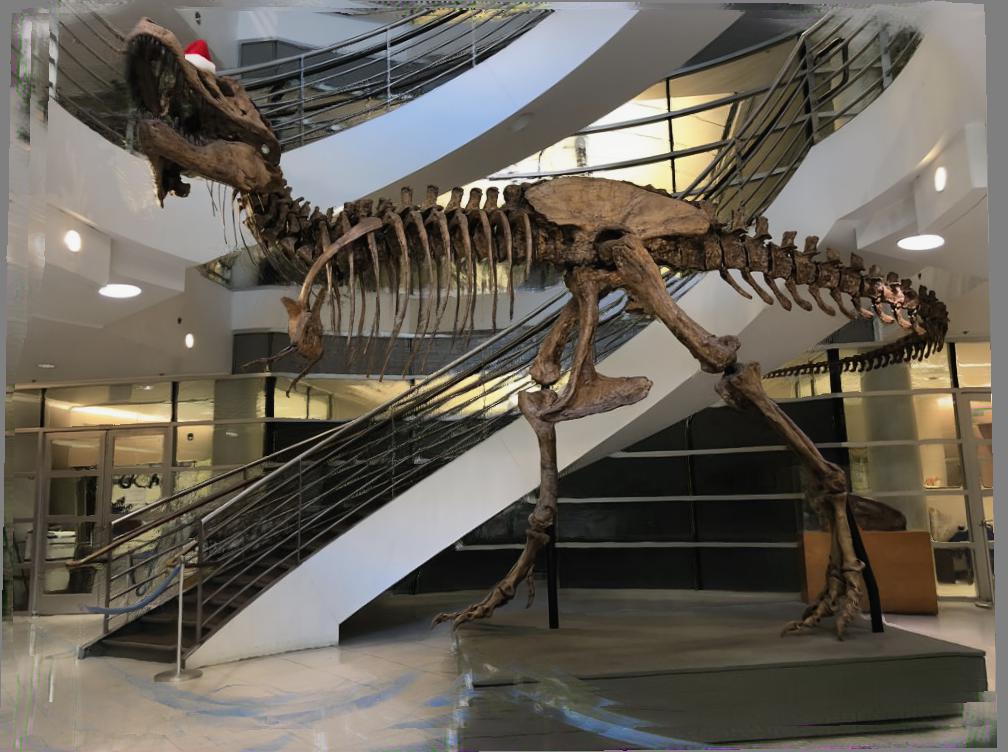} &  \hspace{\mrgone}
\hspace{\mrgone} \includegraphics[width=\wid]{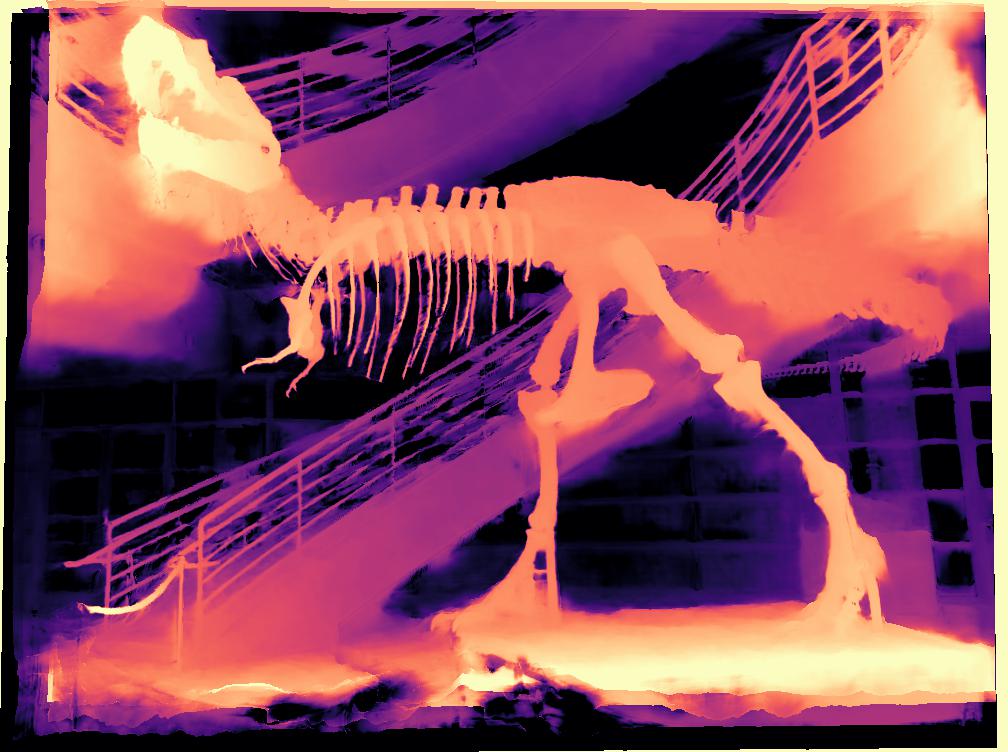} &  \hspace{\mrgone}
\includegraphics[width=\wid]{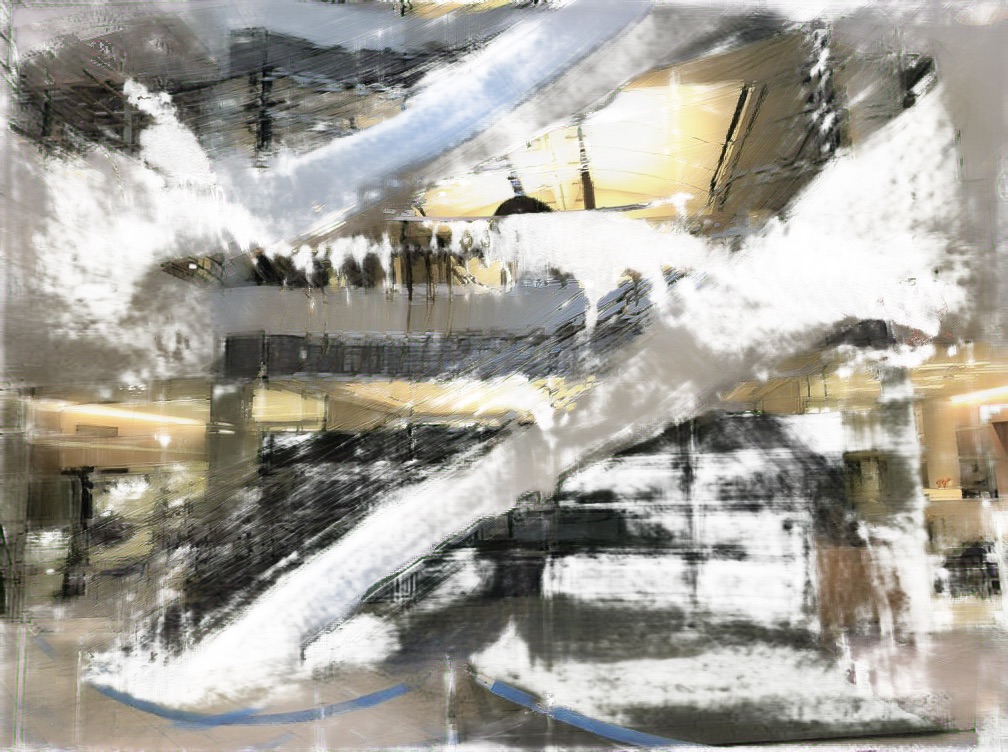} & \hspace{\mrgone}
\includegraphics[width=\wid]{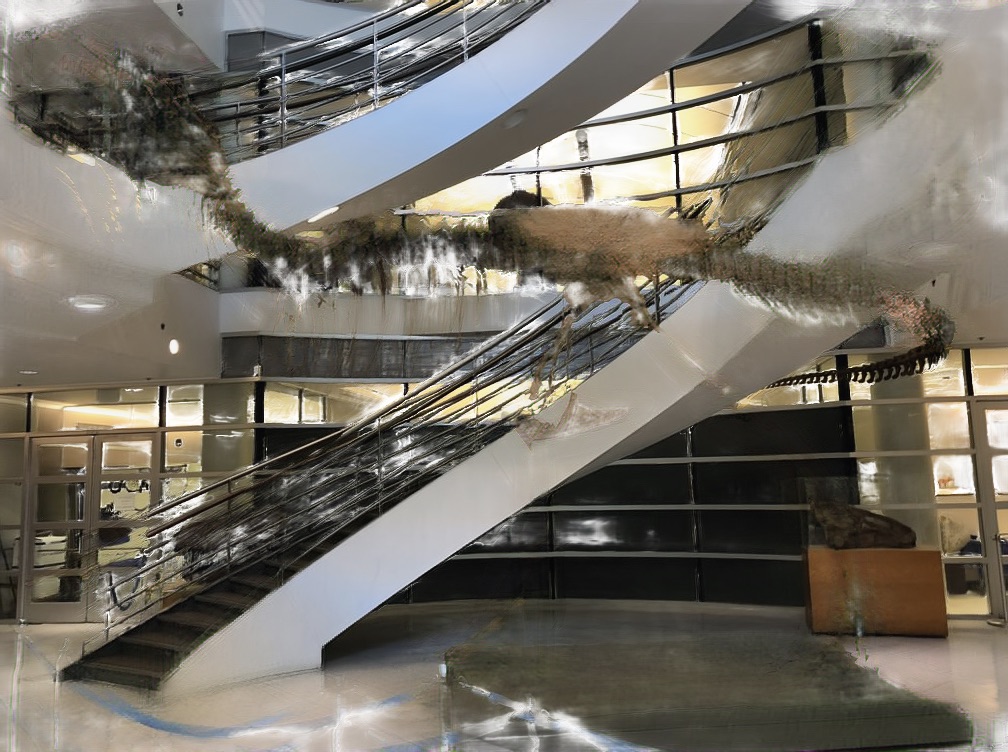} & \hspace{\mrgone}
\includegraphics[width=\wid]{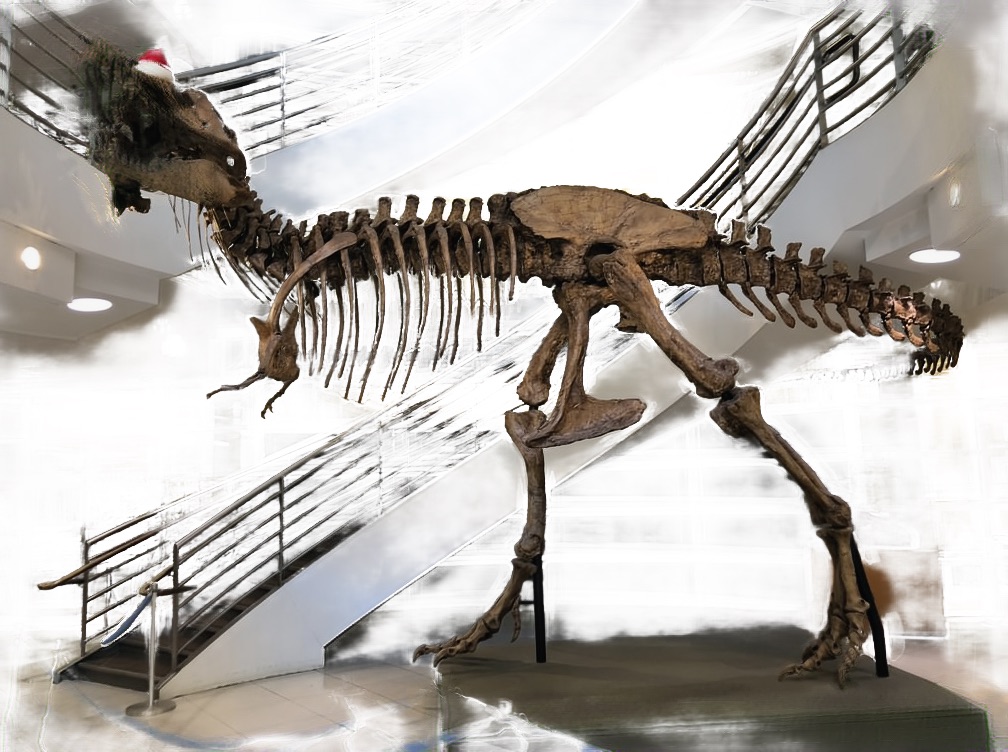} & \hspace{\mrgone}
\includegraphics[width=\wid]{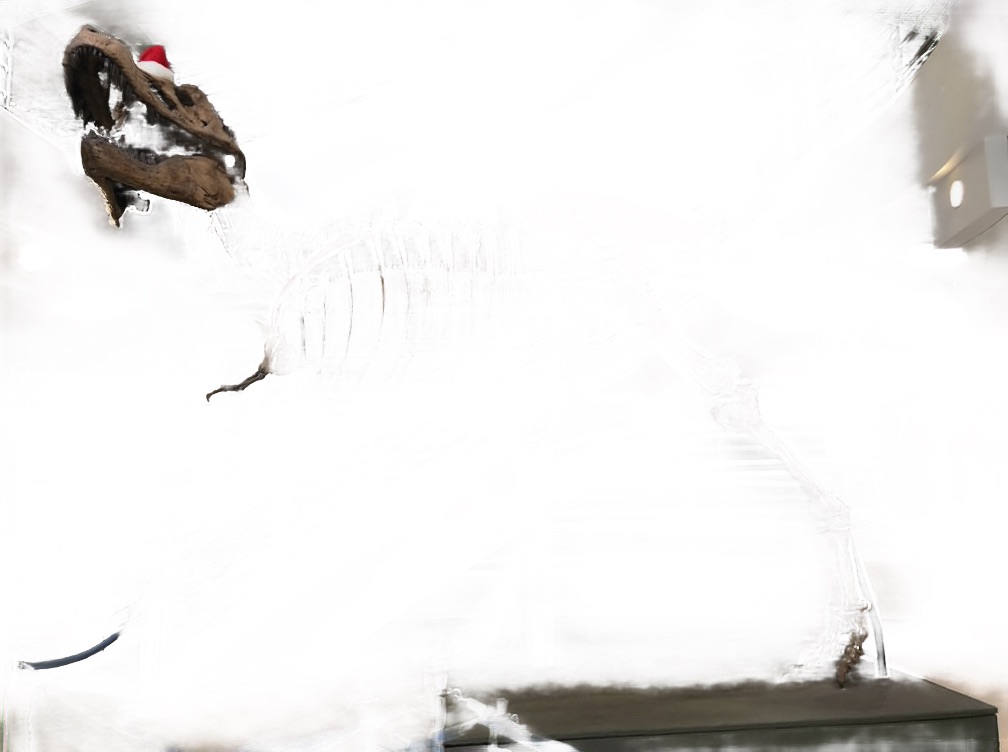}
\\

\hspace{\mrgone} \includegraphics[width=\wid]{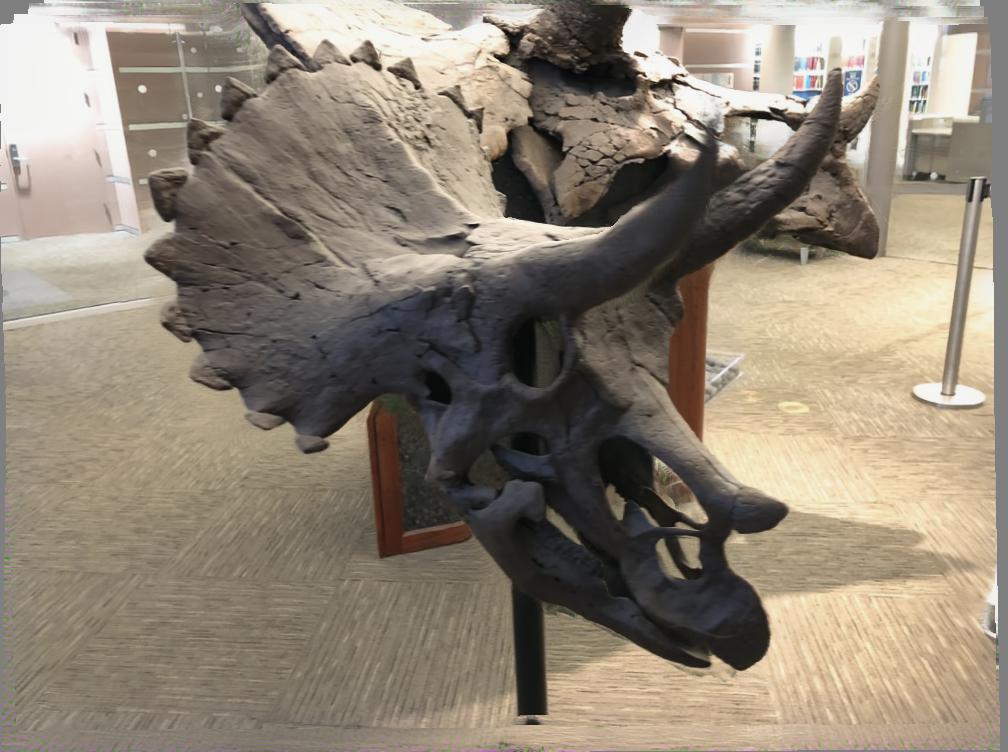} &  \hspace{\mrgone}
\hspace{\mrgone} \includegraphics[width=\wid]{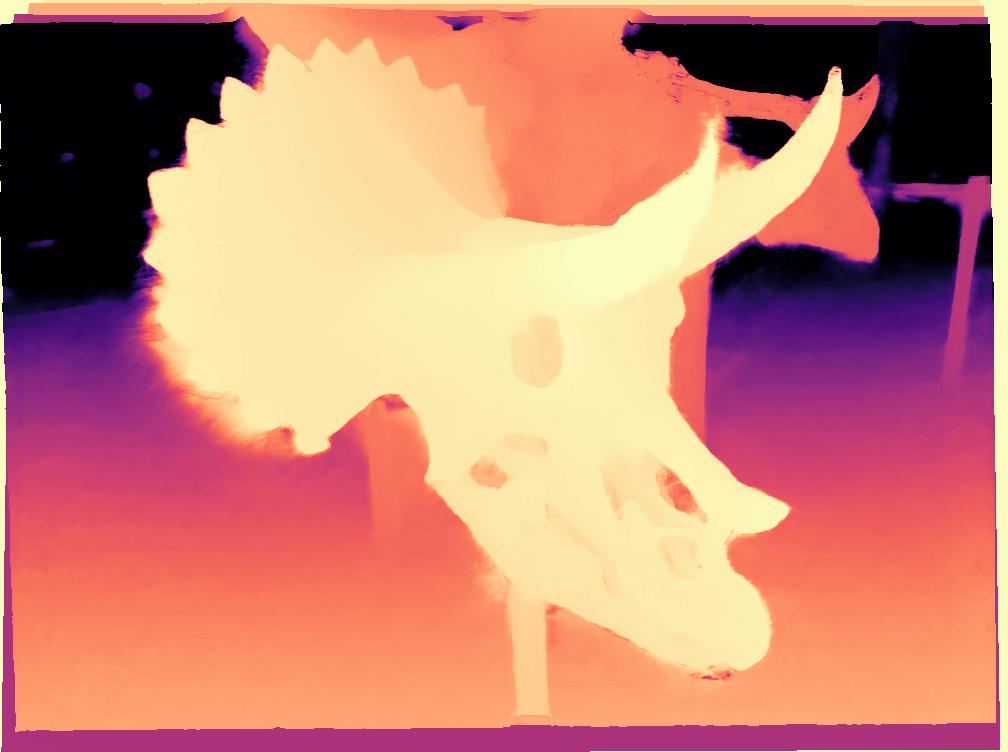} &  \hspace{\mrgone}
\includegraphics[width=\wid]{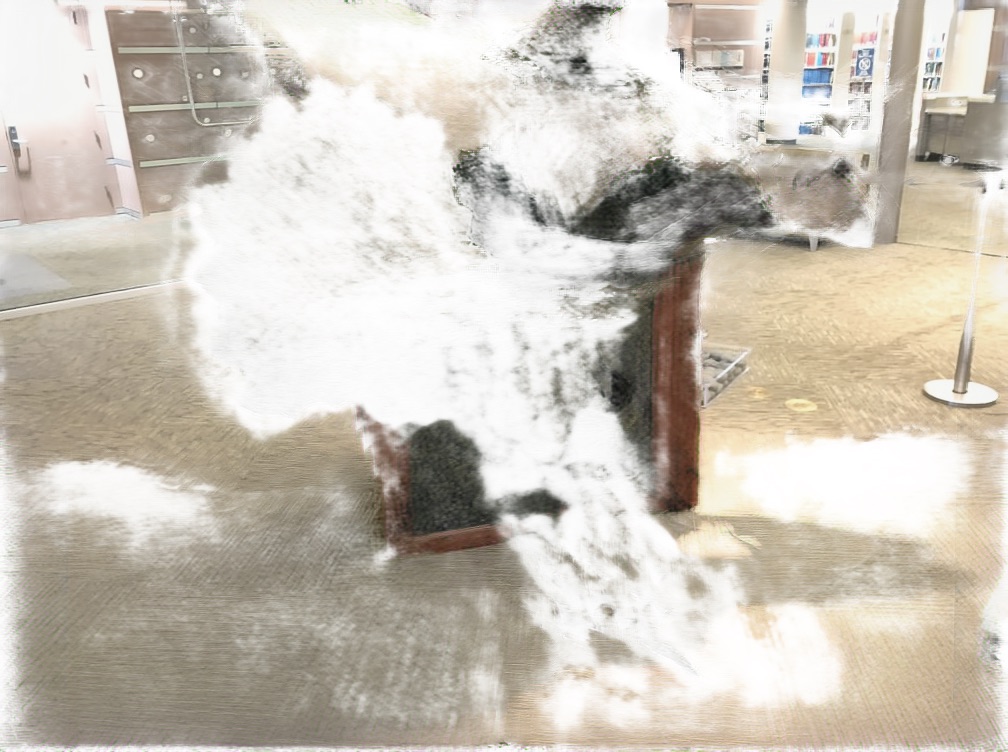} & \hspace{\mrgone}
\includegraphics[width=\wid]{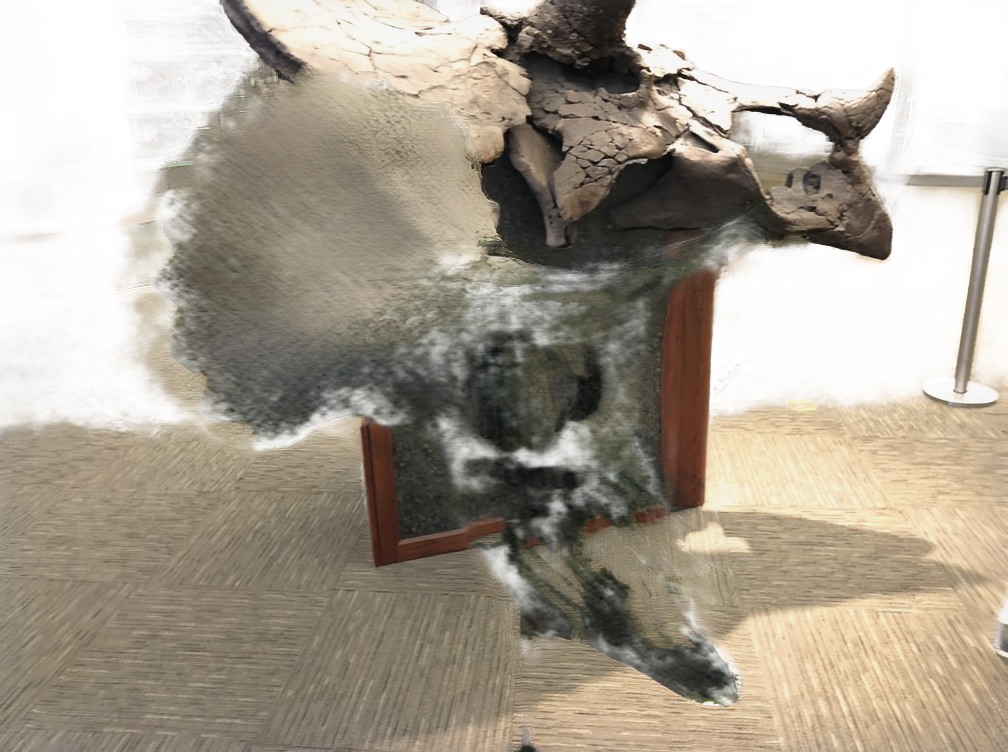} & \hspace{\mrgone}
\includegraphics[width=\wid]{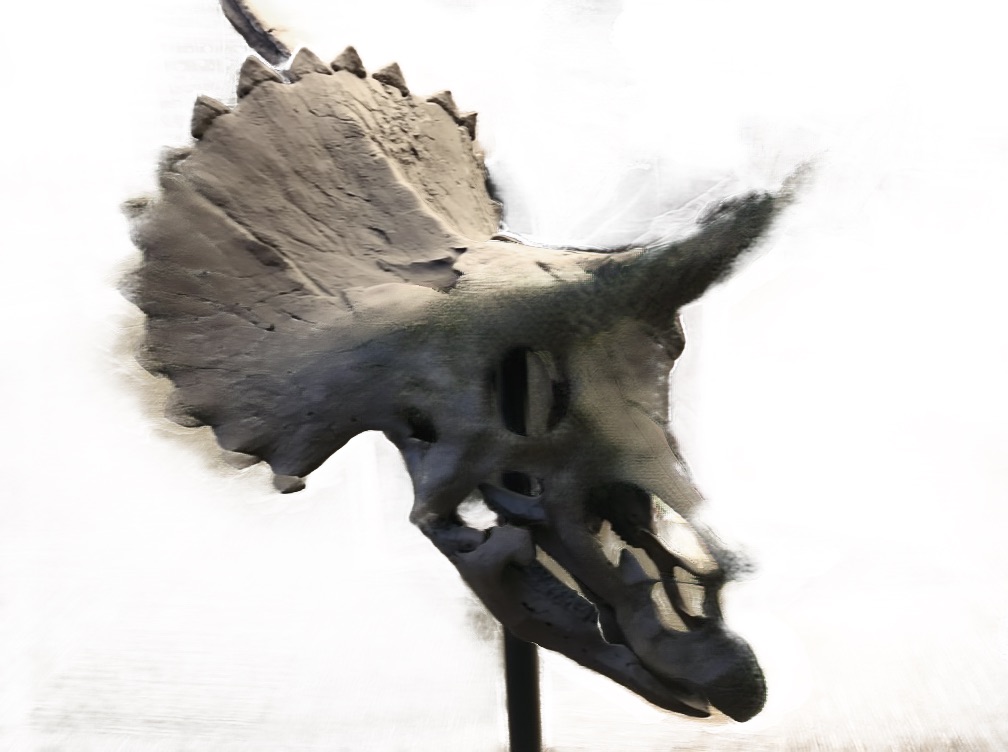} & \hspace{\mrgone}
\includegraphics[width=\wid]{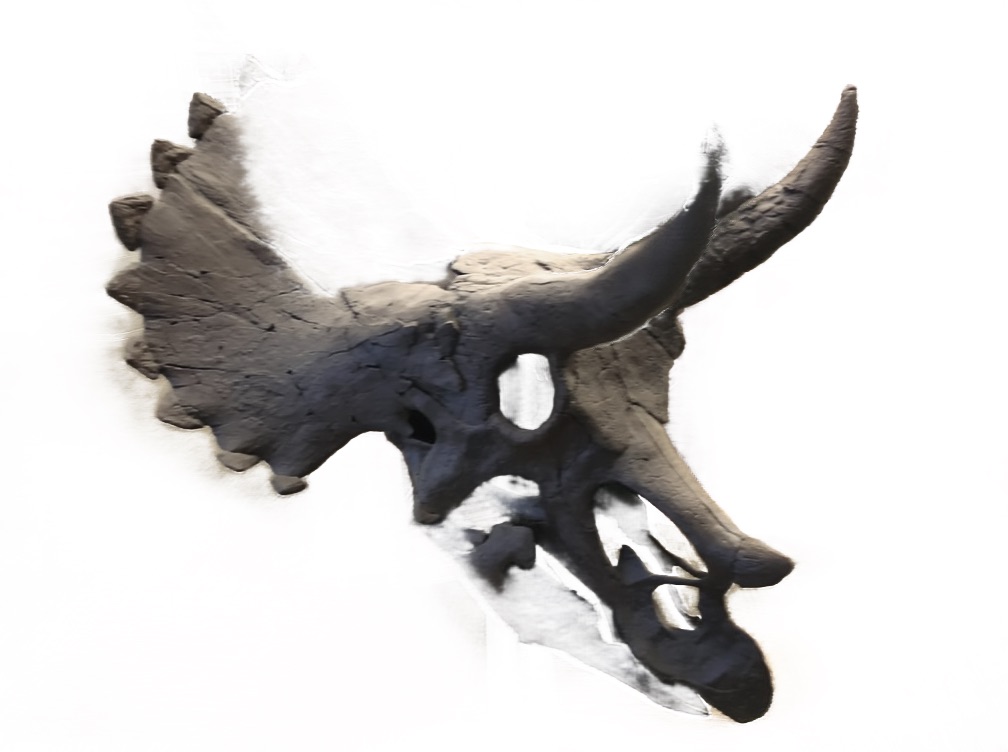}
\\

\hspace{\mrgone} \includegraphics[width=\wid]{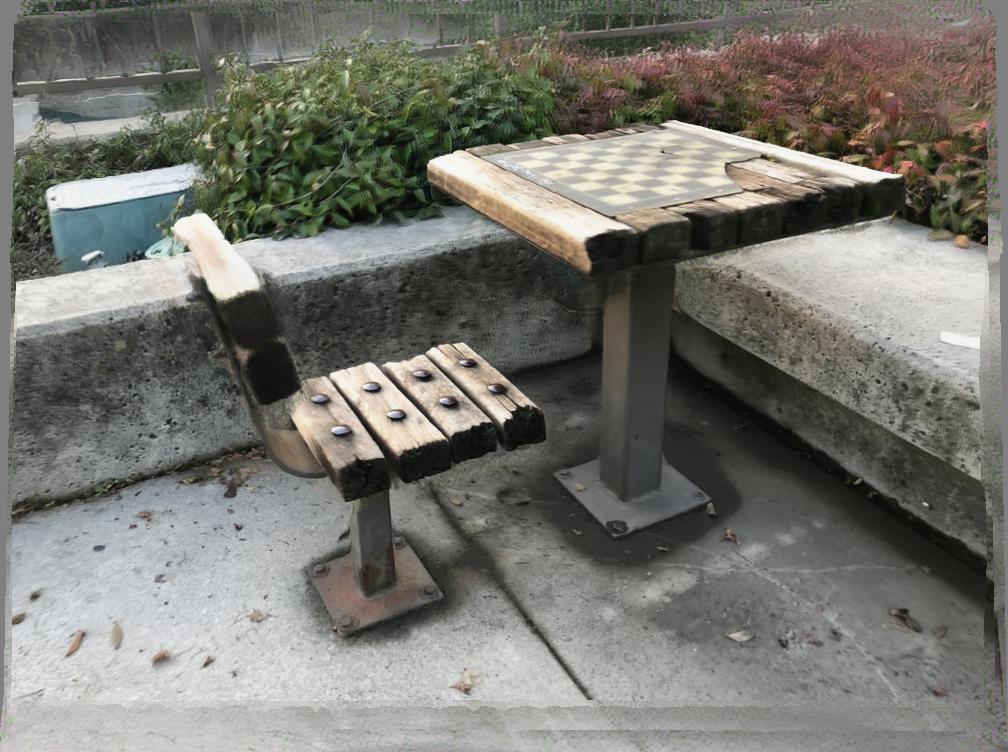} &  \hspace{\mrgone}
\hspace{\mrgone} \includegraphics[width=\wid]{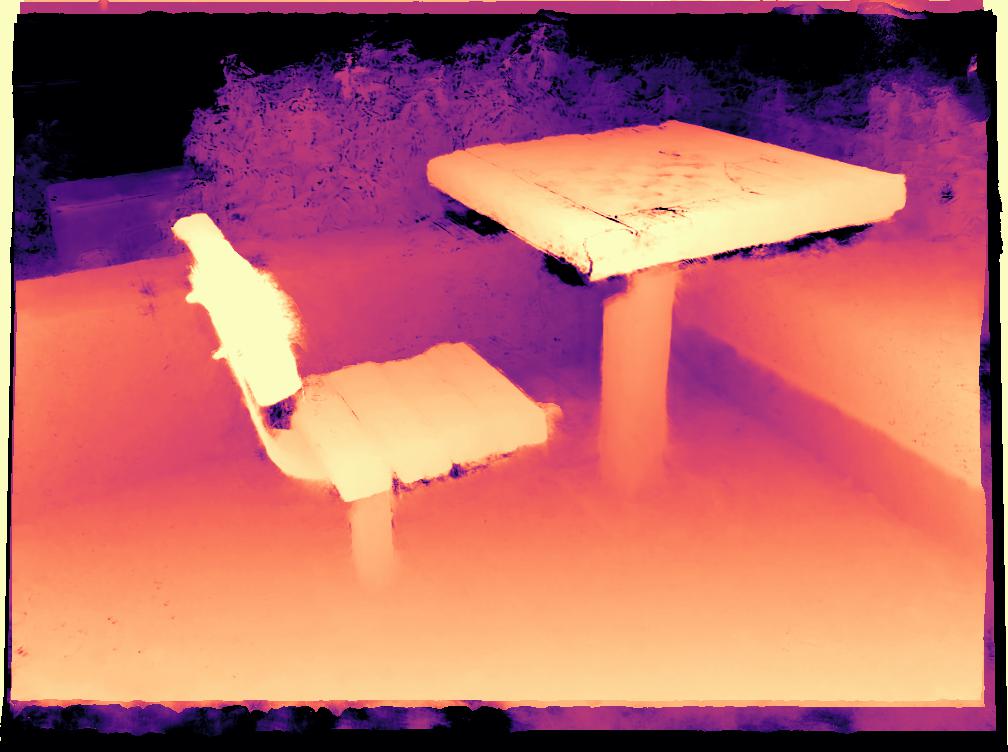} &  \hspace{\mrgone}
\includegraphics[width=\wid]{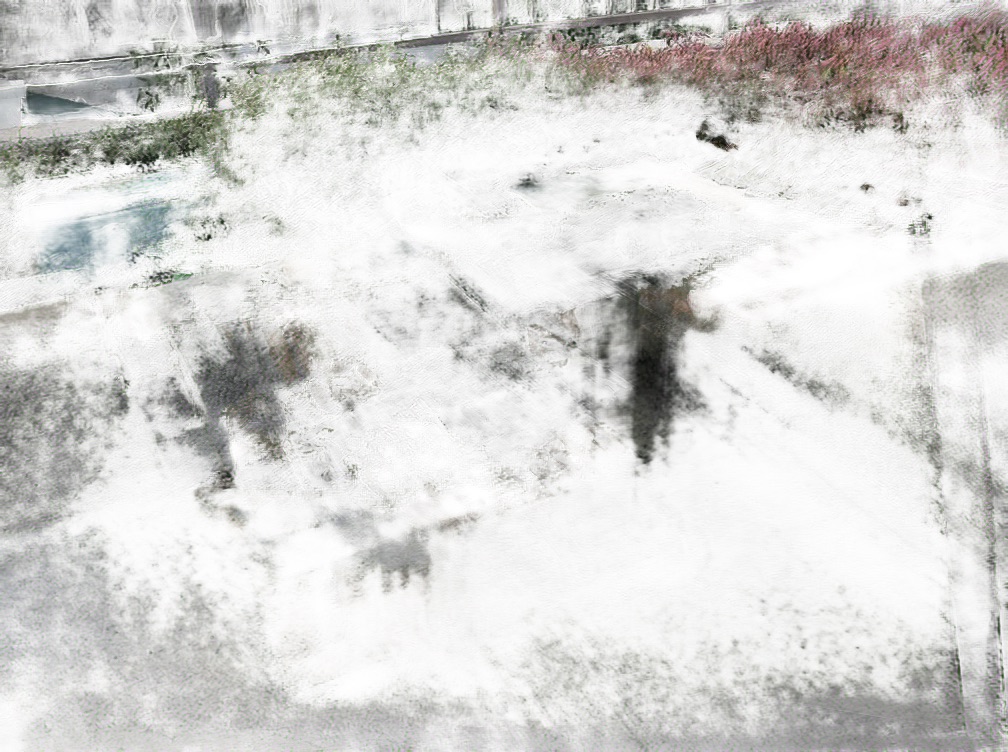} & \hspace{\mrgone}
\includegraphics[width=\wid]{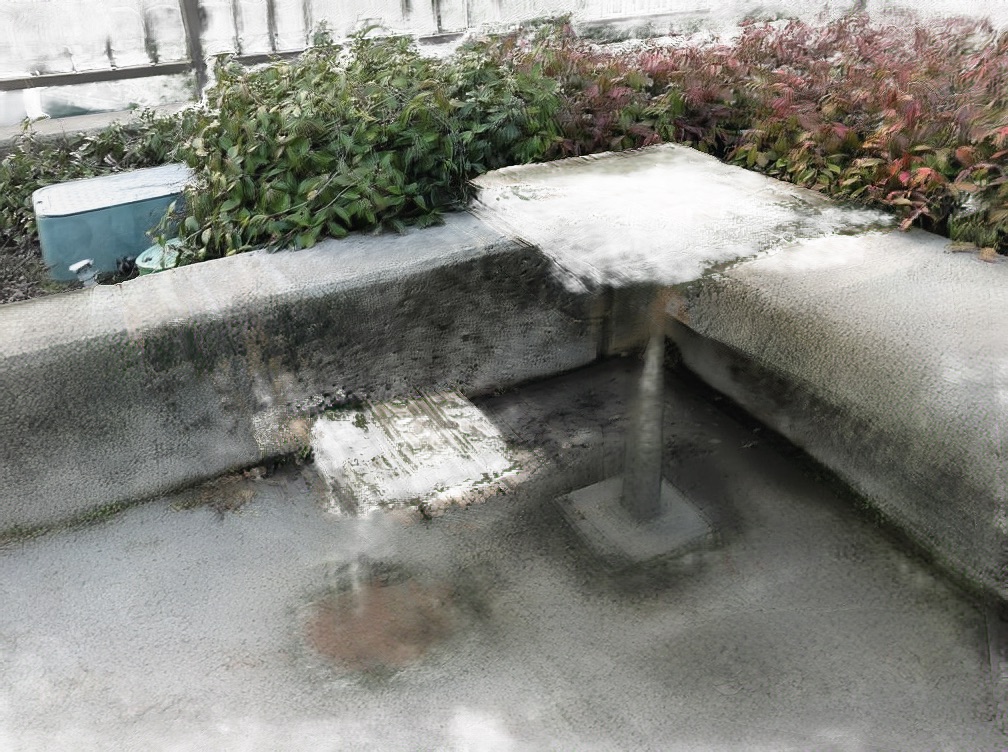} & \hspace{\mrgone}
\includegraphics[width=\wid]{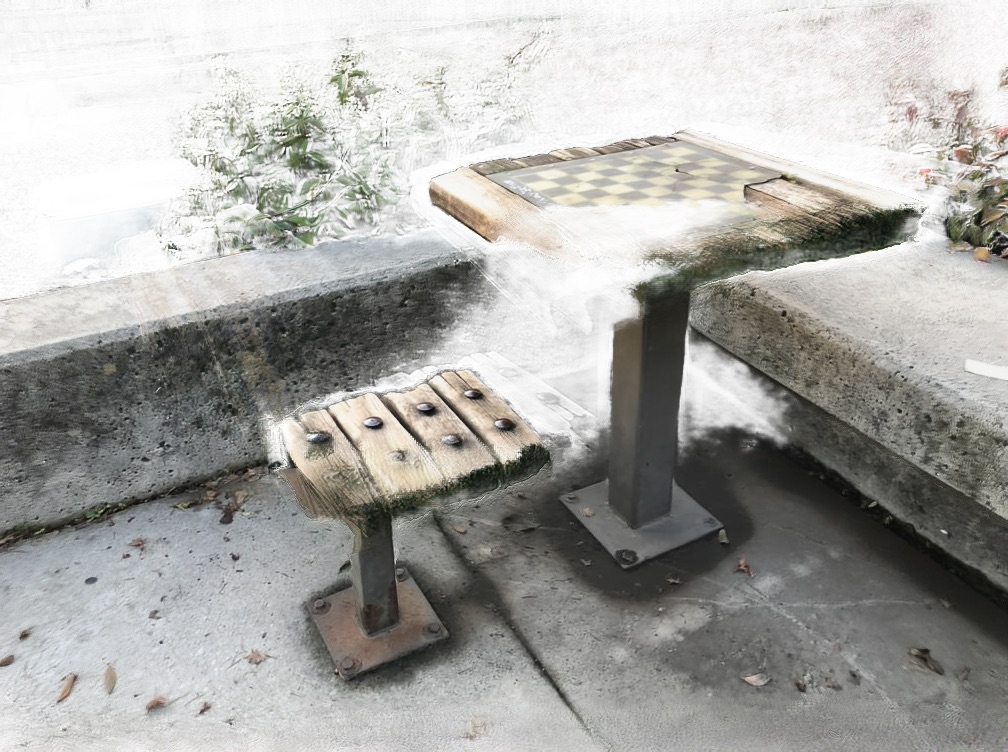} & \hspace{\mrgone}
\includegraphics[width=\wid]{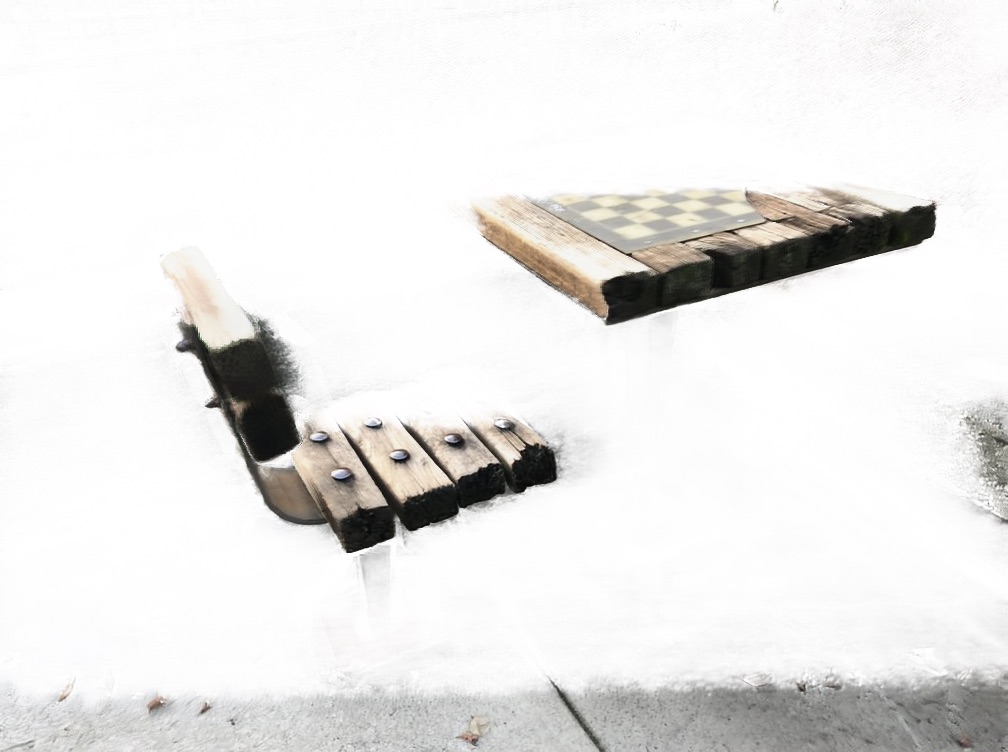}
\\

\hspace{\mrgone} \includegraphics[width=\wid]{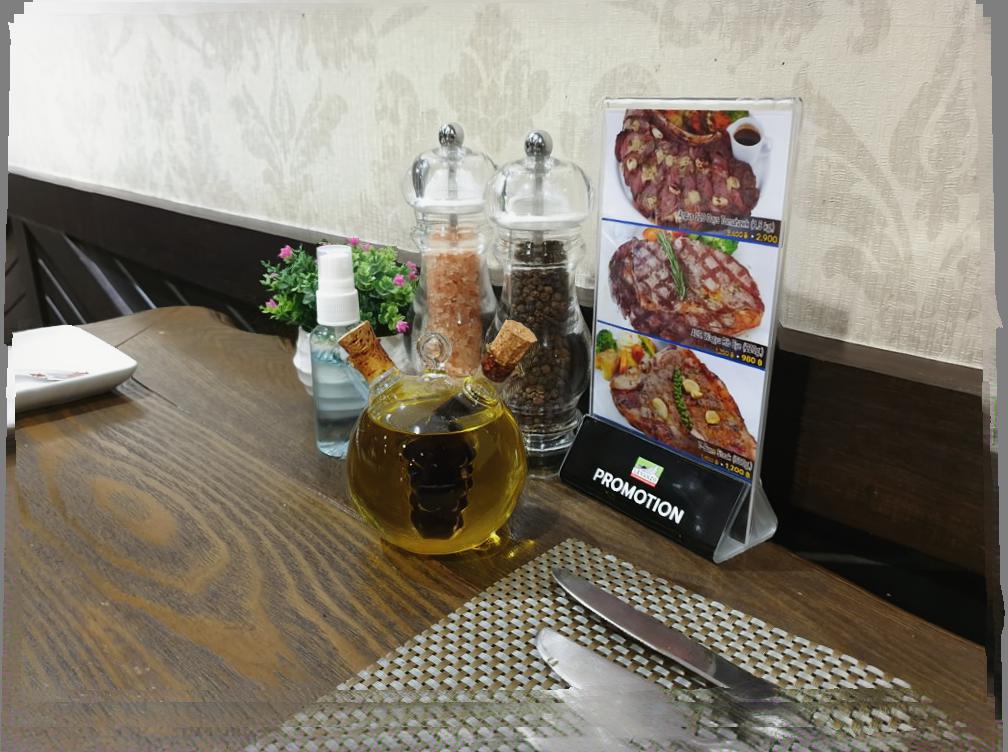} &  \hspace{\mrgone}
\hspace{\mrgone} \includegraphics[width=\wid]{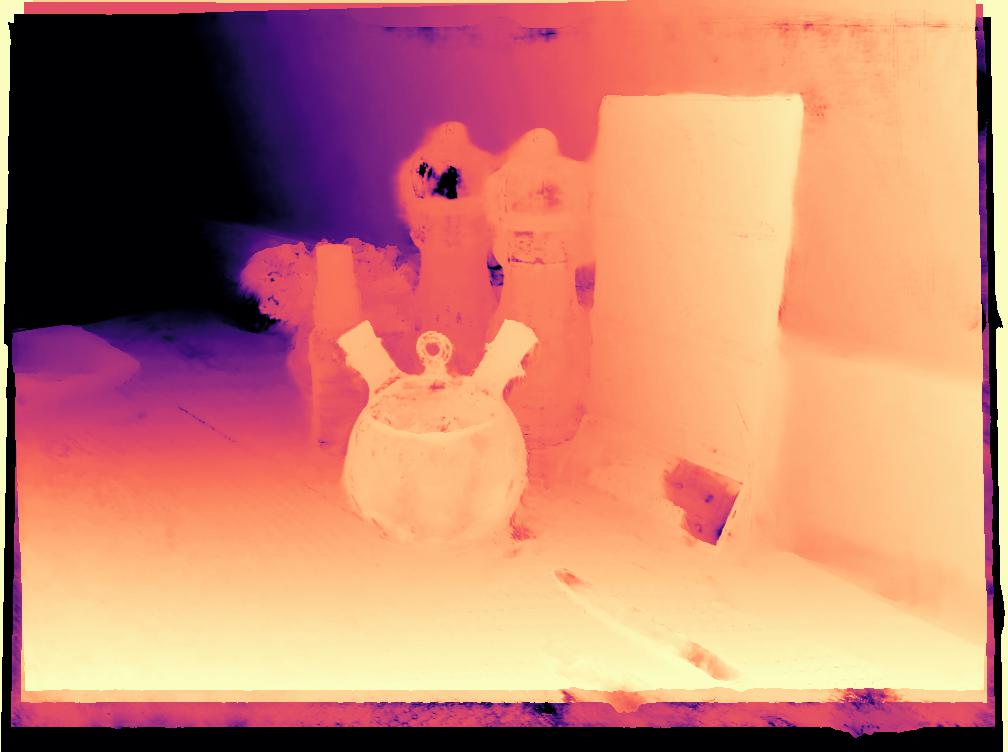} &  \hspace{\mrgone}
\includegraphics[width=\wid]{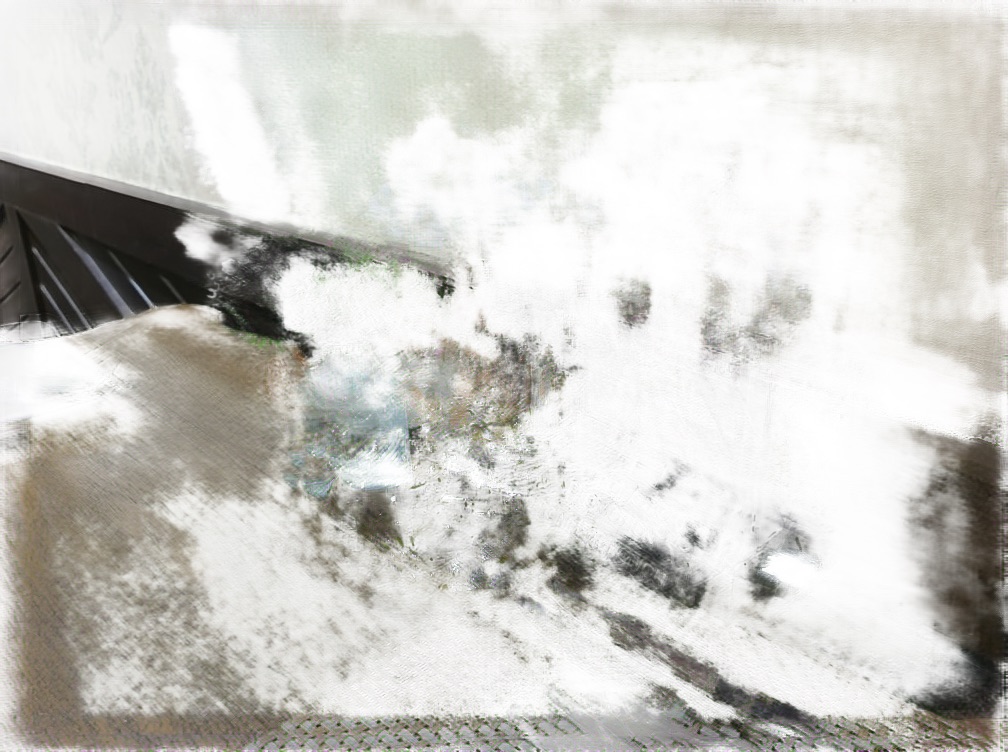} & \hspace{\mrgone}
\includegraphics[width=\wid]{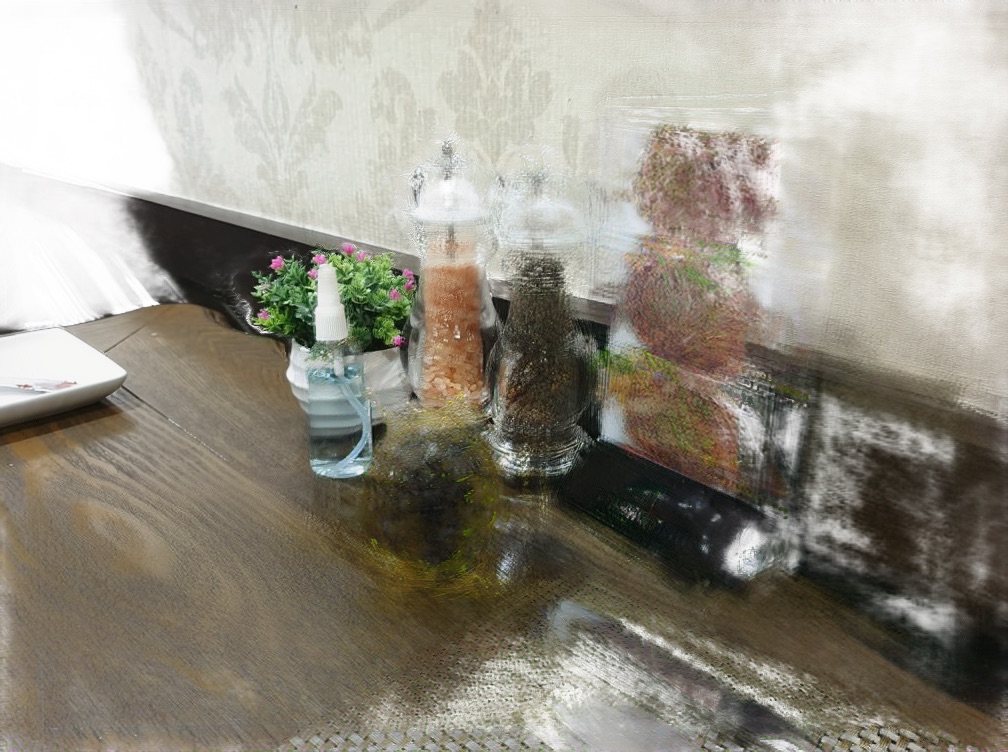} & \hspace{\mrgone}
\includegraphics[width=\wid]{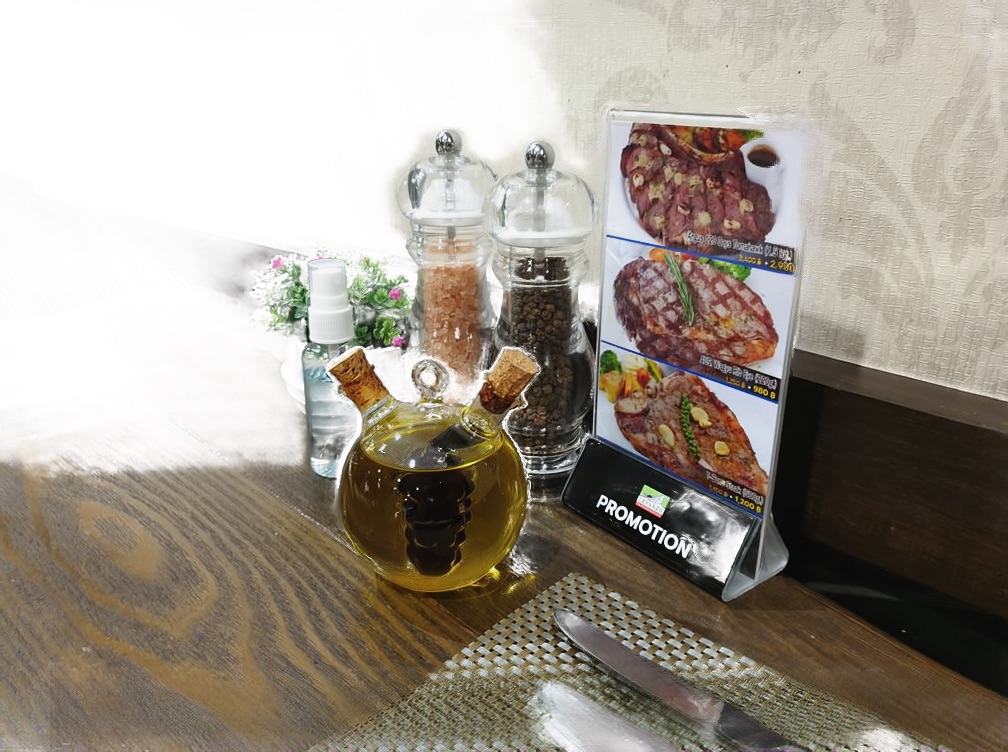} & \hspace{\mrgone}
\includegraphics[width=\wid]{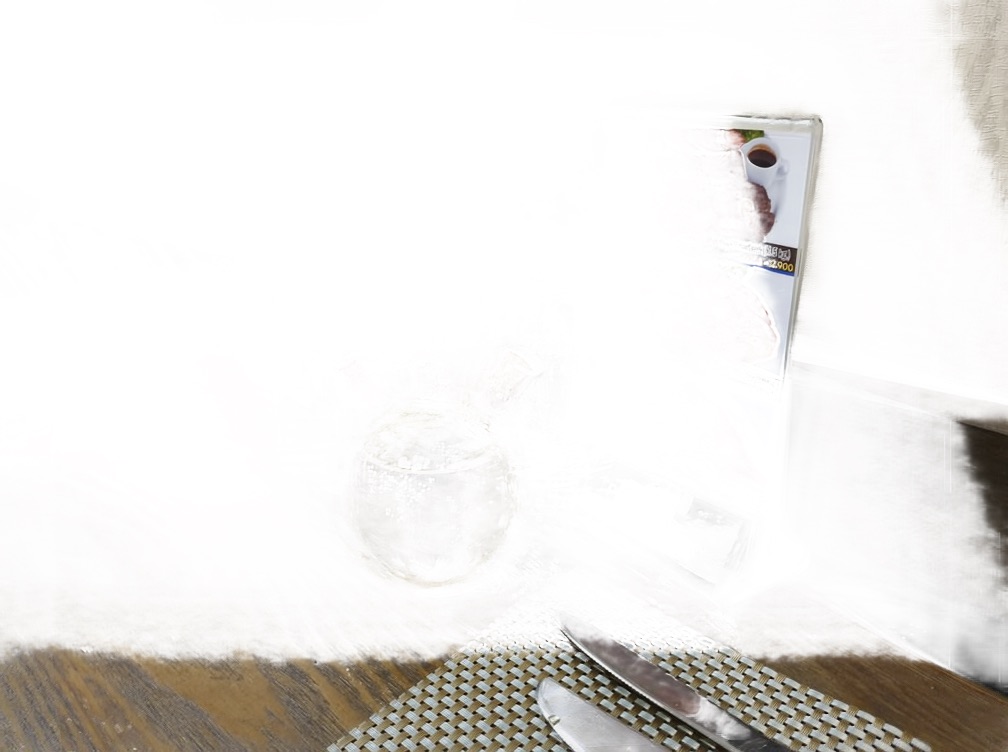}
\\

\hspace{\mrgone} \includegraphics[width=\wid]{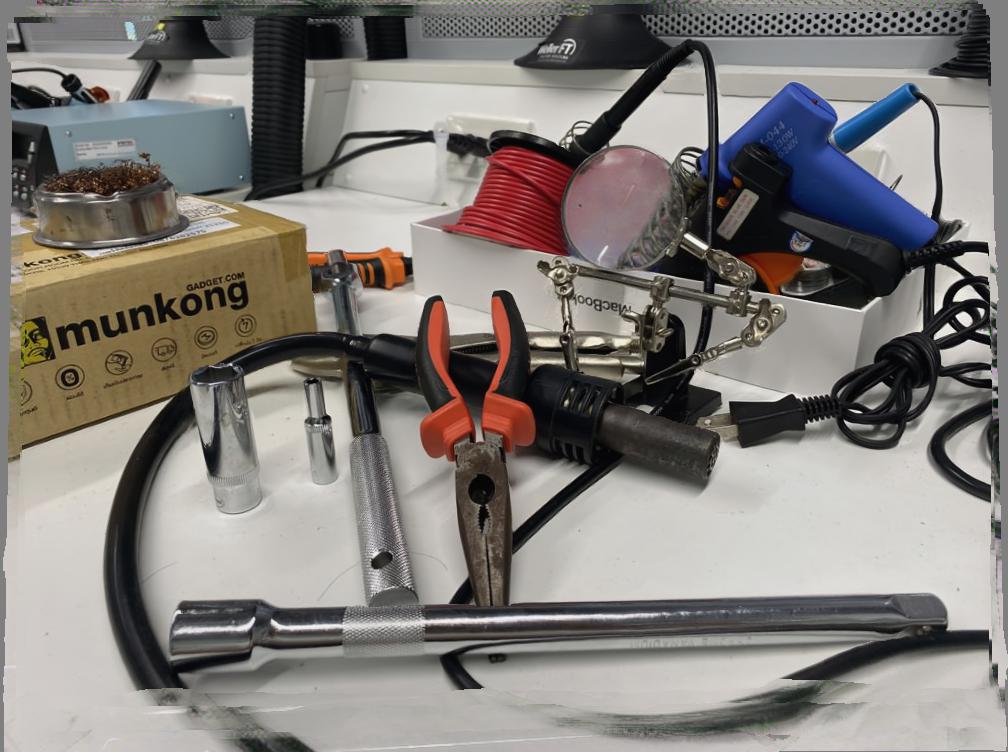} &  \hspace{\mrgone}
\hspace{\mrgone} \includegraphics[width=\wid]{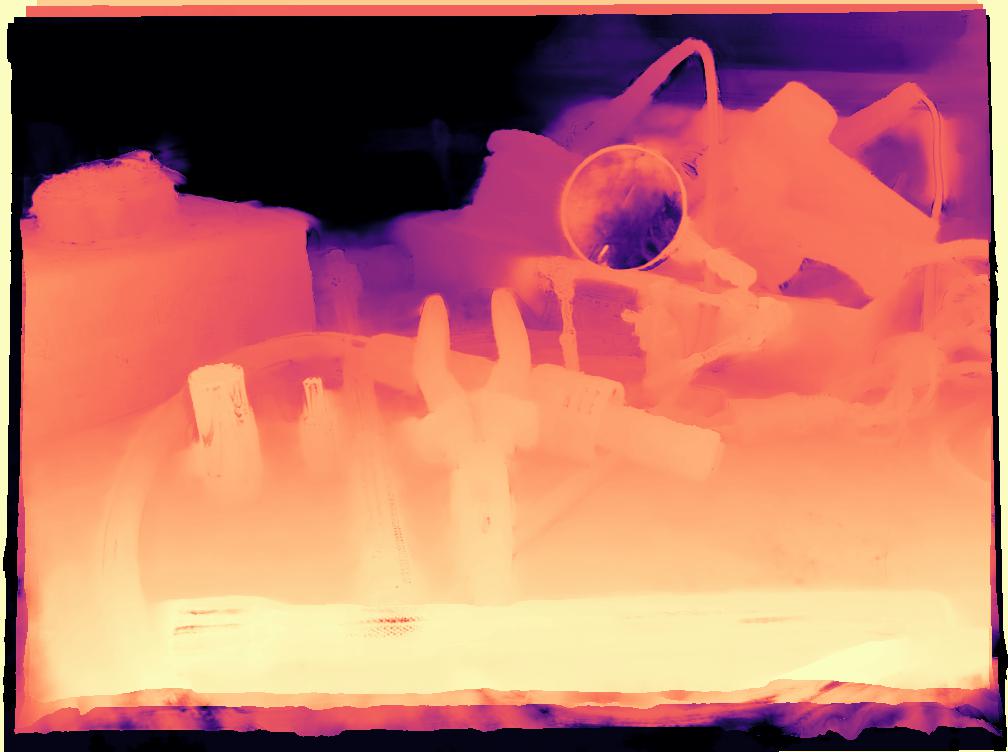} &  \hspace{\mrgone}
\includegraphics[width=\wid]{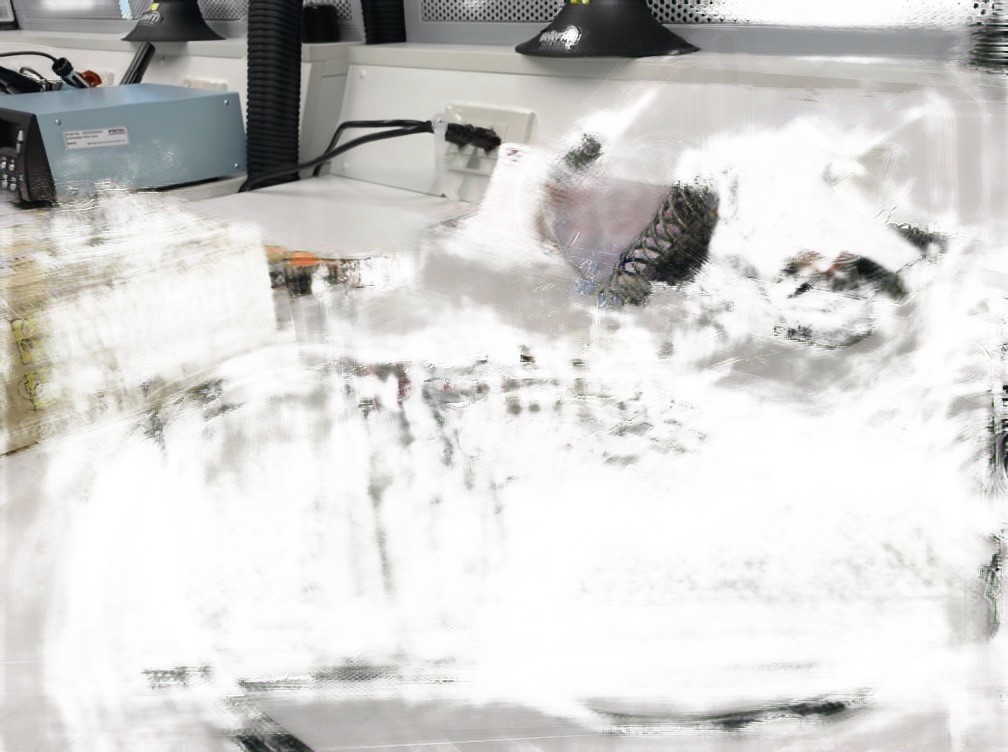} & \hspace{\mrgone}
\includegraphics[width=\wid]{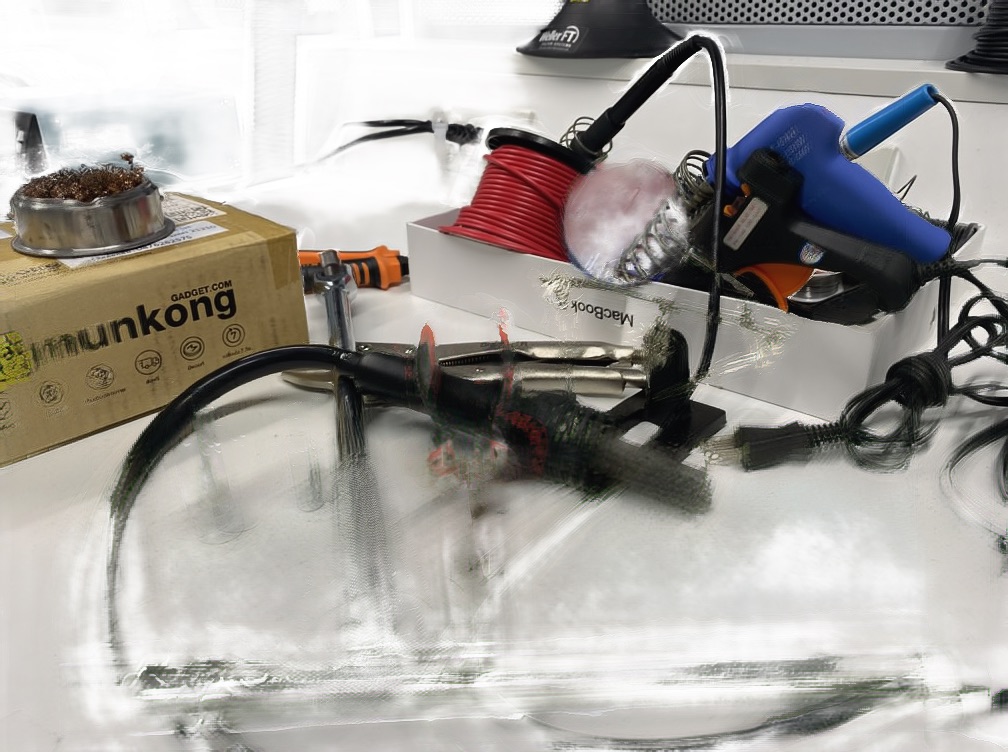} & \hspace{\mrgone}
\includegraphics[width=\wid]{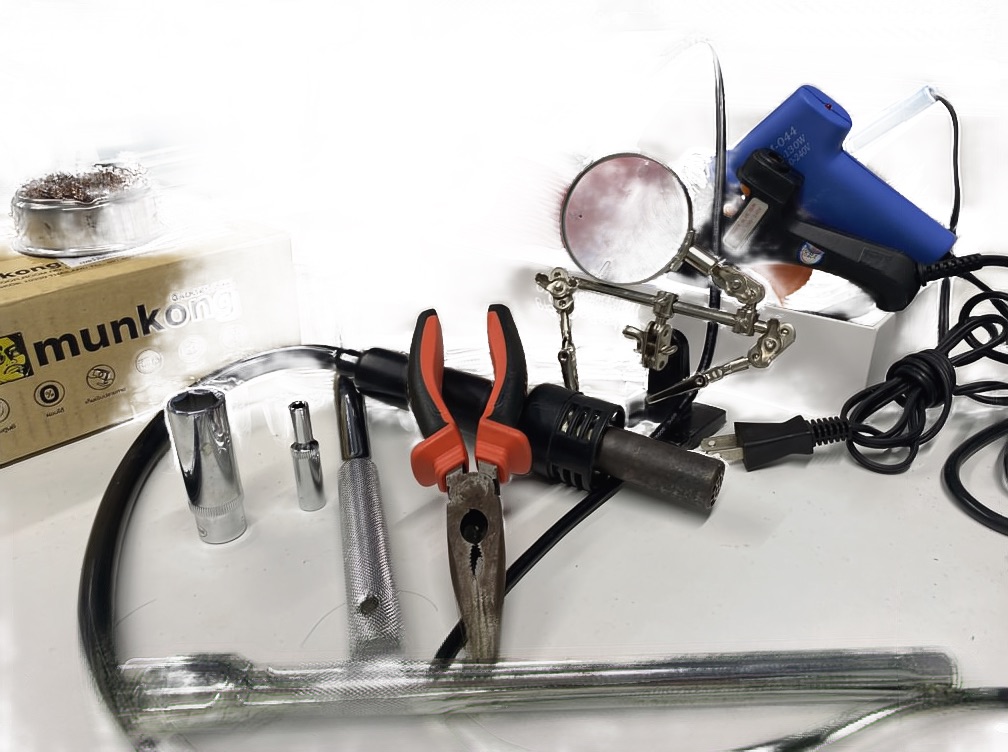} & \hspace{\mrgone}
\includegraphics[width=\wid]{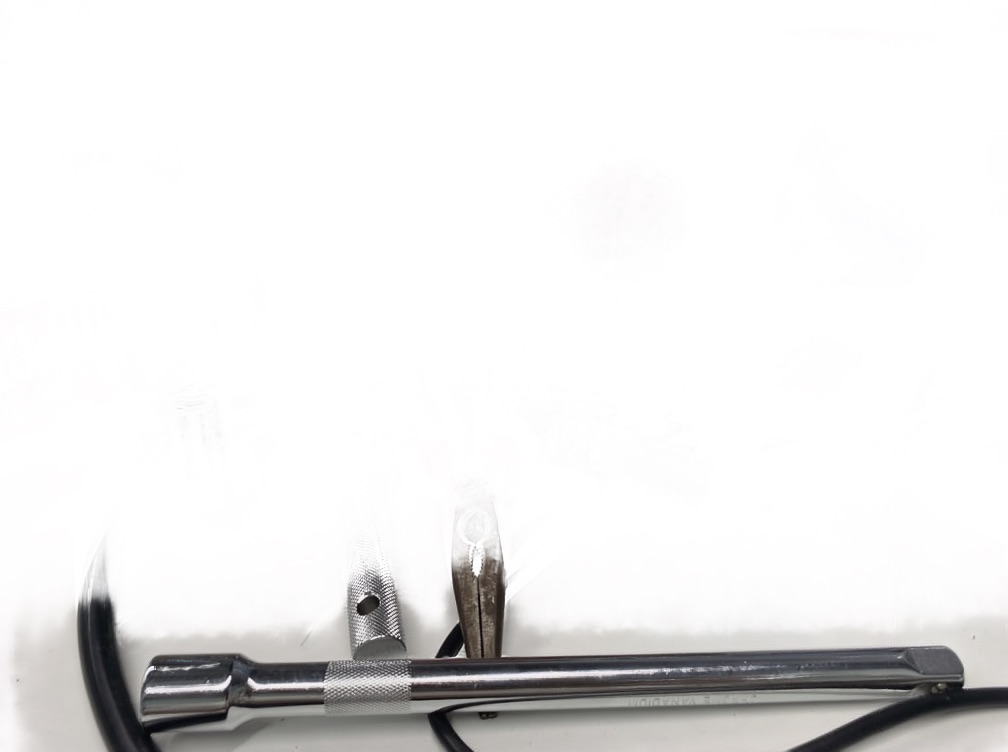}
\\

\hspace{\mrgone} \includegraphics[width=\wid]{wacv_figures/layers_depth/55/000046.jpg} &  \hspace{\mrgone}
\hspace{\mrgone} \includegraphics[width=\wid]{wacv_figures/layers_depth/55/000046_depth.jpg} &  \hspace{\mrgone}
\includegraphics[width=\wid]{wacv_figures/layers_depth/55/layer_00.jpeg} & \hspace{\mrgone}
\includegraphics[width=\wid]{wacv_figures/layers_depth/55/layer_01.jpeg} & \hspace{\mrgone}
\includegraphics[width=\wid]{wacv_figures/layers_depth/55/layer_02.jpeg} & \hspace{\mrgone}
\includegraphics[width=\wid]{wacv_figures/layers_depth/55/layer_03.jpeg}
\end{tabular}
\vspace{0.7ex}
\caption{Extension of \switchArxiv{\cref{fig:mli_representation}}{Fig.~1 from the main text}.
The textures of MLI representation  with 4 deformable layers estimated by \modelname. 
Left to right: generated novel view, corresponding depth map, four semitransparent textures in the back-to-front order.
The inferred depth map is computed by overcomposing the per-layer depth maps \wrt the opacity extracted from the corresponding RGBA textures.
}
\label{fig:mli_representation_appendix}
\vspace{-0.25cm}
\end{figure*}

\begin{figure*}[t]
    \vspace{0pt}
    \centering
    \begin{minipage}{0.8\textwidth}
    \makeatletter
    \@for\items:={Full image, Ground truth,IBRNet,LLFF,\modelname-2L,\modelname-4L,\modelname-8L}%
    \do{\minipage{0.14\textwidth}\centering\scriptsize \items \endminipage\hfill}%
    \makeatother\\
    \includegraphics[width=\textwidth]{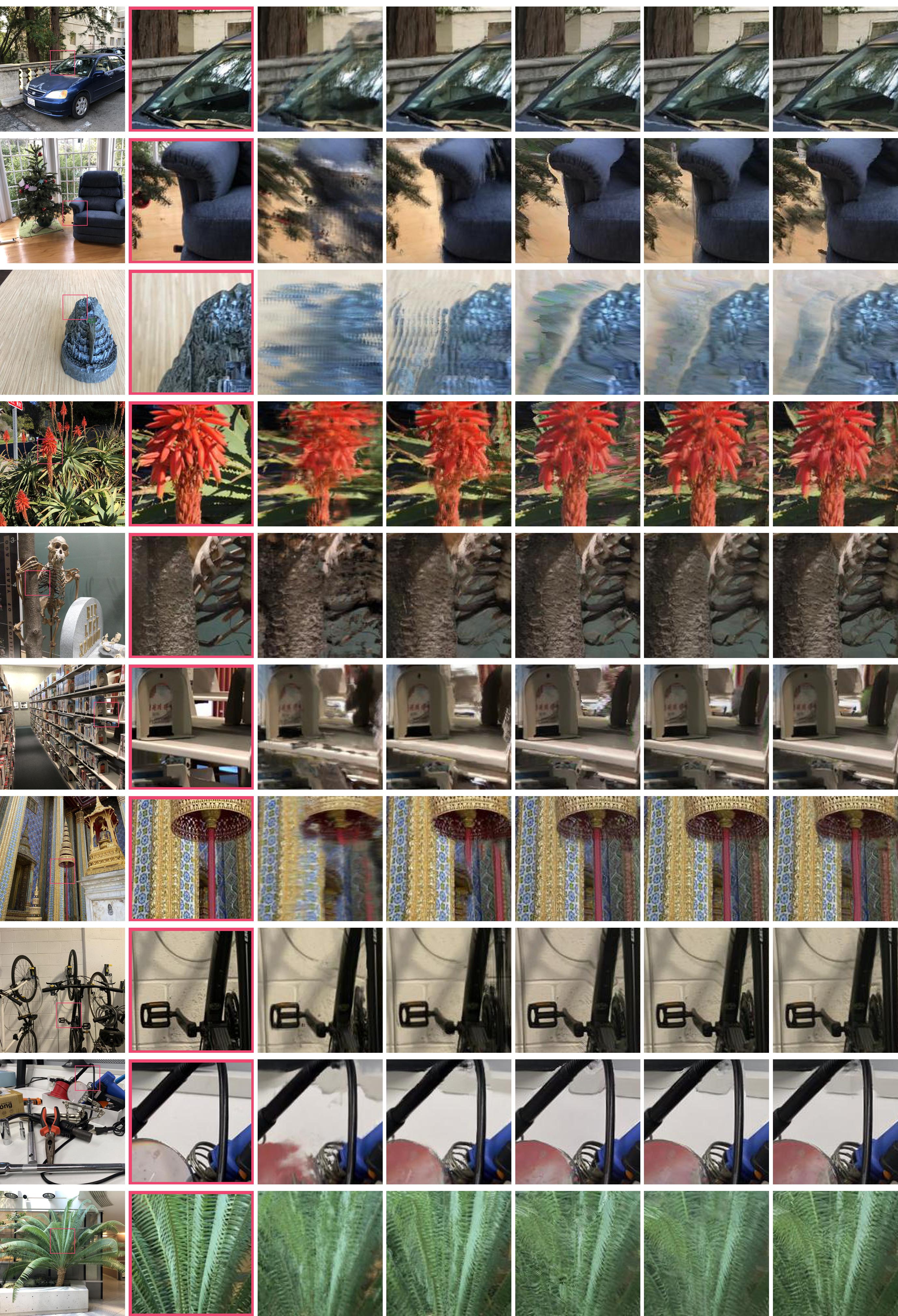}
    \end{minipage} 
    \vspace{0.7ex}
    \caption{ 
        Results for real novel cameras with \textbf{two} input views given.
        Note that IBRNet cannot produce any information for areas unobserved from the source views.
    }
\label{fig:2views}
\end{figure*}

\begin{figure*}[t]
    \vspace{0pt}
    \centering
    \begin{minipage}{0.8\textwidth}
    \makeatletter
    \@for\items:={Full image, Ground truth,IBRNet,LLFF,\modelname-2L,\modelname-4L,\modelname-8L}%
    \do{\minipage{0.14\textwidth}\centering\scriptsize \items \endminipage\hfill}%
    \makeatother\\
    \includegraphics[width=\textwidth]{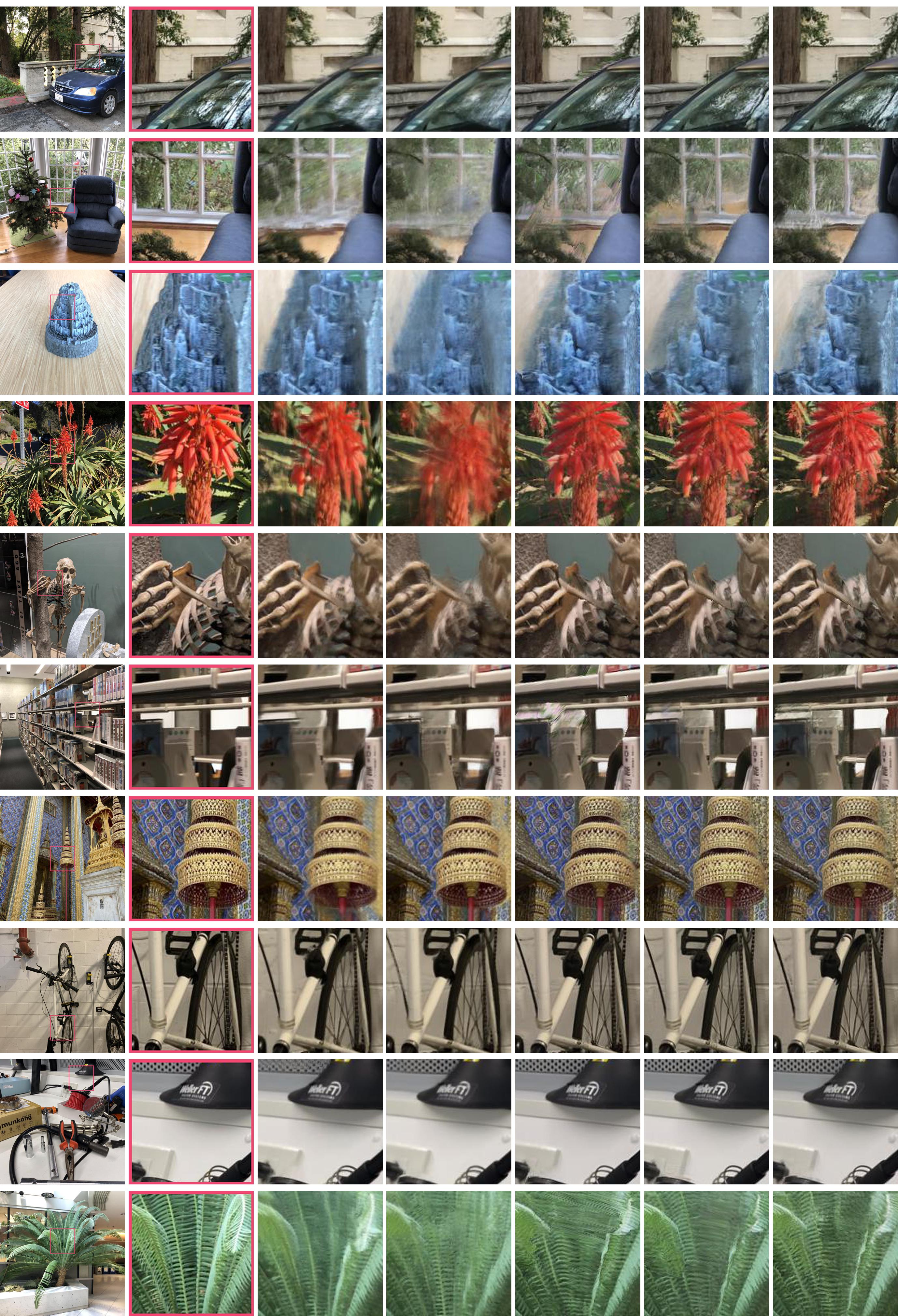}
    \end{minipage} 
    \vspace{0.7ex}
    \caption{ 
        Results for real novel cameras with \textbf{five} input views given.  
        The outputs of \modelname-8L are the most similar to the ground truth frames and have less artifacts than other models.
    }
\label{fig:5views}
\end{figure*}

\begin{figure*}[t]
    \vspace{0pt}
    \centering
    \begin{minipage}{0.8\textwidth}
    \makeatletter
    \@for\items:={Full image, Ground truth,IBRNet,LLFF,\modelname-2L,\modelname-4L,\modelname-8L}%
    \do{\minipage{0.14\textwidth}\centering\scriptsize \items \endminipage\hfill}%
    \makeatother\\
    \includegraphics[width=\textwidth]{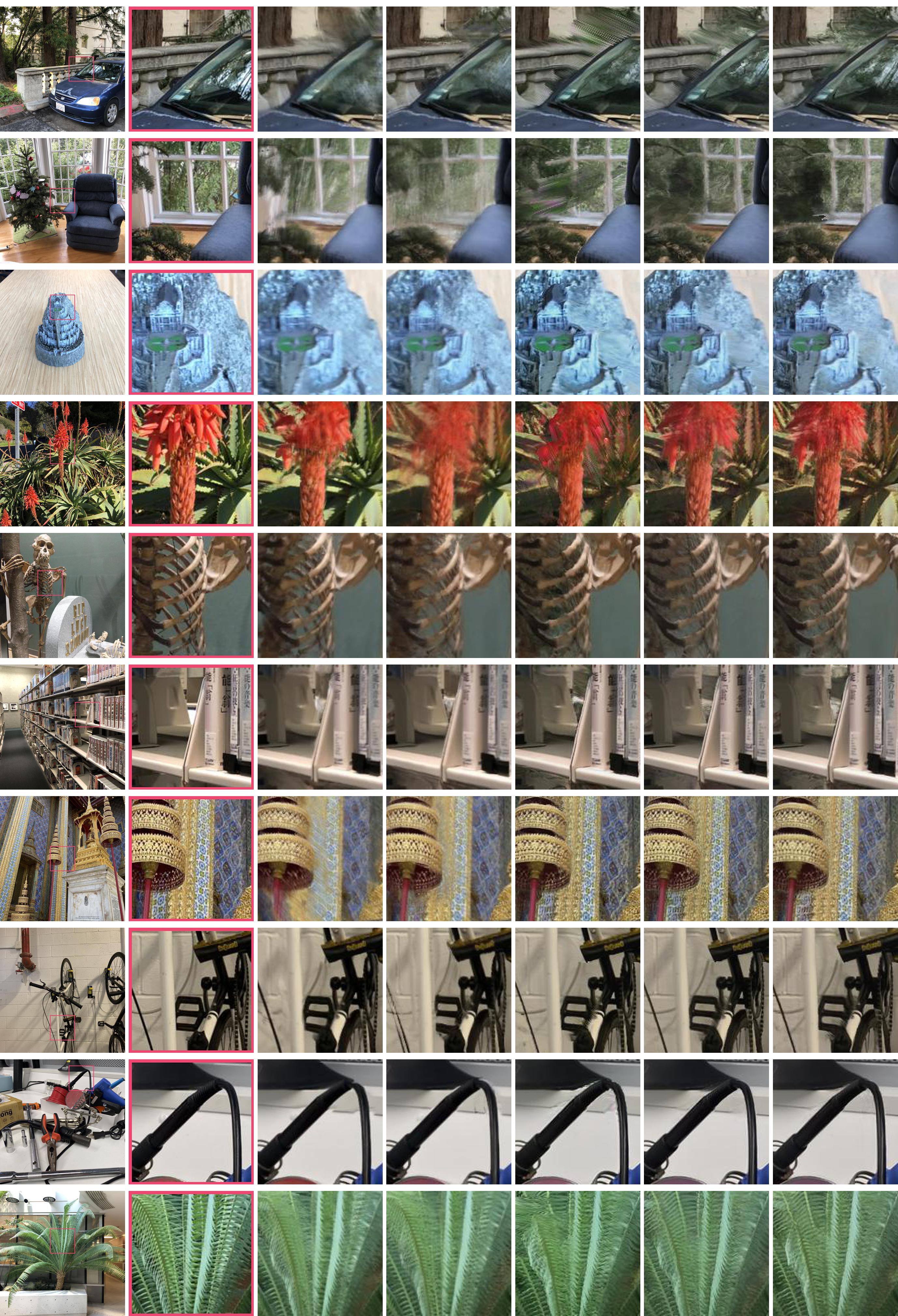}
    \end{minipage} 
    \vspace{0.7ex}
    \caption{ 
        Results for real novel cameras with  \textbf{eight} input views given.
        This is the most competitive scenario. 
        Note that both IBRNet and LLFF tend to produce many artifacts, \eg blurriness, while frames rendered by \modelname are more sharp.
    }
\label{fig:8views}
\end{figure*}

\cref{fig:mli_representation_appendix} demonstrates the results of our \modelname method in the case of 4 layers.
Also we provide an additional comparison with the baseline methods on publicly available datasets~\cite{ibrnet,llff}. 
We show visual results for 2 input views in~\cref{fig:2views}, for 5 input views in~\cref{fig:5views} and for 8 -- in~\cref{fig:8views}.
These results correlate with the metrics reported in the main text. 
For two or five input views, our method clearly outperforms all baselines and produces the most visually pleasant results.
Also, most of the crops for the eight input images show that our method is at least on par with existing approaches.
Note that all the demonstrated crops and scenes are uncurated.

\begin{figure*}[t]
    \centering
    \begin{minipage}{\linewidth}
    \makeatletter
    \begin{minipage}{0.49\linewidth}
    \@for\items:={Ground truth,DeepView, IBRNet,\modelname-4L,\modelname-8L}%
    \do{\minipage{0.2\linewidth}\centering\scriptsize \items \endminipage\hfill}%
    \end{minipage}\hfill
    \begin{minipage}{0.49\linewidth}
    \@for\items:={Ground truth,DeepView, IBRNet, \modelname-4L,\modelname-8L}%
    \do{\hfill\minipage{0.2\linewidth}\centering\scriptsize \items \endminipage}%
    \end{minipage}%
    \makeatother\\
    \begin{tabular}{c|c}
    \includegraphics[width=0.49\linewidth]{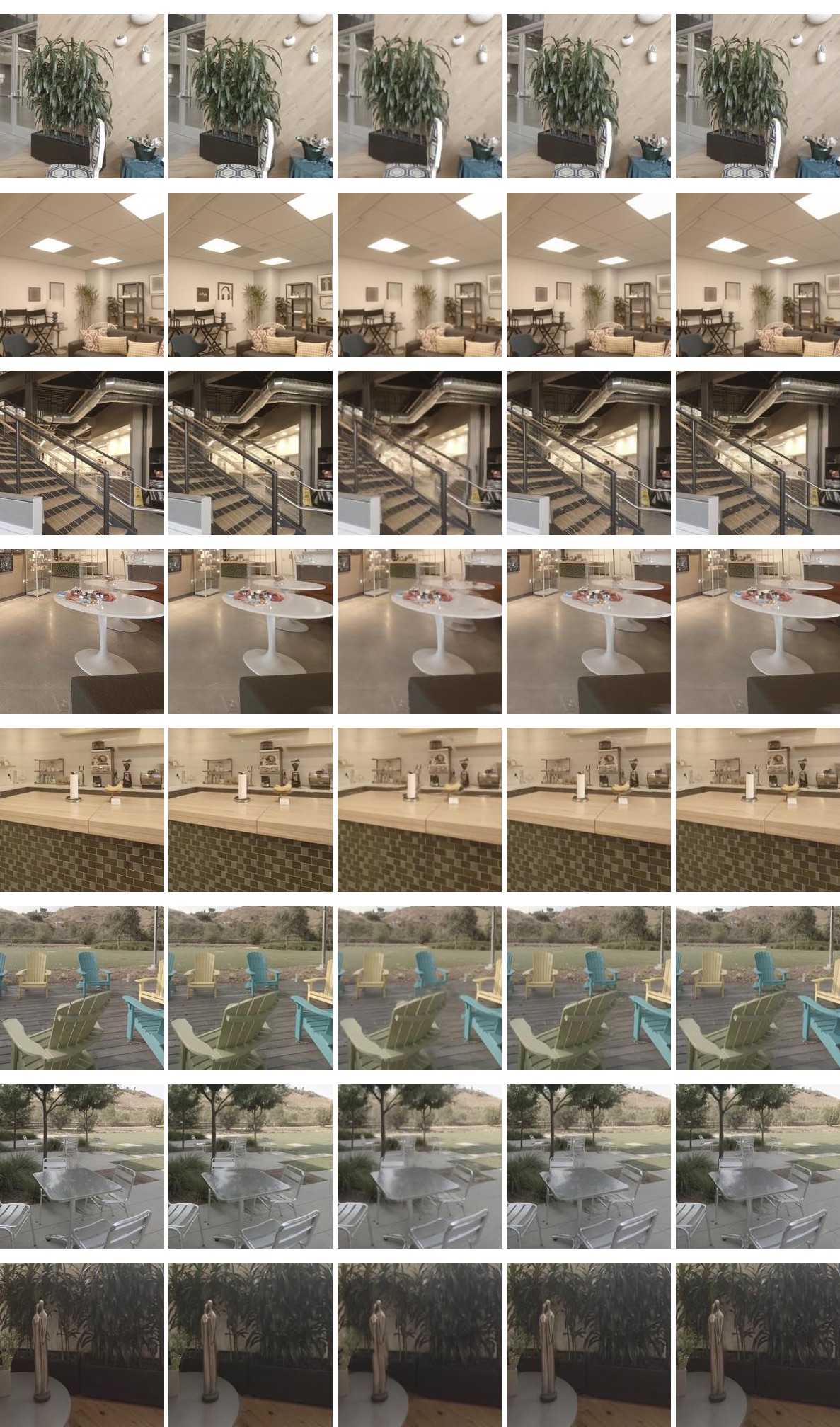} 
    & \includegraphics[width=0.49\linewidth]{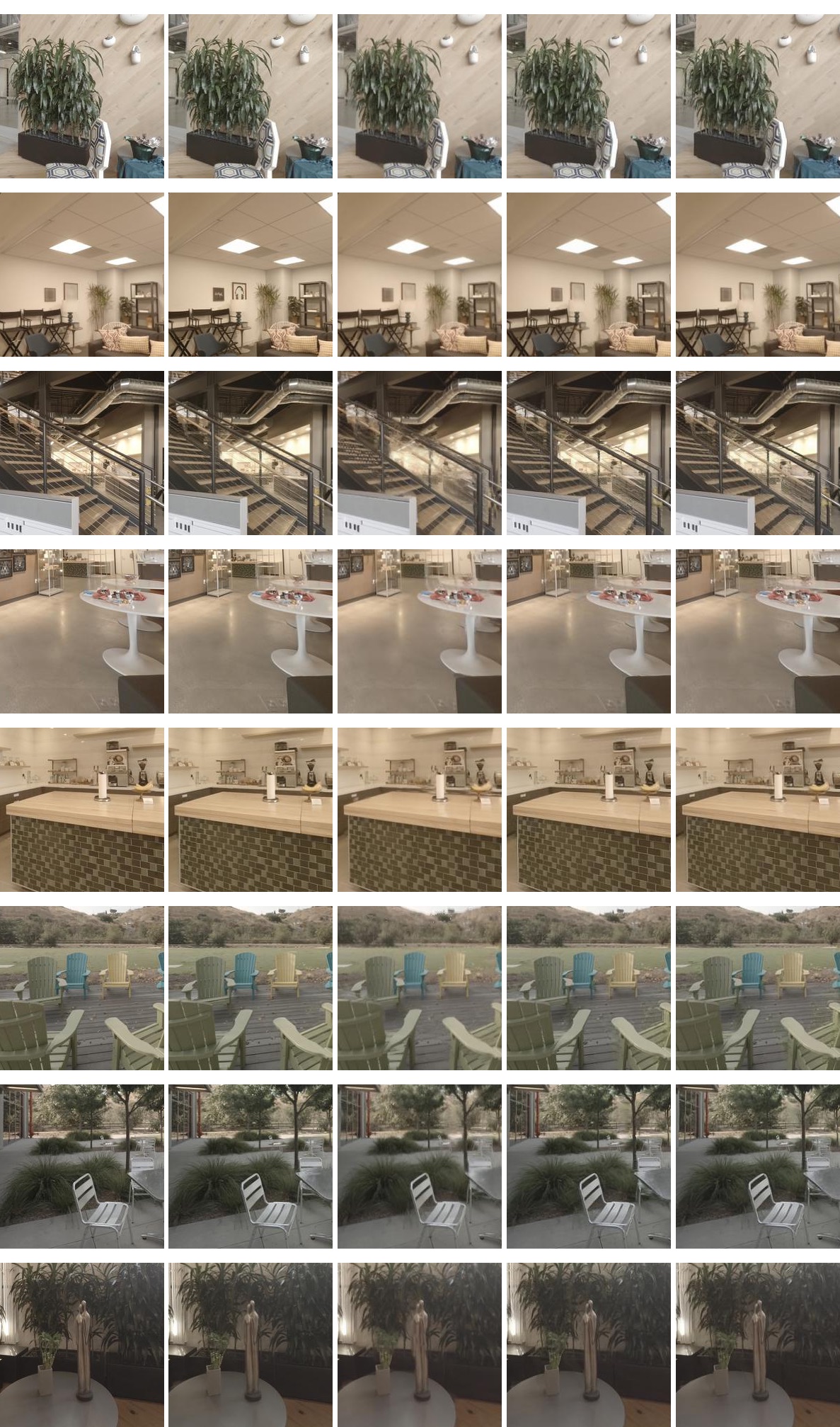} \\
    \textbf{Small camera baseline} &  \textbf{Large camera baseline}
    \end{tabular}
    \end{minipage}
    \vspace{0.7ex}
    \caption{
        Results for the Spaces dataset for DeepView (40 planes) and \modelname with 4 and 8 layers. 
        Our model produces more blurry and less bright results, trading off for more compact representation of the scene.
    }
\label{fig:spaces_results}
\end{figure*}

\cref{fig:spaces_results} demonstrates a comparison of our model with the DeepView system~\cite{deep_view} on the Spaces dataset~\cite{deep_view}.
We show the results for the small and large camera baselines separately.
As may be seen from the figures, our model produces slightly blurrier results than DeepView does. 
However, it allows us to get a much more compact scene representation, as was discussed in the main text.
We demonstrate the visual comparison for \modelname with a different number of layers in the MLI representation in~\cref{fig:2views,fig:5views,fig:8views}.
We observe the degradation of the quality with decreasing of the number of layers in the final representation.

{\small
\bibliographystyle{ieee_fullname}
\bibliography{egbib}
}

\end{document}